\documentclass[journal]{IEEEtran}
\usepackage{ragged2e} 
\usepackage{float} 
\usepackage{graphicx,subfigure}
\usepackage[table,xcdraw]{xcolor}
\usepackage{amsmath,amsfonts}
\usepackage{algorithmic}
\usepackage{algorithm}
\usepackage{array}
\usepackage[caption=false,font=normalsize,labelfont=sf,textfont=sf]{subfig}
\usepackage{textcomp}
\usepackage{stfloats}
\usepackage{url}
\usepackage{verbatim}
\usepackage{graphicx}
\usepackage{cite}
\usepackage{subfig}
\usepackage{mathrsfs}
\usepackage{multirow}
\usepackage{booktabs}
\usepackage {color}
\usepackage{microtype}

\usepackage{cuted} 
\usepackage{lipsum}

\begin{document}
%
\title{High-order Knowledge Based Network Controllability Robustness Prediction: A Hypergraph Neural Network Approach}

\author{Shibing~Mo,
        Jiarui~Zhang\textsuperscript{†},
        Jiayu~Xie\textsuperscript{†},
        Xiangyi~Teng*,~\IEEEmembership{Member,~IEEE},
        and~Jing~Liu,~\IEEEmembership{Senior Member,~IEEE}
\IEEEcompsocitemizethanks{
\IEEEcompsocthanksitem This work was supported in part by the National Natural Science Foundation of China under Grant 62306224 and Grant 62471371, in part by the Guangdong Highlevel Innovation Research Institution Project under Grant 2021B0909050008, in part by the Guangzhou Key Research and Development Program under Grant 202206030003 and in part by the Fundamental Research Funds for the Central Universities under Grant YJSJ25014. 
\IEEEcompsocthanksitem Shibing Mo, Jiarui Zhang and Jing Liu are with the School of Artifcial Intelligence, Xidian University, Xian 710126, China. Jiayu Xie is with the School of Computer Science and Technology, Xidian University, Xian 710126, China. Xiangyi Teng is with the Guangzhou Institute of Technology, Xidian University, Guangzhou 510555, China.\protect
\IEEEcompsocthanksitem Email:~msb@stu.xidian.edu.cn;~2901999089@qq.com;~3630729140@qq.com;
~tengxiangyi@xidian.edu.cn;~neouma@mail.xidian.edu.cn.
\IEEEcompsocthanksitem Xiangyi~Teng~is~the~corresponding~author.
\IEEEcompsocthanksitem Authors marked with † contributed equally to this work.
\IEEEcompsocthanksitem The source code and supplementary materials for our work are available at https://github.com/Explorermomo/NCR-HoK.
}}
\markboth{IEEE TRANSACTIONS ON NETWORK SCIENCE AND ENGINEERING}%
{Shell \MakeLowercase{\textit{et al.}}: A Sample Article Using IEEEtran.cls for IEEE Journals}

\maketitle

\begin{abstract}
\justifying
In order to evaluate the invulnerability of networks against various types of attacks and provide the guidance for potential performance enhancement as well as controllability maintenance, network controllability robustness (NCR) has attracted increasing attention in recent years. Traditionally, the controllability robustness is determined by attack simulations, which is computationally time consuming and only applicable for small-scale networks. Although some machine learning based methods for predicting network controllability robustness have been proposed lately, they only focus on pairwise interactions of complex networks and the underlying relationships between high-order structural information and network controllability robustness are not explored. In this paper, a dual hypergraph attention neural network model based on high-order knowledge (NCR-HoK) is proposed to accomplish the tasks of robustness learning and controllability robustness curve prediction. Through the proposed node feature encoder, the construction of hypergraph with high-order relation and dedicatedly designed dual hypergraph attention module, our method can effectively learn three types of network information simultaneously: the explicit structural information in the original graph, the high-order connection information in local neighborhoods and more hidden features in the embedding space. Notably, we explore for the first time the impact of high-order knowledge on the network’s controllable robustness. Compared to the state-of-the-art methods for network robustness learning, our method has superior performance on both synthetic and real-world networks with low overheads.
\end{abstract}

\begin{IEEEkeywords}
Complex network, hypergraph, controllability, graph attention neural network, controllability learning.
\end{IEEEkeywords}

\section{Introduction}\label{sec:introduction}

\IEEEPARstart{C}{omplex} network is an interdisciplinary subject covering computational mathematics, physics, computer, biology, and other fields \cite{1,2,3,4}. Nowadays,  researches on complex network can help us analyze the composition of large-scale systems, the operating dynamics of network members and the distribution rules of functional structures. Therefore, network science has received enormous attention in recent years \cite{6,10,13,17}.  Generally, systems with different structures undertake different types of tasks and inevitably will face complicated and changeable application scenarios. Various external factors such as natural disasters or man-made attacks could interfere with the normal operation of the system \cite{+1}. Under such circumstances, a robust network needs to maintain relatively stable functions even when it suffers certain attacks or damage \cite{5}. In order to evaluate the invulnerability or controllable capacity of different networks and provide a reasonable reference for potential performance enhancement, network controllability robustness has become one of the core issues in the field of network science \cite{14,15,16,18,19}.

Generally speaking, the measure of the network controllability robustness is quantified by a sequence of values that record the remaining controllability of the network after a sequence of attacks \cite{8, 9, 11,12}. Here, the attacks can be divided into random attack or malicious attack. The former means that the attacker randomly selects targets from the network to attack, and the latter indicating that the attacker targets the most important members (nodes or edges) of the current network. For a random network, the distribution of the significance of network members follows Poisson distribution; that is, the importance of each node is similar and there are no apparent key nodes. Therefore, random attacks are relatively more destructive to this kind of network since random network has strong tolerance for malicious attacks \cite{20}. For scale-free networks, the distribution of network members exhibits a power law distribution characteristic; that is, very few nodes are critical and need to be highly protected. This makes the network very vulnerable to malicious attacks but robust to random attacks \cite{21}.

Although the correlation between local topological structure and network controllability has been investigated during last decades, there is no systematic study on underlying relationships between high-order structural information and network controllability robustness. Existing methods often rely on measures like degree and betweenness \cite{22,23} or focus on pairwise interactions and local structures such as rings and chains \cite{24,25,26}. While insightful, these approaches largely overlook crucial high-order connection information, where multiple nodes might interact collectively or share complex dependencies which are not captured by simple dyadic relationships. This issue is significant because such multi-way interactions, representing functional groups, community structures, or complex influence pathways, are hypothesized to fundamentally affect how a network's controllability responds to perturbations. To address this challenge, the introduction of hypergraph is a promising solution. Unlike traditional graphs, hypergraphs employ hyperedges that can connect any number of nodes, offering a more expressive framework to naturally model these many-body interactions and capture a richer understanding of the network's intricate organization, which we believe is key to more accurately assessing controllability robustness.

Traditionally, the controllability robustness is determined by attack simulation, which is computationally time consuming. Lately, some machine learning based methods for predicting network controllability robustness have been proposed. In \cite{27} and \cite{28}, the convolutional neural network (CNN) based predictor for controllability robustnes (PCR) and a knowledge-based predictor for controllability robustnes (iPCR) are proposed, which achieve high-precision controllability robustness learning when the training and test sets maintain the same distribution. In \cite{CRL-SGNN}, the authors proposed to learn network robustness via Low-Complexity spectral graph neural networks. However, these methods, while innovative, also tend to dismiss the impact of underlying high-order structural information and can be challenging to train due to their high complexity (e.g., iPCR), making them less practical for large-scale networks.

To overcome these limitations and effectively integrate high-order structural insights, this paper introduces a graph attention neural network model based on high-order knowledge (NCR-HoK). We explicitly construct hypergraphs from the input network to capture this vital high-order information. Specifically, we employ methods such as K-hop neighborhood analysis and K-Nearest Neighbors (K-NN) for hypergraph construction due to their ability to reveal distinct aspects of network topology. The motivation and main contributions of our paper are as follows:
\begin{itemize}
    \item The K-hop approach defines hyperedges by encompassing nodes within a specific path distance (K hops). This is motivated by the understanding that a node's influence and its contribution to overall network controllability often extend beyond its immediate connections, and capturing these extended local environments as single hypergraph structures can reveal how local clusters collectively impact robustness.
    \item The K-NN method, typically applied to node embeddings or structural feature vectors, groups nodes that exhibit similar topological roles or characteristics, even if not directly linked. Hyperedges formed from such K-NN groups can uncover latent communities or functionally related node sets whose collective behavior underpins network stability. 
    \item By leveraging these diverse hypergraph construction strategies, we aim to extract multifaceted high-order structural data. This rich high-order knowledge, embedded within the hypergraph, is then processed by our NCR-HoK model. The attention mechanism within this model is particularly well-suited to learn and weigh the significance of these complex multi-node interactions and high-order patterns for the tasks of robustness learning and controllability robustness curve prediction.
     \item Compared to traditional network attack simulation-based methods and cutting-edge machine learning methods, the proposed NCR-HoK aims to consume significantly less time to obtain an accurate controllability robustness measurement for different types of real-world and synthetic networks, with lower overheads in training.
\end{itemize}

In summary, the core contribution of NCR-HoK lies in designing a novel hypergraph-based model, which is the first to integrate K-hop/K-NN hypergraph construction with a dual-channel attention mechanism. Through comprehensive experimental and theoretical analysis, it is demonstrated that NCR-HoK can fully extract high-order knowledge from networks, thereby enabling the accurate prediction of controllable robustness curves for networks.

The remainder of this article is organized as follows. Section II reviews the research of network controllability and controllability robustness against various attacks. Section III provides preliminary and basic knowledge for the proposed method. Section IV describes how HCR-HoK works in detail. In Section V, extensive experimental studies are conducted with comprehensive analysis. Finally, Section VI concludes the whole paper. The supplemental materials are given in Section VII and can be downloaded along with this article.

\section{Related Work}
\subsection{Network Controllability Robustness Predictors}

In the last few years, some machine learning based methods for predicting network controllability robustness have been proposed to address the limitation of traditional simulation-based methods. Among them, PCR\cite{27} is a predictor based on a single CNN. This framework transforms complex network data into grayscale images, typically by using the adjacency matrix as input to the CNN. iPCR\cite{28} is an improved version of PCR, employing a set of CNNs to predict network controllability robustness. This approach leverages a large amount of simulation-generated training data to train the ensemble of CNNs, which are used for both classification and regression tasks. iPCR outperforms the classical single-CNN predictor in terms of prediction accuracy, and it also achieves better performance than traditional spectral metrics and network heterogeneity measures. Recently, several graph neural network(GNN) based methods have also been proposed. \cite{lu2024graph} introduced a graph convolutional network(GCN) model that directly takes network topology and node features as input, rather than converting them into grayscale images. Compared to CNN-based methods, this GCN model requires only 1\% of the parameters, yet achieves better prediction performance across various synthetic networks and six real-world networks. \cite{CRL-SGNN} proposed a controllability robustness learning model based on a spectral graph neural network, named CRL-SGNN. This model first uses graph convolutional layers to generate network representations incorporating topological information and node degree centrality, and then employs a multi-branch prediction module to output robustness metrics.

However, these methods largely overlook critical high-order connectivity information. This issue is significant, as multi-node interactions—representing functional groups, community structures, or complex influence paths—are believed to fundamentally affect how network controllability responds to perturbations.

\subsection{Hypergraph Neural Network}

A hypergraph is a generalization of a graph that introduces hyperedges, which can connect multiple nodes simultaneously, enabling the representation of high-order relationships. HGNNs\cite{+5} are designed to capture such high-order associations by employing hyperedge convolution or attention mechanisms, allowing node representations to incorporate high-order structural information. Compared to GCNs, HGNNs are capable of modeling latent structures in complex multimodal data and have demonstrated superior performance on tasks such as node classification and visual recognition. In recent years, various HGNN models have been developed. HGNN+ \cite{+6}, for instance, is tailored for multimodal or heterogeneous data. It first constructs sets of hyperedges for each modality to represent latent high-order correlations. These modality-specific hypergraphs are then merged into a unified hypergraph via an adaptive fusion strategy, upon which spatial-domain convolution is applied. Besides, some approaches incorporate attention mechanisms to weigh the importance of different hyperedges. For example, the Hypergraph Attention Network\cite{HGAT} introduces attention coefficients between nodes and their associated hyperedges to enhance information filtering. DHHNN \cite{mei2025dhhnn} combines the advantages of hyperbolic geometry, dynamic hypergraphs, and the self-attention mechanism to enhance multimodal data representation learning. \cite{yang2025recent} proposed the Dual-view HGNN, which constructs two sets of hyperedges from the topological structure and node attributes, respectively. It employs both shared and view-specific hypergraph convolutional layers—augmented with attention mechanisms—to deeply model the complex relationships among nodes. In summary, as a high-level form of graphical representation, hypergraphs find numerous applications in the field of graph data mining, as detailed in \textbf{Supplementary Materials VII. A}.

This paper proposes NCR-HoK, which captures high-order semantics from multiple perspectives by jointly leveraging the original feature space and the embedding space. The learnability and dynamic adaptability of the embedding space further enhance the model’s capacity to extract and integrate high-order information.


\section{PRELIMINARIES}

This section first introduces the concept and calculation of controllability robustness for a given complex network, which reflects the network system's ability to maintain a controllable state. Then we discuss the concept of high-order hypergraph structures in complex networks. Finally, the error measurement and evaluation criteria used in this study for learning network controllability robustness are presented.

\subsection{Network Controllability Robustness}
The controllable robustness of complex networks reflects the ability of network systems to maintain or restore their controllability at the lowest cost. For a linear time-invariant network system $\dot{x}=Ax+Bu$, where $A$ and $B$ are the transposed adjacency matrix and input matrix, and $x$ and $u$ are the state vector and input vector. The full rank of the controllability matrix $[B\textit{ AB A}^2B...A^{N-1}B]$ is a necessary and sufficient condition for state controllability, where $N$ represents the dimension of $A$ and the total number of nodes in the network. Based on the minimum input theorem \cite{5}, the minimum number of driver nodes $N_{D}$ for a directed network can be determined by $N_{D}=\max\left\{N-|E*|,1\right\}$, where $|E*|$ is the number of edges in the maximum matching $E*$. For undirected networks, the minimum number of driver nodes $N_{D}$ can be determined by $N_{D}=\max\{N-rank(A),1\}$ according to the framework of exact controllability \cite{+1}.

Then, the calculation of network controllable robustness is as follows:
\begin{equation}
\label{1}
n_D(i)=\frac{N_D(i)}{N-i},i=1,2,...,N-1
\end{equation}
where $N_{D}(i)$ is the number of driver nodes needed to retain the network controllability after removing $i$ nodes, and $N-i$ is the number of remaining nodes after the $i$-th attack. By recording the controllable robustness of each state during a network attack, we can obtain a curve of the network's controllable robustness, which intuitively reflects the changes in the network's robustness throughout the attack process.

\begin{figure*}[ht]
  \centering
  \includegraphics[width=\textwidth]{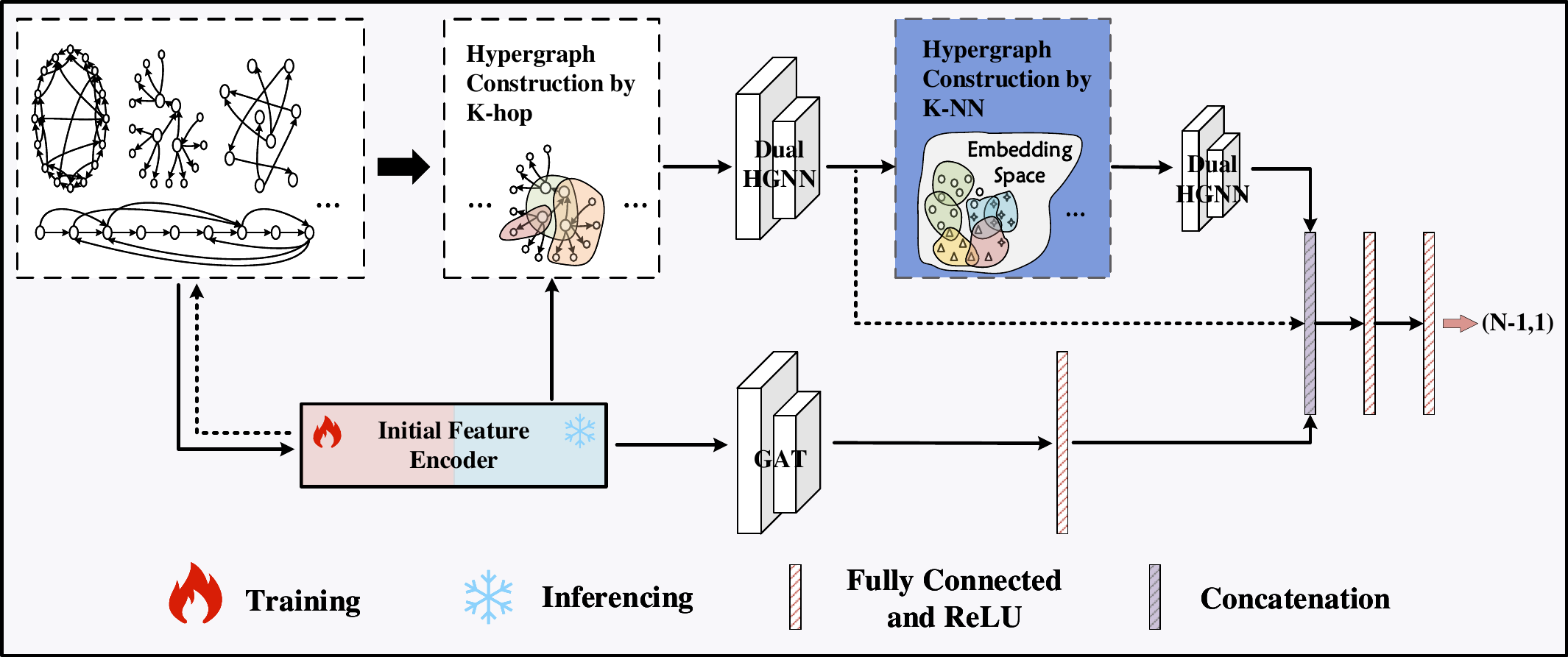}
  \caption{The Overview of the proposed NCR-HoK framework. NCR-HoK can effectively learn three types of network information simultaneously: the explicit structural information in the original graph, the high-order connection information in local neighborhoods and more hidden structures and features in the embedding space.}
  \label{fig1}
\end{figure*}

\subsection{Hypergraph Representation Learning}
A hypergraph is defined as $G_{H}=\{V, E, F\}$, where $V=\{v_{1}, v_{2}, \dots, v_{N}\}$ represents the set of nodes in the graph and $E=\{e_{1}, e_{2}, \dots, e_{M}\}$ represents the set of hyperedges. For any hyperedge, it can contain two or more nodes. It is worth noting that the topology of hyperedge $G_{H}$ can be also represented by the association matrix $A_{Hij}\in R^{N\times M}$, with entries defined as: 
\begin{equation}
\label{2}
A_{H_{ij}}=\begin{cases}1,if~\nu_i\in e_j\\0,if~\nu_i\notin e_j\end{cases}
\end{equation}

In general, each node in the hypergraph can have a $d$-dimensional attribute vector. Therefore, all node attributes can be expressed as $F=[F_{1},F_{2},...,F_{N}]^{T}\in R^{N\times d}$.

Hypergraphs \cite{+2,+3,+4,+5} have attracted more and more attention in recent years. For mining hypergraph structures, \cite{+6} gives four generation methods, namely hyperedge group using pairwise edge, hyperedge group using k-Hop neighbors, hyperedge group using attributes, and hyperedge group using features. The generated hypergraph structures can effectively mine the high-order knowledge of graph structures and provide unique data analysis perspectives for downstream tasks such as recommender systems \cite{+7}, text categorization \cite{+8}, disease diagnosis \cite{+9}, and chip layout wiring \cite{+10}. As for the controlled robustness analysis of complex networks, how to use hypergraphs and analyze the impact of high-order knowledge on network controlled robustness is an interesting research topic.

\subsection{Measurement of Network Controllability Robustness}
This paper employs four types of curves to measure and describe the controllable robustness of networks. For $M$ networks with the node size denoted as $N$, their true robustness profile is expressed as $tv\in R^{M\times(N-1)}$ and the robustness profile predicted by the model is expressed as $pv\in R^{M\times(N-1)}$.  The four types of curves are introduced as follows: 1) The actual controllable robustness curve of the network $TC=mean(tv)\in R^{1\times (N-1)}$, obtained through attack simulation methods combined with the formulas from references \cite{+1} or \cite{5}. 2) The predicted controllable robustness curve $PC=mean(pv)\in R^{1\times (N-1)}$. 3) The error curve between the predicted and actual values $er=mean(|pv-tv|)\in R^{1\times (N-1)}$. 4) The standard deviation curve between the predicted and actual values $\sigma=std(|pv-tv|)\in R^{1\times (N-1)}$.

In addition to analyzing relevant curve graphs, this article also utilizes two evaluation metrics, $\overline{er}$ and $\overline{\sigma}$, where $\overline{er}$ represents the mean accuracy of the model's predictions for controllable robustness curve and $\overline{\sigma}$ represents the average deviation in the accuracy of these predictions. For both $\overline{er}$ and $\overline{\sigma}$, lower values indicate better performance, and the $\overline{er}$ and $\overline{\sigma}$ are calculated as follows:
\begin{equation}
\overline{er}=mean(er)\in R
\end{equation}
\begin{equation}
\overline{\sigma}=mean(\sigma)\in R
\end{equation}

\section{PREDICTOR FOR NETWORK CONTROLLABILITY ROBUSTNESS}
In this section, we construct high-order network structure using hypergraphs and introduce our model (NCR-HoK) that combines Attention Neural Networks to obtain high-order knowledge of graphs for predicting the controllable robustness curve of graphs. The overall framework of our model is shown in Fig.\ref{fig1}. In NCR-HoK, the node feature encoder encodes the in-degree, out-degree, and betweenness centrality of the graph to obtain initial node features. It integrates features of the graph structure using both the Graph Attention (GAT) Network and the Dual Hypergraph Attention (Dual HGNN) Network. Initially, the GAT network embeds the original graph structure to learn the explicit features of the graph. Subsequently, the Dual HGNN network embeds the generated hypergraph structure to learn hidden and high-order features of the graph. Additionally, the K-NN method adaptively generates hypergraphs in the embedding space of the graph. The Dual HGNN network then embeds these generated hypergraph structures to further explore the potential connections between the graph structure and the graph's controllable robustness curve. Detailed information about NCR-HoK is described below.

\subsection{Node Initial Feature Encoder}
In the study of controllable robustness of graphs, structural properties such as node in-degree, out-degree, and betweenness centrality play crucial roles. They provide deep insights into the structure and function of graphs, thereby influencing their performance in responding to external disturbances. Consequently, the node initial feature encoder of NCR-HoK encodes these aspects separately and then concatenates the results to obtain the initial features of the nodes.




The betweenness centrality of a node $v$ is defined as:

\begin{equation}
C_B(v) = \sum_{s \ne v \ne t} \frac{Path_{st}(v)}{Path_{st}}
\end{equation}

where $Path_{st}$ denotes the total number of shortest paths from node $s$ to node $t$, and $Path_{st}(v)$ denotes the number of those paths that pass through node $v$. Currently, the most commonly used method for computing the exact betweenness centrality (BC) is Brandes' algorithm\cite{brandes2001faster}. For an unweighted graph with $n$ nodes and $m$ edges, its time complexity is $\mathcal{O}(nm)$. This implies that for dense graphs, the complexity approaches $\mathcal{O}(n^2)$, making it computationally expensive. To address this issue, we propose training a GAT with a global receptive field as a predictor for betweenness centrality, denoted as $BC_{\text{GAT}}$. The computational complexity of this method is reduced to $\mathcal{O}(H_{ead}h_{hidden}^2m)$, where $H_{ead}$ is the sum of the attention heads of all layers in the $BC_{GAT}$, $h_{hidden}$ is the output dimension size at attention head, and $m$ is the number of edges in the graph. Experimental results demonstrate that the BC values predicted by $BC_{GAT}$ model are sufficiently accurate to allow the NCR-HoK framework to effectively perform network robustness curve prediction tasks. The loss of $BC_{GAT}$ training can be expressed as follows:
\begin{equation}
Loss_{BC_{GAT}}=MSE(BC_{GAT}({v_{i}})-BC_{True}(v_i))
\end{equation}
where $BC_{GAT}({v_{i}})$ denotes the BC of node $v_i$ predicted by the GAT\cite{gat} and $BC_{True}(v_i)$ denotes the ground true BC of node $v_i$.

Specifically, for each node $v_i$, the initial features $F(v_i)$ can be defined as follows:
\begin{equation}
F(v_{i})=Linear(X_{v_{i}}^{out-deg}\left\|X_{v_{i}}^{in-deg}\right\|BC_{GAT}(v_{i}))
\end{equation}
where $X_{v_{i}}^{out-deg}$, $X_{v_{i}}^{in-deg}$ and $BC_{GAT}(v_{i})$ represent the node's in-degree, out-degree, and betweenness centrality inferred by the $BC_{GAT}$ with frozen parameters, respectively. Notably, these three embedding vectors are learned through three independent encoders. Besides, $||$ represent the concatenation operation.

\begin{figure}[!t]
\centering
\includegraphics[width=3.5in]{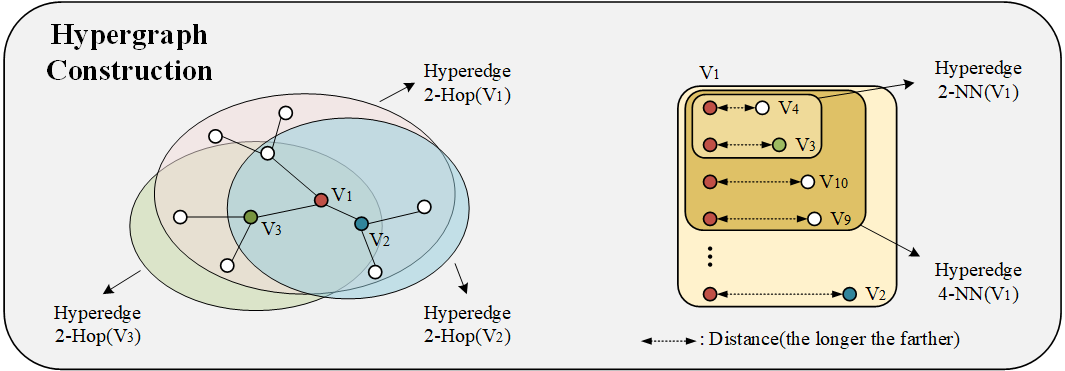}
\caption{The Principle of Hypergraph Generation by K-Hop and K-NN.}
\label{fig2}
\end{figure}

\subsection{Network High-Order Relation Generator}
The core strength of our model NCR-HoK lies in its ability to uncover the latent relationships between high-order knowledge of graphs and network controllability robustness. This is achieved through the integration of hypergraph attention neural networks, where the generation of hypergraphs plays a crucial role. As shown in Fig.\ref{fig2}, we utilized the K-Hop method to generate the initial structure of the hypergraph. Specifically, this method considers any set of nodes in a graph, where the distance between any two nodes does not exceed K hops, as a hyperedge, forming a hypergraph $HG_{K-Hop}(g)$ where $g$ denotes the network index. This approach allows NCR-HoK to capture high-order connectivity patterns within local neighborhoods, thereby gaining a deeper understanding of the intrinsic characteristics of graph structures. The construction of K-Hop hypergraphs helps reveal how nodes in a graph influence each other within a limited number of steps, providing a new perspective for understanding the dynamic changes and controllable robustness of graphs.

Moreover, we utilize the K-NN method to generate a more complex hypergraph structure. By calculating the distances between the embedding representations of nodes within the hypergraph embedding space, we connect each node to its K nearest neighbors to form hyperedges, generating hypergraph $HG_{K-NN}(g)$, where $g$ denotes the network index. This approach not only considers the direct connectivity between nodes but also their similarity in the embedding space, thereby capturing more covert structures and features within the graph. The hypergraphs generated through the K-NN method enable NCR-HoK to adaptively explore and learn the graph's hidden and high-order features, providing robust support for predicting the graph's controllable robustness curve.
\subsection{Graph Attention Neural Network Module}
In order to learn underlying graph structures and improve the prediction of controllable robustness curves of graphs, we employ a dual-channel embedding approach combining GAT Neural Network Module and Dual HGNN Neural Network Module. The GAT Neural Network Module is particularly inclined towards learning explicit structural features present in the original graph.

Let $h^l(v_{i})$ represent the embedding features of node $v_i$ at layer $l$, where $h^0(v_i)=F(v_i)$. Input the adjacency matrix and encoded node features of the network into the Graph Attention Neural Network, where the feature transfer and aggregation between nodes can be represented as:
\begin{equation}
h^{l}(v_{i})=\Phi(\sum_{e_{ij}\neq0}\alpha_{ij}h^{l-1}(v_{j}))
\end{equation}
where $\Phi$ refers to a nonlinear layer and ReLU is used as activation function in our method. $\alpha_{ij}$ represents the attention coefficient between node $v_i$ and its connected nodes $v_j$, which can be calculated using the following formula:
\begin{equation}
\alpha_{ij}=\frac{\exp{({a_{1}}^{T}\cdot u_{j})}}{\Sigma_{e_{ip}\neq0}\exp{({a_{i}}^{T}\cdot u_{p})}}
\end{equation}

\begin{equation}
u=ReLU(W\cdot h^{l-1}(v_{i})+b)
\end{equation}
where $a_{i}^{T}$ is a trainable context vector, $W$ and $b$ are the transformation weights and biases of nodes in the trainable attention-ready state. 

The above describes the working process of a single layer GAT Network. A GAT Neural Network Module consists of two layers of GAT Network. For each graph $G_{g}$, we concatenate its nodes to form the output of the GAT Network Module, serving as the graph embedding results for the graph.

\subsection{Dual Hypergraph Attention Neural Network Module}
By integrating the Dual HGNN Network, NCR-HoK not only focuses on the relationship between individual node and their neighbors, but also pays attention to complex interactions across multiple nodes, thereby providing richer and more in-depth feature representations. For the generated hypergraphs $HG_{K-Hop}(g)$ and $HG_{K-NN}(g)$, we introduced a dual attention message passing module to learn the implicit and high-order features within these hypergraphs. Below, we describe the learning process of a single-layer hypergraph attention architecture for a hypergraph $HG_{K-Hop}(g)$.

\subsubsection{Node Attention on Hyperedge}
First, we use an attention-based message passing scheme to aggregate node information on hyperedges, thereby obtaining the embedding vectors of the hyperedge $h_{E-KHop}^{l}(e_{j})$ generated in the hypergraph. Embedded representations of nodes in the $l$-th layer of the network can be regarded as $h_{N-KHop}^{l}(v_{i})$, where $h_{N-KHop}^{0}(v_{i})=F(v_{i})$. For hyperedge $e_j$ in hypergraph $HG_{K-Hop}(g)$, the learned feature of the layer can be described as:
\begin{equation} \label{eq:eq11}
h_{E-KHop}^{l}(e_{j})=\Phi(\sum_{v_{i}\in e_{j}}\alpha_{E}(e_{j},v_{i})\cdot h_{N-Kop}^{l-1}(v_{i}))
\end{equation}
where $\alpha_{E}(e_j,v_i)$ is the attention coefficient of each node $v_i$ on the hyperedge $e_j$, which can be calculated by the following formula:
\begin{equation} \label{eq:eq12}
\alpha_{E}(e_{j},v_{i})=\frac{\exp{(a_{2}^{T}\cdot s_{N}(e_{j},v_{i}))}}{\Sigma_{v_{i^{\prime}}\in e_{j}}(a_{2}^{T}\cdot s_{N}(e_{j},v_{i^{\prime}}))}
\end{equation}
\begin{equation} \label{eq:eq13}
s_{N}(e_{j},v_{i})=ReLU(W_{N}\cdot h_{N-KHop}^{l-1}(v_{i})+b_{N})
\end{equation}
where $a_{2}^T$ is the trainable context vector, $W_N$ and $b_N$ are the transformation weights and biases with respect to the attention-ready state of the node, respectively.

\subsubsection{Hyperedge Attention on Node}
In the hypergraph, to update the embedding of node $v_i$ at layer $l$, we use the following attention mechanism to aggregate information through hyperedges $e_j$ connected to node $v_i$, which can be expressed as:
\begin{equation} \label{eq:eq14}
h_{N-KHop}^{l}(v_{i})=\Phi(\sum_{e_{j}\supset v_{i}}\alpha_{N}(e_{j},v_{i})\cdot h_{E-KHop}^{l-1}(e_{i}))
\end{equation}
where $\alpha_{N}(e_j,v_i)$ is the attention coefficient about each hyperedge $e_j$ passing through the node $v_i$, which can be calculated by the following formula:
\begin{equation} \label{eq:eq15}
\alpha_N(e_j,v_i)=\frac{\exp(a_3^T\cdot s_E(e_j,v_i))}{\sum_{e_{j^{\prime}}\supset v_i}(a_3^T\cdot s_E(e_{j^{\prime}},v_i))}
\end{equation}
\begin{equation} \label{eq:eq16}
s_E(e_j,v_i)=ReLU(W_E\cdot h_{E-KHop}^{l-1}(e_j)+b_E)
\end{equation}
where $a_{3}^T$ is the trainable context vector, $W_E$ and $b_E$ are the transformation weights and biases with respect to the attention-ready state of the node, respectively.

The process described above involves using a single-layer Dual HGNN Network for learning the embedding of hypergraph $HG_{K-Hop}(g)$. The Dual HGNN Neural Network Module includes two layers of Dual HGNN Network. We concatenate all the embedded nodes $h_{N-KHop}^l(v)$ in the hypergraph to serve as the output of the Dual HGNN Neural Network Module, which is used as the graph embedding results of the hypergraph $HG_{K-Hop}(g)$

\subsubsection{The Dual Hypergraph Attention Neural Network Module on $HG_{K-NN}(g)$}
Similarly, for the hypergraphs $HG_{K-NN}(g)$ generated in the embedding space, we also use a Dual HGNN Neural Network Module to embed them. The feature of nodes in hypergraph $HG_{K-NN}(g)$ is represented by $h_{N-KNN}^l(v)$, where $h_{N-KNN}^0(v)=h_{N-KHop}^l(v)$.

\subsection{Prediction of Network Controllable Robustness Curve}
By using the GAT Neural Network Module and Dual HGNN Neural Network Module to learn embeddings from the original graph structure and hypergraphs $HG_{K-Hop}$ and $HG_{K-NN}$, our model is supposed to capture explicit structural information in the original graph, high-order connectivity within local neighborhoods, and more hidden features in the embedding space. Before predicting controllable robustness curves using the input MLP module, the embedding features of the original graph $G(g)$ are formed by concatenating the features of each node from the three parts of embeddings mentioned above.
\begin{equation}
G(g)_{v_i}=h^l(v)\|h_{N-KHop}^l(v_i)\|h_{N-KNN}^l(v)
\end{equation}

Then, using the MLP module and transforming the original graph $G(g)$ features into the final network controllable robustness curve, denoted as PC(G(g)), we get the controlled robustness curve prediction results made by NCR-HoK for the network $G(g)$.

Furthermore, we employ the SmoothL1Loss function to optimize our model, formulated as:
\begin{equation}
loss(x,y)=\frac{1}{N}\sum_{i=1}^{N}\begin{cases}0.5\cdot(PC_i-TC_i)^2,if\left|PC_i-TC_i\right|<1\\\left|PC_i-TC_i\right|-0.5,otherwise\end{cases}
\label{loss}
\end{equation}
where $N$ is the number of nodes of the network, $PC\in R^{(N-1)\times1}$ is the controlled robustness profile of the network obtained from the model prediction, and $TC\in R^{(N-1)\times1}$ is the true controlled robustness profile of the network.

\section{EXPERIMENTAL STUDIES}
In this section, we evaluate the effectiveness and superiority of NCR-HoK through extensive experimental assessments. Initially, we introduce the datasets used in the experiments and the relevant parameter settings for our method as well as baseline methods. Subsequently, NCR-HoK is employed to predict the controlled robustness curves across various types of networks and corresponding experimental analyses are provided. Besides, ablation studies and parameter analysis are conducted to explore potential factors or modules that might influence the performance of NCR-HoK, shedding light on the potential improvement and application of our method. 


\begin{figure*}[ht]
  \centering
  \includegraphics[width=5.5in]{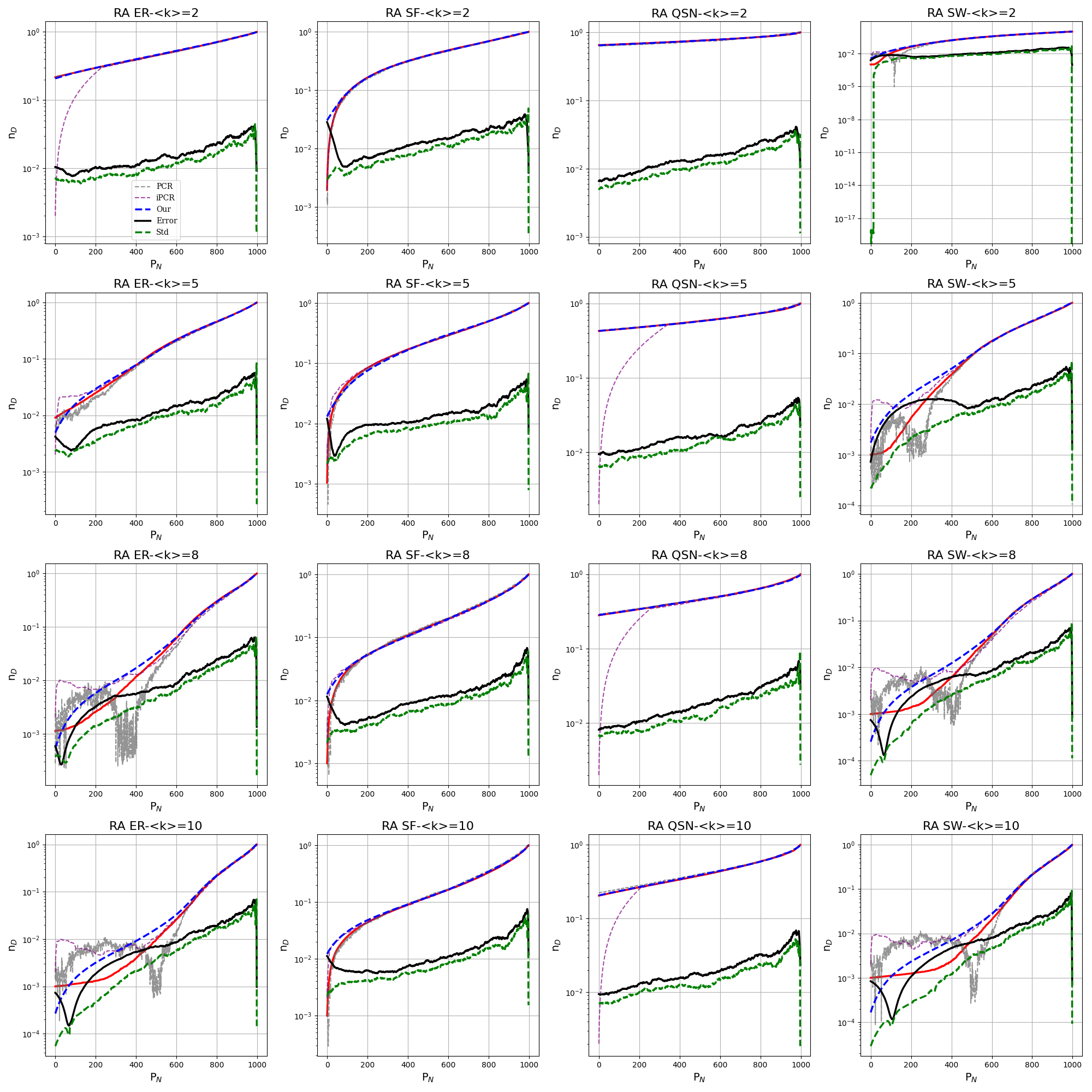}
  \caption{The predictions of the controllable robustness curves for different network types under the RAs condition by NCR-HoK, PCR, and iPCR models. $P_N$ represents the number of nodes having been removed from the network; and $n_D$ is calculated by \ref{1}. RA (x)-$\textless{}k\textgreater{}$=(y) represents the true values of the controllable robustness curve under the RA condition for a network of type (x) with an average degree of (y), as well as the prediction results of each model.}
  \label{fig3}
\end{figure*}

\subsection{Experimental Settings}
\subsubsection{Datasets}
We evaluate the performance of proposed NCR-HoK on four typical directed and unweighted synthetic network models, including the Erdös-Rényi (ER) \cite{+11}, generic scale-free (SF) \cite{22, +13, +14}, q-snapback \cite{+13}, Newman–Watts small-world(NW) \cite{24}, Barabási–Albert (BA) scale-free networks \cite{+16}.

Specifically, the edges in the ER network are added completely at random, where the direction of each edge is evenly randomly assigned. The edges of the SF network are generated according to their weights defined by:
\begin{equation}
w_i = (i+\theta)^{-\beta}, i=1,2,\dots, N
\end{equation}
where $\beta\in \left[ {0,1} \right)$ and $\theta \ll N$. Two nodes $v_j$ and $v_i(i\neq j, i,j = 1,2,\dots, N)$ are randomly picked with a probability proportional to the weights $w_i$ and $w_j$, respectively. Then, an edge $e_{ij}$ from $i$ to $j$ is added (if the two nodes are already connected, do nothing). The resulting SF network has a power-law distribution $k^{-\gamma}$, where $k$ is the degree variable, with constant $\gamma = 1 + (1/\beta)$, which is independent of $\theta$. In this article, $\beta$ is set to 0.999 such that the power-law distribution has a reasonable scaling exponent $\gamma = 2.001$. The QSN has one layer $r_q$ generated with a backbone chain and multiple snapback edges created for each node $i=r_q +1, r_q +2,\dots, N$, which connects backward to the previously appeared nodes $i-l\times r_q(l=1,2,\dots,\lfloor {i/r_q} \rfloor)$, with the same probability $q \in \left[ 0,1 \right]$. The SW network is initiated by a directed ring with $K=2$ nearest-neighbors connected; that is, node $v_i$ is connected to nodes $v_{i-1}$, $v_{i-2}$ and $v_{i+2}$ via edges $e_{i-1,i}, e_{i,i+1}, e_{i-2,i}$ and $e_{i,i+2}$. Then, adding or removing edges, until the predefined number of edges is reached.


Based on the adjacency matrix of each network topology, the initial hypergraph structure can be generated using the K-Hop method, with the corresponding adjacency matrix represented as follows:
\begin{equation}
A_{ij}=\begin{cases}0,&if~v_i \notin e_j \\1,&if~v_i \in e_j\end{cases}
\end{equation}

Considering the limited length of the paper, after considering the simulation methodology of Randomized Attacks (RA), other types of attacks such as Target Degree Based Attacks (TDA) as well as Target Betweenness Based Attacks (TBA) can be investigated in the same manner, detailed results are presented in \textbf{Supplementary Materials VII. F}. All experiments are carried out on a computer with a 64-bit operating system, installed with two Intel i7-11700K (3.6GHz) and four NVIDIA Geforce RTX 3090. Baseline models are implemented using the default settings described in their original papers.

\begin{table*}[ht]
\centering
\caption{The average errors $\overline{er}$ and average standard deviations $\overline{\sigma}$ of the controllable robustness curve predicting for various types of networks under the RA condition are compared among the NCR-HoK, PCR\cite{27}, iPCR\cite{28} models and CRL-SGNN\cite{CRL-SGNN} , along with a comprehensive ranking.}
\label{tab:table 1}
\begin{tabular}{|ccc|cccc|c|}
\hline
\multicolumn{3}{|c|}{}    & \textbf{\textless{}k\textgreater{}=2}  & \textbf{\textless{}k\textgreater{}=5}  & \textbf{\textless{}k\textgreater{}=8}  & \textbf{\textless{}k\textgreater{}=10} & \textbf{Average}    \\ \hline
\multicolumn{1}{|c|}{}   & \multicolumn{1}{l|}{}    & $\overline{er}$    & 0.020    & 0.016  & 0.014    & 0.014    & 0.016(\#2)    \\
\multicolumn{1}{|c|}{}   & \multicolumn{1}{c|}{\multirow{-2}{*}{\textbf{PCR}}}  & \cellcolor[HTML]{D9D9D9}$\overline{\sigma}$ & \cellcolor[HTML]{D9D9D9}0.015   & \cellcolor[HTML]{D9D9D9}0.019          & \cellcolor[HTML]{D9D9D9}0.020          & \cellcolor[HTML]{D9D9D9}0.022          & \cellcolor[HTML]{D9D9D9}0.019(\#2)          \\ \cline{2-2}

\multicolumn{1}{|c|}{}                               & \multicolumn{1}{l|}{}                                & $\overline{er}$                         & 0.043                                  & 0.015                                  & 0.017                                  & 0.013                                  & 0.022(\#3)                                  \\
\multicolumn{1}{|c|}{}                               & \multicolumn{1}{c|}{\multirow{-2}{*}{\textbf{iPCR}}} & \cellcolor[HTML]{D9D9D9}$\overline{\sigma}$ & \cellcolor[HTML]{D9D9D9}0.054          & \cellcolor[HTML]{D9D9D9}0.018          & \cellcolor[HTML]{D9D9D9}0.022          & \cellcolor[HTML]{D9D9D9}0.020          & \cellcolor[HTML]{D9D9D9}0.029(\#4)          \\ \cline{2-2}

\multicolumn{1}{|c|}{}                               & \multicolumn{1}{l|}{}                                & $\overline{er}$                         & 0.029                                  & 0.034                                  & 0.020                                  & 0.014                                  & 0.024(\#4)                                  \\
\multicolumn{1}{|c|}{}                               & \multicolumn{1}{c|}{\multirow{-2}{*}{\textbf{CRL-SGNN}}} & \cellcolor[HTML]{D9D9D9}$\overline{\sigma}$ & \cellcolor[HTML]{D9D9D9}0.016          & \cellcolor[HTML]{D9D9D9}0.043          & \cellcolor[HTML]{D9D9D9}0.021          & \cellcolor[HTML]{D9D9D9}0.022          & \cellcolor[HTML]{D9D9D9}0.026(\#3)          \\ \cline{2-2}

\multicolumn{1}{|c|}{}                               & \multicolumn{1}{l|}{}                                & $\overline{er}$                         & \textbf{0.016}                         & \textbf{0.015}                         & \textbf{0.012}                         & \textbf{0.010}                         & \textbf{0.013(\#1)}                         \\
\multicolumn{1}{|c|}{\multirow{-8}{*}{\textbf{ER}}}  & \multicolumn{1}{c|}{\multirow{-2}{*}{\textbf{Our}}}  & \cellcolor[HTML]{D9D9D9}$\overline{\sigma}$ & \cellcolor[HTML]{D9D9D9}\textbf{0.012} & \cellcolor[HTML]{D9D9D9}\textbf{0.011} & \cellcolor[HTML]{D9D9D9}\textbf{0.009} & \cellcolor[HTML]{D9D9D9}\textbf{0.008} & \cellcolor[HTML]{D9D9D9}\textbf{0.010(\#1)} \\ \hline

\multicolumn{1}{|l|}{}                               & \multicolumn{1}{l|}{}                                & $\overline{er}$                         & 0.017                                  & 0.020                                  & \textbf{0.021}                         & 0.024                                  & 0.021(\#2)                                  \\
\multicolumn{1}{|l|}{}                               & \multicolumn{1}{c|}{\multirow{-2}{*}{\textbf{PCR}}}  & \cellcolor[HTML]{D9D9D9}$\overline{\sigma}$ & \cellcolor[HTML]{D9D9D9}0.014          & \cellcolor[HTML]{D9D9D9}0.018          & \cellcolor[HTML]{D9D9D9}0.020          & \cellcolor[HTML]{D9D9D9}0.022          & \cellcolor[HTML]{D9D9D9}0.019(\#2)          \\ \cline{2-2}
\multicolumn{1}{|l|}{}                               & \multicolumn{1}{l|}{}                                & $\overline{er}$                         & 0.017                                  & 0.100                                  & 0.061                                  & 0.042                                  & 0.055(\#4)                                  \\
\multicolumn{1}{|l|}{}                               & \multicolumn{1}{c|}{\multirow{-2}{*}{\textbf{iPCR}}} & \cellcolor[HTML]{D9D9D9}$\overline{\sigma}$ & \cellcolor[HTML]{D9D9D9}0.015          & \cellcolor[HTML]{D9D9D9}0.012          & \cellcolor[HTML]{D9D9D9}0.073          & \cellcolor[HTML]{D9D9D9}0.050          & \cellcolor[HTML]{D9D9D9}0.038(\#4)          \\ \cline{2-2}
\multicolumn{1}{|l|}{}                               & \multicolumn{1}{l|}{}                                & $\overline{er}$                         & 0.023                         & 0.027                         & 0.022                         & \textbf{0.014}                         & 0.022(\#3)                         \\

\multicolumn{1}{|l|}{}                               & \multicolumn{1}{c|}{\multirow{-2}{*}{\textbf{CRL-SGNN}}} & \cellcolor[HTML]{D9D9D9}$\overline{\sigma}$ & \cellcolor[HTML]{D9D9D9}0.028          & \cellcolor[HTML]{D9D9D9}0.034         & \cellcolor[HTML]{D9D9D9}0.018          & \cellcolor[HTML]{D9D9D9}0.020          & \cellcolor[HTML]{D9D9D9}0.025(\#3)          \\ \cline{2-2}
\multicolumn{1}{|l|}{}                               & \multicolumn{1}{l|}{}                                & $\overline{er}$                         & \textbf{0.016}                         & \textbf{0.019}                         & \textbf{0.021}                         & 0.021                        & \textbf{0.019(\#1)}                         \\

\multicolumn{1}{|c|}{\multirow{-8}{*}{\textbf{SF}}}  & \multicolumn{1}{c|}{\multirow{-2}{*}{\textbf{Our}}}  & \cellcolor[HTML]{D9D9D9}$\overline{\sigma}$ & \cellcolor[HTML]{D9D9D9}\textbf{0.012} & \cellcolor[HTML]{D9D9D9}\textbf{0.015} & \cellcolor[HTML]{D9D9D9}\textbf{0.015} & \cellcolor[HTML]{D9D9D9}\textbf{0.016} & \cellcolor[HTML]{D9D9D9}\textbf{0.015(\#1)} \\ \hline
\multicolumn{1}{|l|}{}                               & \multicolumn{1}{l|}{}                                & $\overline{er}$                         & \textbf{0.015}                         & 0.016                                  & \textbf{0.012}                         & \textbf{0.011}                         & 0.014(\#2)                                  \\
\multicolumn{1}{|l|}{}                               & \multicolumn{1}{c|}{\multirow{-2}{*}{\textbf{PCR}}}  & \cellcolor[HTML]{D9D9D9}$\overline{\sigma}$ & \cellcolor[HTML]{D9D9D9}0.015          & \cellcolor[HTML]{D9D9D9}0.016          & \cellcolor[HTML]{D9D9D9}0.016          & \cellcolor[HTML]{D9D9D9}0.017          & \cellcolor[HTML]{D9D9D9}0.016(\#2)          \\ \cline{2-2}
\multicolumn{1}{|l|}{}                               & \multicolumn{1}{l|}{}                                & $\overline{er}$                         & \textbf{0.015}                         & \textbf{0.013}                         & 0.013                                  & 0.011                                  & \textbf{0.013(\#1)}                         \\
\multicolumn{1}{|l|}{}                               & \multicolumn{1}{c|}{\multirow{-2}{*}{\textbf{iPCR}}} & \cellcolor[HTML]{D9D9D9}$\overline{\sigma}$ & \cellcolor[HTML]{D9D9D9}0.015          & \cellcolor[HTML]{D9D9D9}0.016          & \cellcolor[HTML]{D9D9D9}0.016          & \cellcolor[HTML]{D9D9D9}0.016          & \cellcolor[HTML]{D9D9D9}0.016(\#2)          \\ \cline{2-2}
\multicolumn{1}{|l|}{}                               & \multicolumn{1}{l|}{}                                & $\overline{er}$                         & 0.024                         & 0.043                                  & 0.023                                 & 0.045                                 & 0.034(\#4)                                  \\

\multicolumn{1}{|l|}{}                               & \multicolumn{1}{c|}{\multirow{-2}{*}{\textbf{CRL-SGNN}}} & \cellcolor[HTML]{D9D9D9}$\overline{\sigma}$ & \cellcolor[HTML]{D9D9D9}0.025          & \cellcolor[HTML]{D9D9D9}0.050         & \cellcolor[HTML]{D9D9D9}0.029          & \cellcolor[HTML]{D9D9D9}0.047          & \cellcolor[HTML]{D9D9D9}0.038(\#4)          \\ \cline{2-2}
\multicolumn{1}{|l|}{}                               & \multicolumn{1}{l|}{}                                & $\overline{er}$                         & \textbf{0.015}                         & 0.015                                  & 0.013                                  & 0.012                                  & 0.014(\#2)                                  \\

\multicolumn{1}{|c|}{\multirow{-8}{*}{\textbf{QSN}}} & \multicolumn{1}{c|}{\multirow{-2}{*}{\textbf{Our}}}  & \cellcolor[HTML]{D9D9D9}$\overline{\sigma}$ & \cellcolor[HTML]{D9D9D9}\textbf{0.011} & \cellcolor[HTML]{D9D9D9}\textbf{0.011} & \cellcolor[HTML]{D9D9D9}\textbf{0.009} & \cellcolor[HTML]{D9D9D9}\textbf{0.008} & \cellcolor[HTML]{D9D9D9}\textbf{0.010(\#1)} \\ \hline
\multicolumn{1}{|l|}{}                               & \multicolumn{1}{l|}{}                                & $\overline{er}$                         & 0.013                                  & \textbf{0.014}                         & 0.014                                  & 0.014                                  & 0.014(\#2)                                  \\
\multicolumn{1}{|l|}{}                               & \multicolumn{1}{c|}{\multirow{-2}{*}{\textbf{PCR}}}  & \cellcolor[HTML]{D9D9D9}$\overline{\sigma}$ & \cellcolor[HTML]{D9D9D9}0.014          & \cellcolor[HTML]{D9D9D9}0.018          & \cellcolor[HTML]{D9D9D9}0.021          & \cellcolor[HTML]{D9D9D9}0.023          & \cellcolor[HTML]{D9D9D9}0.019(\#2)          \\ \cline{2-2}
\multicolumn{1}{|l|}{}                               & \multicolumn{1}{l|}{}                                & $\overline{er}$                         & 0.014                                  & \textbf{0.014}                         & 0.017                                  & 0.012                                  & 0.014(\#2)                                  \\
\multicolumn{1}{|l|}{}                               & \multicolumn{1}{c|}{\multirow{-2}{*}{\textbf{iPCR}}} & \cellcolor[HTML]{D9D9D9}$\overline{\sigma}$ & \cellcolor[HTML]{D9D9D9}0.014          & \cellcolor[HTML]{D9D9D9}0.016          & \cellcolor[HTML]{D9D9D9}0.022          & \cellcolor[HTML]{D9D9D9}0.017          & \cellcolor[HTML]{D9D9D9}0.017(\#2)          \\ \cline{2-2}
\multicolumn{1}{|l|}{}                               & \multicolumn{1}{l|}{}                                & $\overline{er}$                         & 0.029                         & 0.040                        & 0.039                         & 0.030                        & 0.035(\#4)                        \\

\multicolumn{1}{|l|}{}                               & \multicolumn{1}{c|}{\multirow{-2}{*}{\textbf{CRL-SGNN}}} & \cellcolor[HTML]{D9D9D9}$\overline{\sigma}$ & \cellcolor[HTML]{D9D9D9}0.032           & \cellcolor[HTML]{D9D9D9}0.042          & \cellcolor[HTML]{D9D9D9}0.044     & \cellcolor[HTML]{D9D9D9}0.032         & \cellcolor[HTML]{D9D9D9}0.038(\#4)          \\ \cline{2-2}
\multicolumn{1}{|l|}{}                               & \multicolumn{1}{l|}{}                                & $\overline{er}$                         & \textbf{0.012}                         & \textbf{0.014}                         & \textbf{0.012}                         & \textbf{0.010}                         & \textbf{0.012(\#1)}                         \\

\multicolumn{1}{|c|}{\multirow{-8}{*}{\textbf{SW}}}  & \multicolumn{1}{c|}{\multirow{-2}{*}{\textbf{Our}}}  & \cellcolor[HTML]{D9D9D9}$\overline{\sigma}$ & \cellcolor[HTML]{D9D9D9}\textbf{0.009} & \cellcolor[HTML]{D9D9D9}\textbf{0.009} & \cellcolor[HTML]{D9D9D9}\textbf{0.008} & \cellcolor[HTML]{D9D9D9}\textbf{0.007} & \cellcolor[HTML]{D9D9D9}\textbf{0.008(\#1)} \\ \hline
\end{tabular}
\end{table*}

\subsubsection{Reproducibility} 
During the training process, the epoch is set to 10 based on experience, and the structure that performed best on the validation data is chosen as the optimal model structure. We use the Adam optimizer to optimize formula \ref{loss} , with a decay rate of 0.5, a learning rate of 0.001, an $l2$ regularization coefficient of 1e-6, and a gradient clipping value of 0.8. Additionally, the hypergraph generation part utilizes both the 3-Hop and 10-NN methods to generate the corresponding hypergraph structures. The selected K values (3-Hop and 10-NN) strike a good balance between capturing comprehensive higher-order knowledge and maintaining computational efficiency. Detailed analysis can be found in \textbf{Supplementary Materials VII. E}. During the pretraining of $BC_{GAT}$, we generated 4000 training samples for each of the four graph types: ER, SF, QSN and SW. For each graph, the number of nodes was randomly selected from the range [800, 1600], and the average node degree $\textless{}k\textgreater{}$ was randomly chosen from 2, 5, 8, and 10. The detailed architecture of the $BC_{GAT}$ model is provided in table \ref{tab:table S0}. 




\subsection{Network Controllability Robustness Learning}
\subsubsection{Controllability Robustness Curve Prediction}
In this section, we demonstrate the prediction of controllable robustness curves under RA conditions for NCR-HoK across different network topologies. Specifically, we address four types of network topologies: ER, SF, QSN, and SW. For each topology, we randomly generate 800 networks with average out-degrees of $\textless{}k\textgreater{}=2,5,8,10$, resulting in a variety of network styles and a total of 12800 network datasets. These datasets are shuffled to serve as training samples for NCR-HoK. Additionally, for each type of network topology, each individually generates 100 network data as a test sample, totaling 1600 networks. The results obtained can be seen in Fig.\ref{fig3} and table \ref{tab:table 1}

\begin{figure*}[!h]
  \centering
  \includegraphics[width=5.5in]{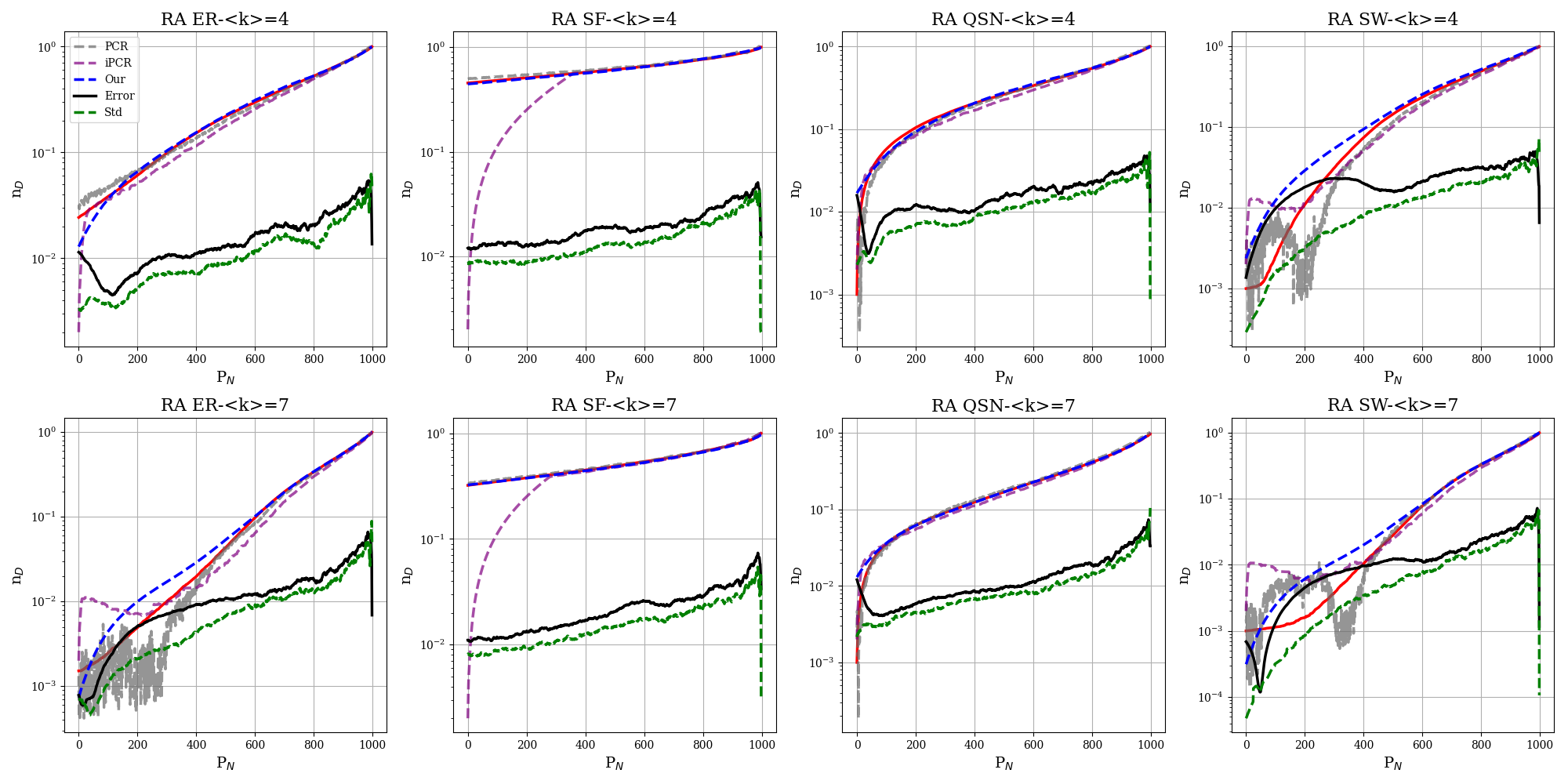}
  \caption{The predictions of the controllable robustness curves for different types of network under the RA condition by NCR-HoK, PCR\cite{27}, iPCR\cite{28} and CRL-SGNN\cite{CRL-SGNN}  models. $P_N$ represents the number of nodes having been removed from the network; and $n_D$ is calculated by \ref{1}. RA (x)-$<k>$=(y) represents the true values of the controllable robustness curve under the RA condition for a network of type (x) with an average degree of (y), as well as the prediction results of each model.}
  \label{fig4}
\end{figure*}

{
\setlength{\tabcolsep}{3.5pt}
\begin{table}[!t]
\centering
\caption{The average errors and standard deviations of the controllable robustness curve predicting for various types of networks with different average degrees under the RA condition are compared among the NCR-HoK, PCR\cite{27}, iPCR\cite{28} and CRL-SGNN\cite{CRL-SGNN} models, along with a comprehensive ranking.}
\label{tab:table 2}
\begin{tabular}{|ccc|cccc|c|}
\hline
\multicolumn{3}{|c|}{}                                                                                                                                           & \textbf{ER}                            & \textbf{SF}                            & \textbf{QSN}                           & \textbf{SW}                            & \textbf{Average}                            \\ \hline
\multicolumn{1}{|c|}{}                                                        & \multicolumn{1}{l|}{}                                & $\overline{er}$                         & 0.019                                  & 0.030                                  & 0.018                                  & 0.021                                  & 0.022(\#2)                                  \\
\multicolumn{1}{|c|}{}                                                        & \multicolumn{1}{c|}{\multirow{-2}{*}{\textbf{PCR}}}  & \cellcolor[HTML]{D9D9D9}$\overline{\sigma}$ & \cellcolor[HTML]{DBDBDB}0.017          & \cellcolor[HTML]{DBDBDB}0.018          & \cellcolor[HTML]{DBDBDB}0.016          & \cellcolor[HTML]{DBDBDB}0.018          & \cellcolor[HTML]{DBDBDB}0.017(\#2)          \\ \cline{2-2}
\multicolumn{1}{|c|}{}                                                        & \multicolumn{1}{l|}{}                                & $\overline{er}$                         & 0.030                                  & 0.109                                  & 0.029                                  & 0.030                                  & 0.050(\#4)                                  \\
\multicolumn{1}{|c|}{}                                                        & \multicolumn{1}{c|}{\multirow{-2}{*}{\textbf{iPCR}}} & \cellcolor[HTML]{D9D9D9}$\overline{\sigma}$ & \cellcolor[HTML]{DBDBDB}0.021          & \cellcolor[HTML]{DBDBDB}0.140          & \cellcolor[HTML]{DBDBDB}0.018          & \cellcolor[HTML]{DBDBDB}0.024          & \cellcolor[HTML]{DBDBDB}0.051(\#4)          \\ \cline{2-2}

\multicolumn{1}{|c|}{}                                                        & \multicolumn{1}{l|}{}                                & $\overline{er}$                         & 0.052                                 & 0.044                                  & 0.035                                 & 0.035                                & 0.042(\#3)                                  \\
\multicolumn{1}{|c|}{}                                                        & \multicolumn{1}{c|}{\multirow{-2}{*}{\textbf{CRL-SGNN}}} & \cellcolor[HTML]{D9D9D9}$\overline{\sigma}$ & \cellcolor[HTML]{DBDBDB}0.057          & \cellcolor[HTML]{DBDBDB}0.045          & \cellcolor[HTML]{DBDBDB}0.040          & \cellcolor[HTML]{DBDBDB}0.041          & \cellcolor[HTML]{DBDBDB}0.028(\#3)          \\ \cline{2-2}

\multicolumn{1}{|c|}{}                                                        & \multicolumn{1}{l|}{}                                & $\overline{er}$                         & \textbf{0.016}                         & \textbf{0.020}                         & \textbf{0.017}                         & \textbf{0.022}                         & \textbf{0.019(\#1)}                         \\
\multicolumn{1}{|c|}{\multirow{-8}{*}{\textbf{\textless{}k\textgreater{}=4}}} & \multicolumn{1}{c|}{\multirow{-2}{*}{\textbf{Our}}}  & \cellcolor[HTML]{D9D9D9}$\overline{\sigma}$ & \cellcolor[HTML]{DBDBDB}\textbf{0.012} & \cellcolor[HTML]{DBDBDB}\textbf{0.015} & \cellcolor[HTML]{DBDBDB}\textbf{0.012} & \cellcolor[HTML]{DBDBDB}\textbf{0.012} & \cellcolor[HTML]{DBDBDB}\textbf{0.013(\#1)} \\ \hline
\multicolumn{1}{|l|}{}                                                        & \multicolumn{1}{l|}{}                                & $\overline{er}$                         & 0.013                                  & 0.024                                  & 0.016                                  & \textbf{0.013}                         & 0.017(\#2)                                  \\
\multicolumn{1}{|l|}{}                                                        & \multicolumn{1}{c|}{\multirow{-2}{*}{\textbf{PCR}}}  & \cellcolor[HTML]{D9D9D9}$\overline{\sigma}$ & \cellcolor[HTML]{DBDBDB}0.018          & \cellcolor[HTML]{DBDBDB}0.020          & \cellcolor[HTML]{DBDBDB}0.021          & \cellcolor[HTML]{DBDBDB}0.020          & \cellcolor[HTML]{DBDBDB}0.020(\#2)          \\ \cline{2-2}
\multicolumn{1}{|l|}{}                                                        & \multicolumn{1}{l|}{}                                & $\overline{er}$                         & 0.023                                  & 0.073                                  & 0.018                                  & 0.018                                  & 0.033(\#4)                                  \\
\multicolumn{1}{|l|}{}                                                        & \multicolumn{1}{c|}{\multirow{-2}{*}{\textbf{iPCR}}} & \cellcolor[HTML]{D9D9D9}$\overline{\sigma}$ & \cellcolor[HTML]{DBDBDB}0.024          & \cellcolor[HTML]{DBDBDB}0.087          & \cellcolor[HTML]{DBDBDB}0.018          & \cellcolor[HTML]{DBDBDB}0.022          & \cellcolor[HTML]{DBDBDB}0.038(\#4)          \\ \cline{2-2}

\multicolumn{1}{|l|}{}                                                        & \multicolumn{1}{l|}{}                                & $\overline{er}$                         & 0.039                                 & 0.030                                  & 0.023                                  & 0.026                                  & 0.030(\#3)                                  \\
\multicolumn{1}{|l|}{}                                                        & \multicolumn{1}{c|}{\multirow{-2}{*}{\textbf{CRL-SGNN}}} & \cellcolor[HTML]{D9D9D9}$\overline{\sigma}$ & \cellcolor[HTML]{DBDBDB}0.025          & \cellcolor[HTML]{DBDBDB}0.030          & \cellcolor[HTML]{DBDBDB}0.026          & \cellcolor[HTML]{DBDBDB}0.030          & \cellcolor[HTML]{DBDBDB}0.028(\#3)          \\ \cline{2-2}

\multicolumn{1}{|l|}{}                                                        & \multicolumn{1}{l|}{}                                & $\overline{er}$                         & \textbf{0.013}                         & \textbf{0.022}                         & \textbf{0.014}                         & 0.014                                  & \textbf{0.016(\#1)}                         \\
\multicolumn{1}{|c|}{\multirow{-8}{*}{\textbf{\textless{}k\textgreater{}=7}}} & \multicolumn{1}{c|}{\multirow{-2}{*}{\textbf{Our}}}  & \cellcolor[HTML]{D9D9D9}$\overline{\sigma}$ & \cellcolor[HTML]{DBDBDB}\textbf{0.009} & \cellcolor[HTML]{DBDBDB}\textbf{0.016} & \cellcolor[HTML]{DBDBDB}\textbf{0.010} & \cellcolor[HTML]{DBDBDB}\textbf{0.009} & \cellcolor[HTML]{DBDBDB}\textbf{0.011(\#1)} \\ \hline
\end{tabular}
\end{table}
}

Fig.\ref{fig3} displays the predicted results of the controllable robustness curves for different network topologies under RA conditions by NCR-HoK. Networks with the same average degree are aligned in a row, and each column represents the same network topology, with average degrees increasing from top to bottom. The red curve represents the average controllable robustness curve obtained through attack simulations. The blue curve (Our) shows the average controllable robustness curve predicted by NCR-HoK for different network topologies. The black curve (Error) indicates the error between predicted and actual values. The blue curve (Std) represents the standard deviation between predicted and actual values. The gray curve (PCR) shows the average controllable robustness curve predicted by the PCR model for different network topologies, and the purple curve (iPCR) represents the average controllable robustness curve predicted by the iPCR model for different network topologies.

It can be seen from Fig.\ref{fig3} that different network topologies exhibit varying trends in controllability robustness curves, and the effectiveness of different models in predicting these curves also varies. Specifically, NCR-HoK accurately predicts the controllability robustness curves for SF and QSN type networks, as evidenced by the blue prediction curve nearly perfectly aligning with the actual values. For ER and SW networks, NCR-HoK still performs well in predictions at lower average degrees. However, as the average degree increases, the predictive accuracy diminishes. This is due to the higher randomness in the topological connections of ER and SW networks, which lessens the impact of topological features on the network's controllability robustness curves. Compared to the PCR and iPCR results, NCR-HoK’s predictions are smoother and more stable, as illustrated by the blue curve.

Table \ref{tab:table 1} shows the comparison results based on average errors $\overline{er}$, standard deviations $\overline{\sigma}$, and overall rankings in the robustness learning of various types of networks under the condition of RA for NCR-HoK, PCR, iPCR and CRL-SGNN. We use “bleak bold” font to indicate the minimum prediction error value and the minimum prediction standard deviation value. It is shown from table \ref{tab:table 1} that aside from the QSN network, our model’s prediction error and prediction standard deviation are the smallest on other graphs. This indicates that NCR-HoK 's network controllability robustness curve predictions are more precise and stable. Consequently, this leads to a more stable and smoother performance of the blue prediction curve in Fig.\ref{fig3}. For the QSN network, we consider that its specific snapback structure may favor the ensemble CNN approach of iPCR.

\begin{figure*}[!t]
  \centering
  \includegraphics[width=5.5in]{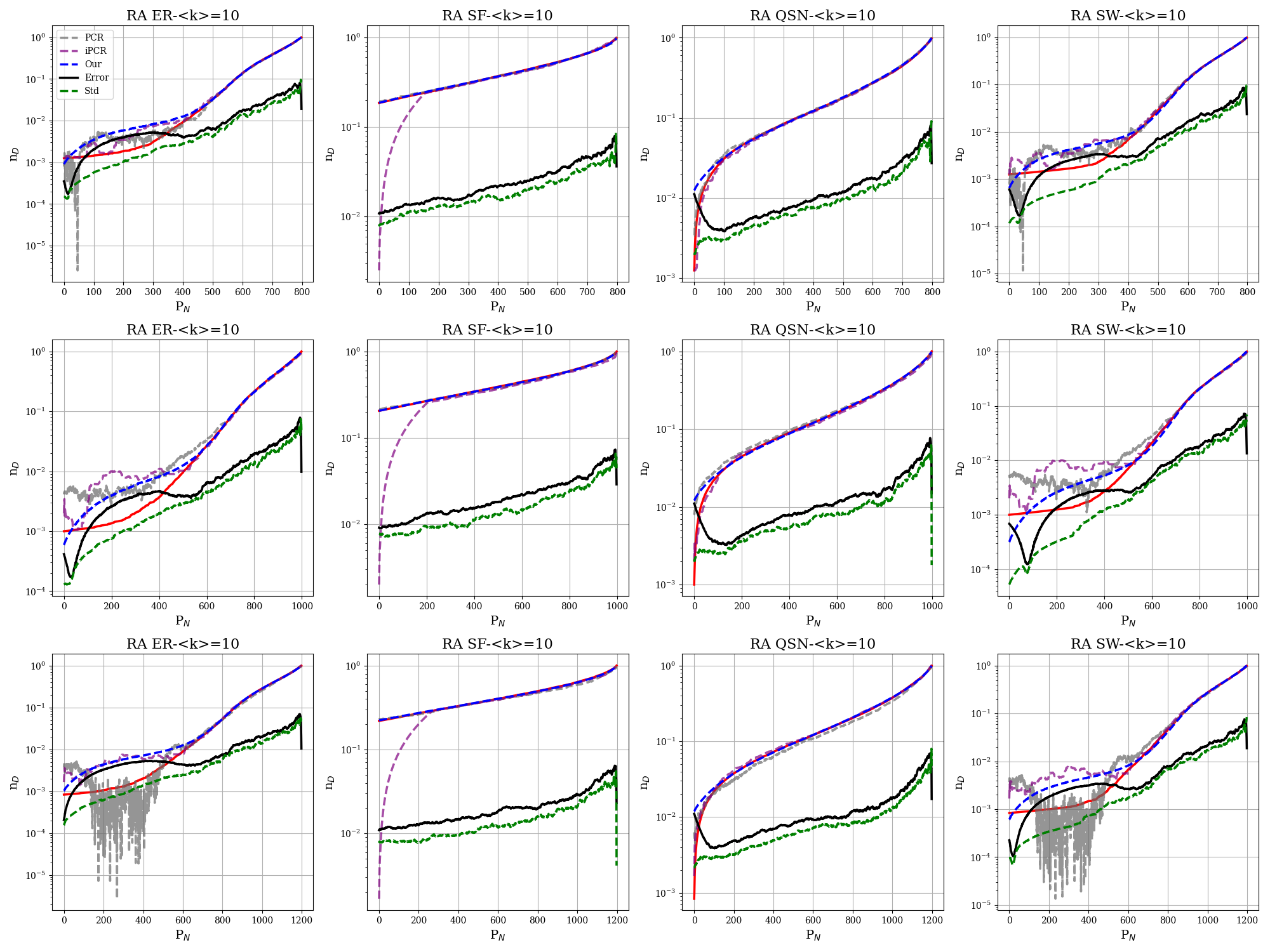}
  \caption{The predictions of the controllable robustness curves for various types of network with different node sizes under the RAs condition by the NCR-HoK, PCR\cite{27}, iPCR\cite{28} and CRL-SGNN\cite{CRL-SGNN} models.}
  \label{fig5}
\end{figure*}

{
\setlength{\tabcolsep}{3.5pt}
\begin{table}[th]
\centering
\caption{The average errors and standard deviations of the controllable robustness curve predicting for various types of networks with different node sizes under the RA condition are compared among NCR-HoK, PCR\cite{27}, iPCR\cite{28} and CRL-SGNN\cite{CRL-SGNN} models, along with a comprehensive ranking.}
\label{tab:table 3}
\begin{tabular}{|ccc|cccc|c|}
\hline
\multicolumn{3}{|l|}{}                                                                                                                              & \textbf{ER}                            & \textbf{SF}                            & \textbf{QSN}                           & \textbf{SW}                            & \textbf{Average}                            \\ \hline
\multicolumn{1}{|c|}{}                                  & \multicolumn{1}{l|}{}                                & $\overline{er}$                                  & \textbf{0.012}                         & \textbf{0.026}                         & \textbf{0.013}                         & \textbf{0.011}                         & \textbf{0.016(\#1)}                         \\
\multicolumn{1}{|c|}{}                                  & \multicolumn{1}{c|}{\multirow{-2}{*}{\textbf{PCR}}}  & \cellcolor[HTML]{D9D9D9}$\overline{\sigma}$          & \cellcolor[HTML]{DBDBDB}0.022          & \cellcolor[HTML]{DBDBDB}0.023          & \cellcolor[HTML]{DBDBDB}0.020          & \cellcolor[HTML]{DBDBDB}0.021          & \cellcolor[HTML]{DBDBDB}0.022(\#2)          \\ \cline{2-2}
\multicolumn{1}{|c|}{}                                  & \multicolumn{1}{l|}{}                                & $\overline{er}$                                  & \textbf{0.012}                         & 0.043                                  & \textbf{0.013}                         & \textbf{0.011}                         & 0.020(\#2)                                  \\
\multicolumn{1}{|c|}{}                                  & \multicolumn{1}{c|}{\multirow{-2}{*}{\textbf{iPCR}}} & \cellcolor[HTML]{D9D9D9}$\overline{\sigma}$          & \cellcolor[HTML]{DBDBDB}0.022          & \cellcolor[HTML]{DBDBDB}0.044          & \cellcolor[HTML]{DBDBDB}0.019          & \cellcolor[HTML]{DBDBDB}0.021          & \cellcolor[HTML]{DBDBDB}0.027(\#4)          \\ \cline{2-2}

\multicolumn{1}{|l|}{}                                  & \multicolumn{1}{l|}{}                                & $\overline{er}$                                  & \textbf{0.012}                         & \textbf{0.026}                          & 0.016                                 & \textbf{0.011}                         & 0.016(\#1)                         \\
\multicolumn{1}{|l|}{}                                  & \multicolumn{1}{c|}{\multirow{-2}{*}{\textbf{CRL-SGNN}}} & \cellcolor[HTML]{D9D9D9}$\overline{\sigma}$ & \cellcolor[HTML]{D9D9D9}0.024          & \cellcolor[HTML]{D9D9D9}0.029          & \cellcolor[HTML]{D9D9D9}0.022          & \cellcolor[HTML]{D9D9D9}0.023          & \cellcolor[HTML]{D9D9D9}0.025(\#3)          \\ \cline{2-2}

\multicolumn{1}{|c|}{}                                  & \multicolumn{1}{l|}{}                                & $\overline{er}$                                  & \textbf{0.012}                         & \textbf{0.026}                         & 0.014                                  & \textbf{0.011}                         & \textbf{0.016(\#1)}                         \\
\multicolumn{1}{|c|}{\multirow{-8}{*}{\textbf{N=800}}}  & \multicolumn{1}{c|}{\multirow{-2}{*}{\textbf{Our}}}  & \cellcolor[HTML]{D9D9D9}$\overline{\sigma}$          & \cellcolor[HTML]{DBDBDB}\textbf{0.008} & \cellcolor[HTML]{DBDBDB}\textbf{0.020} & \cellcolor[HTML]{DBDBDB}\textbf{0.010} & \cellcolor[HTML]{DBDBDB}\textbf{0.008} & \cellcolor[HTML]{DBDBDB}\textbf{0.012(\#1)} \\ \hline
\multicolumn{1}{|l|}{}                                  & \multicolumn{1}{l|}{}                                & $\overline{er}$                                  & 0.012                                  & 0.023                                  & 0.014                                  & 0.012                                  & 0.015(\#2)                                  \\
\multicolumn{1}{|l|}{}                                  & \multicolumn{1}{c|}{\multirow{-2}{*}{\textbf{PCR}}}  & \cellcolor[HTML]{D9D9D9}$\overline{\sigma}$          & \cellcolor[HTML]{DBDBDB}0.018          & \cellcolor[HTML]{DBDBDB}0.021          & \cellcolor[HTML]{DBDBDB}0.017          & \cellcolor[HTML]{DBDBDB}0.019          & \cellcolor[HTML]{DBDBDB}0.019(\#2)          \\ \cline{2-2}
\multicolumn{1}{|l|}{}                                  & \multicolumn{1}{l|}{}                                & $\overline{er}$                                  & 0.012                                  & 0.050                                  & 0.014                                  & 0.012                                  & 0.022(\#3)                                  \\
\multicolumn{1}{|l|}{}                                  & \multicolumn{1}{c|}{\multirow{-2}{*}{\textbf{iPCR}}} & \cellcolor[HTML]{D9D9D9}$\overline{\sigma}$          & \cellcolor[HTML]{DBDBDB}0.019          & \cellcolor[HTML]{DBDBDB}0.049          & \cellcolor[HTML]{DBDBDB}0.022          & \cellcolor[HTML]{DBDBDB}0.020          & \cellcolor[HTML]{DBDBDB}0.028(\#4)          \\ \cline{2-2}

\multicolumn{1}{|l|}{}                                  & \multicolumn{1}{l|}{}                                & $\overline{er}$                                  & 0.016                        & 0.036                       & 0.021                                  & 0.017                         & 0.023(\#4)                        \\
\multicolumn{1}{|l|}{}                                  & \multicolumn{1}{c|}{\multirow{-2}{*}{\textbf{CRL-SGNN}}} & \cellcolor[HTML]{D9D9D9}$\overline{\sigma}$ & \cellcolor[HTML]{D9D9D9}0.023          & \cellcolor[HTML]{D9D9D9}0.027          & \cellcolor[HTML]{D9D9D9}0.026          & \cellcolor[HTML]{D9D9D9}0.024          & \cellcolor[HTML]{D9D9D9}0.025(\#3)          \\ \cline{2-2}

\multicolumn{1}{|l|}{}                                  & \multicolumn{1}{l|}{}                                & $\overline{er}$                                 & \textbf{0.011}                         & \textbf{0.023}                         & \textbf{0.013}                         & \textbf{0.010}                         & \textbf{0.014(\#1)}                         \\
\multicolumn{1}{|l|}{\multirow{-8}{*}{\textbf{N=1000}}} & \multicolumn{1}{c|}{\multirow{-2}{*}{\textbf{Our}}}  & \cellcolor[HTML]{D9D9D9}$\overline{\sigma}$          & \cellcolor[HTML]{DBDBDB}\textbf{0.008} & \cellcolor[HTML]{DBDBDB}\textbf{0.017} & \cellcolor[HTML]{DBDBDB}\textbf{0.009} & \cellcolor[HTML]{DBDBDB}\textbf{0.008} & \cellcolor[HTML]{DBDBDB}\textbf{0.011(\#1)} \\ \hline
\multicolumn{1}{|l|}{}                                  & \multicolumn{1}{l|}{}                                & $\overline{er}$                                  & 0.011                                  & 0.026                                  & 0.018                                  & 0.011                                  & 0.017(\#3)                                  \\
\multicolumn{1}{|l|}{}                                  & \multicolumn{1}{c|}{\multirow{-2}{*}{\textbf{PCR}}}  & \cellcolor[HTML]{D9D9D9}$\overline{\sigma}$ & \cellcolor[HTML]{D9D9D9}0.018          & \cellcolor[HTML]{D9D9D9}0.024          & \cellcolor[HTML]{D9D9D9}0.020          & \cellcolor[HTML]{D9D9D9}0.018          & \cellcolor[HTML]{D9D9D9}0.020(\#2)          \\ \cline{2-2}
\multicolumn{1}{|l|}{}                                  & \multicolumn{1}{l|}{}                                & $\overline{er}$                                  & \textbf{0.010}                         & 0.046                                  & \textbf{0.011}                         & 0.011                                  & 0.020(\#2)                                  \\
\multicolumn{1}{|l|}{}                                  & \multicolumn{1}{c|}{\multirow{-2}{*}{\textbf{iPCR}}} & \cellcolor[HTML]{D9D9D9}$\overline{\sigma}$ & \cellcolor[HTML]{D9D9D9}0.016          & \cellcolor[HTML]{D9D9D9}0.053          & \cellcolor[HTML]{D9D9D9}0.017          & \cellcolor[HTML]{D9D9D9}0.018          & \cellcolor[HTML]{D9D9D9}0.026(\#4)          \\ \cline{2-2}

\multicolumn{1}{|l|}{}                                  & \multicolumn{1}{l|}{}                                & $\overline{er}$                                  & 0.016                         & 0.040                         & 0.023                                  & 0.019                         & 0.025(\#4)                         \\
\multicolumn{1}{|l|}{}                                  & \multicolumn{1}{c|}{\multirow{-2}{*}{\textbf{CRL-SGNN}}} & \cellcolor[HTML]{D9D9D9}$\overline{\sigma}$ & \cellcolor[HTML]{D9D9D9}0.022          & \cellcolor[HTML]{D9D9D9}0.025          & \cellcolor[HTML]{D9D9D9}0.026          & \cellcolor[HTML]{D9D9D9}0.024         & \cellcolor[HTML]{D9D9D9}0.024(\#3)          \\ \cline{2-2}

\multicolumn{1}{|l|}{}                                  & \multicolumn{1}{l|}{}                                & $\overline{er}$                                  & \textbf{0.010}                         & \textbf{0.021}                         & 0.012                                  & \textbf{0.010}                         & \textbf{0.013(\#1)}                         \\
\multicolumn{1}{|c|}{\multirow{-8}{*}{\textbf{N=1200}}} & \multicolumn{1}{c|}{\multirow{-2}{*}{\textbf{Our}}}  & \cellcolor[HTML]{D9D9D9}$\overline{\sigma}$          & \cellcolor[HTML]{D9D9D9}\textbf{0.007} & \cellcolor[HTML]{D9D9D9}\textbf{0.015} & \cellcolor[HTML]{D9D9D9}\textbf{0.009} & \cellcolor[HTML]{D9D9D9}\textbf{0.007} & \cellcolor[HTML]{D9D9D9}\textbf{0.010(\#1)} \\ \hline
\end{tabular}
\end{table}
}

\begin{table*}[t]
\centering
\caption{The details of real-world networks used in the experiment.}
\label{tab:table 4}
\begin{tabular}{llll}
\hline
\textbf{Real-World Networks} & \textbf{Brief description}                            & \textbf{Node Number} & \textbf{Edge Number} \\ \hline
DDG(DD-g97)                  & Protein                                               & 1021                 & 5778                 \\
DEL(delaunay-n10)            & DIMACS10 problem                                      & 1024                 & 3056                 \\
DW5(dwt-1005)                & \multirow{2}{*}{Symmetric connection from Washington} & 1005                 & 4813                 \\
DW7(dwt-1007)                &                                                       & 1007                 & 4791                 \\
LSH(lshp1009)                & Alan George’s   L-shape problem                       & 1009                 & 3937                 \\
ORS(orsirr-1)                & oil reservoir   simulation                            & 1030                 & 6858                 \\ \hline
\end{tabular}
\end{table*}

\begin{table*}[ht]
\centering
\caption{The evaluation performance of NCR-HoK, PCR\cite{27}, iPCR\cite{28} and CRL-SGNN\cite{CRL-SGNN}  on real-world datasets.}
\label{tab:table 5}
\begin{tabular}{ccccccccc}
\hline
\multicolumn{2}{l}{\textbf{Real-World Networks}} & \textbf{DDG}   & \textbf{DEL}   & \textbf{DW5}   & \textbf{DW7}   & \textbf{LSH}   & \textbf{ORS}   & \textbf{Average}    \\ \hline
\multirow{2}{*}{\textbf{PCR}}    & $\overline{er}$   & 0.053          & 0.210          & 0.157          & 0.146          & 0.195          & 0.147          & 0.151(\#3)          \\
                                 & $\overline{\sigma}$    & 0.037          & 0.076          & 0.222          & 0.211          & 0.223          & 0.210          & 0.163(\#3)          \\ \hline
\multirow{2}{*}{\textbf{iPCR}}   & $\overline{er}$    & 0.039          & 0.285          & 0.154 & 0.145          & 0.228          & 0.118          & 0.162(\#4)          \\
                                 & $\overline{\sigma}$    & 0.023          & 0.116          & 0.221          & 0.214          & 0.223          & 0.193          & 0.165(\#4)          \\ \hline

\multirow{2}{*}{\textbf{CRL-SGNN}}   & $\overline{er}$    & 0.033          & 0.043          & \textbf{0.119}  & \textbf{0.091}          & \textbf{0.123}          & 0.205         & 0.102(\#2)          \\
                                 & $\overline{\sigma}$    & 0.028          & 0.039          & 0.186          & 0.164         & 0.184        & 0.063         & 0.111(\#2)          \\ \hline

\multirow{2}{*}{\textbf{Our}}    & $\overline{er}$   & \textbf{0.005} & \textbf{0.007} & 0.162          & 0.118 & 0.184 & \textbf{0.003} & \textbf{0.080(\#1)} \\
                                 & $\overline{\sigma}$    & \textbf{0.013} & \textbf{0.017} & \textbf{0.012} & \textbf{0.015} & \textbf{0.011} & \textbf{0.008} & \textbf{0.013(\#1)} \\ \hline
\end{tabular}
\end{table*}

\subsubsection{Predicting the Controllability and Robustness Curves of Networks with Different Average Degrees}
To explore the transferability of NCR-HoK's learning on controllable robustness across networks with different average degrees, we conduct tests under the same RA conditions for four types of topological structures: ER, SF, QSN, and SW. We randomly generate 100 test samples for each network type with average out-degrees of $\textless{}k\textgreater{}=4$ and 7, respectively. The networks are tested using the NCR-HoK trained in Section V. A, and the results are shown in table \ref{tab:table 2} and Fig.\ref{fig4}.

As can be seen from Fig.\ref{fig4}, although NCR-HoK is trained on datasets with $\textless{}k\textgreater{}=2,5,8,10$, it has good migration ability, and can still predict the controllable robustness curves of the network stably with $\textless{}k\textgreater{}=4,7$. As shown by the blue line in the figure, the prediction curve is smoother and closer to the ground true relative to the baseline. On the other hand, the PCR and iPCR predictions are much steeper, especially the PCR, which shows a jittery behavior at the beginning. Combined with table \ref{tab:table 2}, it can be seen that the average error of NCR-HoK predictions is better than that of the baseline in most cases. In addition, the average standard deviation of NCR-HoK predictions is the smallest regardless of the network topology, which indicates that NCR-HoK is able to maintain stable prediction results.

\subsubsection{Predicting the Controllability Robustness Curve of Networks with Different Node Sizes}
In order to explore the performance of NCR-HoK across networks of varying node sizes, we conduct experiments under the same RA conditions across four different topologies: ER, SF, QSN, and SW. We randomly generate 800 networks for each topology with an average out-degree of $\textless{}k\textgreater{}=10$ and different node scales. For each network topology, 100 individual networks are independently generated as test samples. We retrain NCR-HoK using datasets from different node sizes and complete the corresponding prediction tests. The results are shown in table \ref{tab:table 3} and Fig.\ref{fig5}. 

As can be seen from Fig.\ref{fig5}, on SF and QSN networks, NCR-HoK is still able to give smooth and accurate predictions of network controllable robustness curves, while on ER and SW networks, NCR-HoK does not have good enough prediction performance in the early stage of the network controllable robustness, but it is still able to quickly approximate the trend of the network controllable robustness in the middle and late stages. And compared with the two baselines of PCR and iPCR, NCR-HoK is more stable. Combined with table \ref{tab:table 3}, for relatively small scale networks, PCR performs better than iPCR and CRL-SGNN. And the NCR-HoK prediction results are ranked first in terms of mean error and mean standard deviation.

\begin{figure}[th]
  \centering
  \includegraphics[width=2.5in]{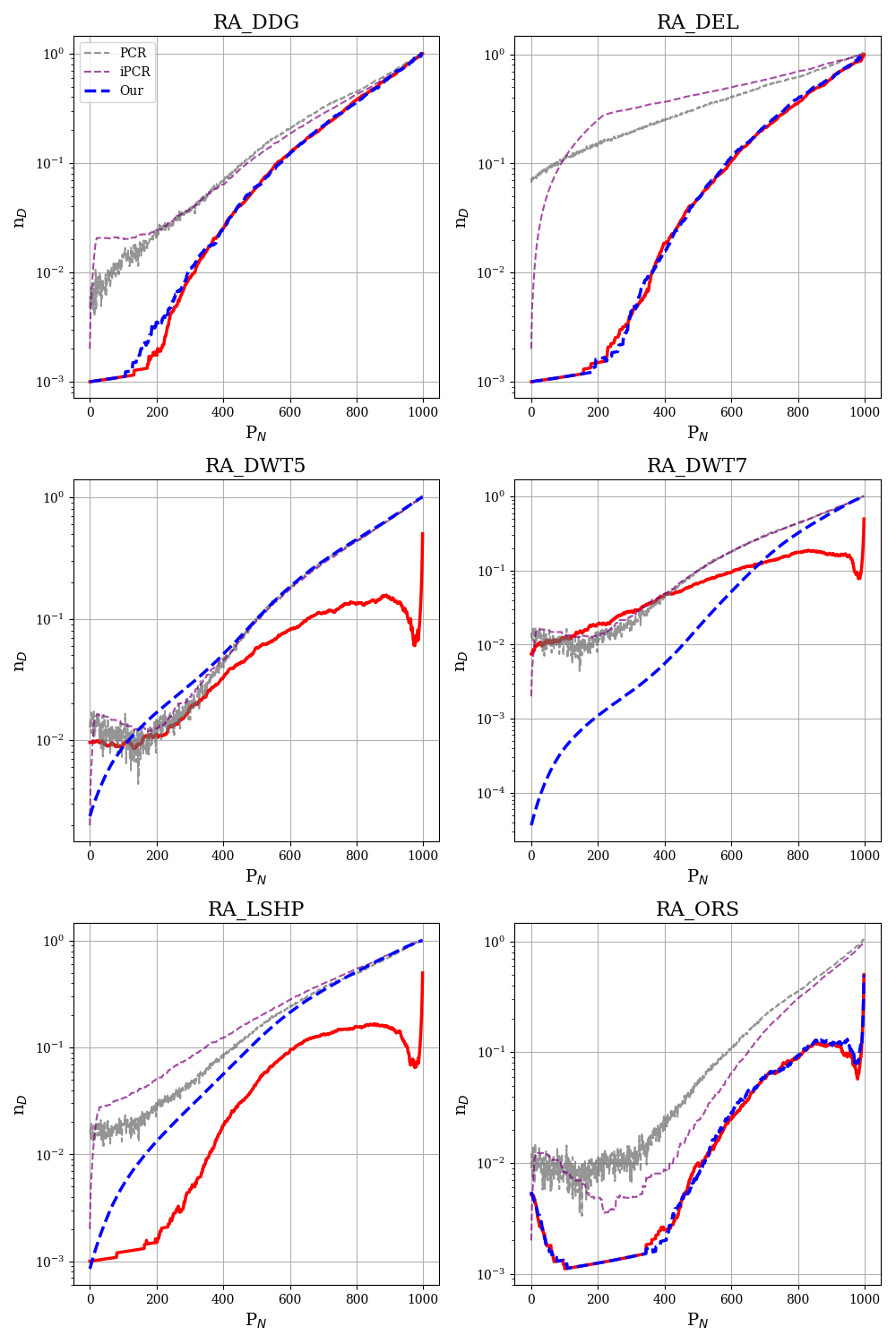}
  \caption{Comparison of controllability robustness curve predicting results of NCR-HoK, PCR\cite{27}, iPCR\cite{28} and CRL-SGNN\cite{CRL-SGNN} on real-world networks.}
  \label{fig6}
\end{figure}

\begin{figure*}[t]
  \centering
  \includegraphics[width=5.5in]{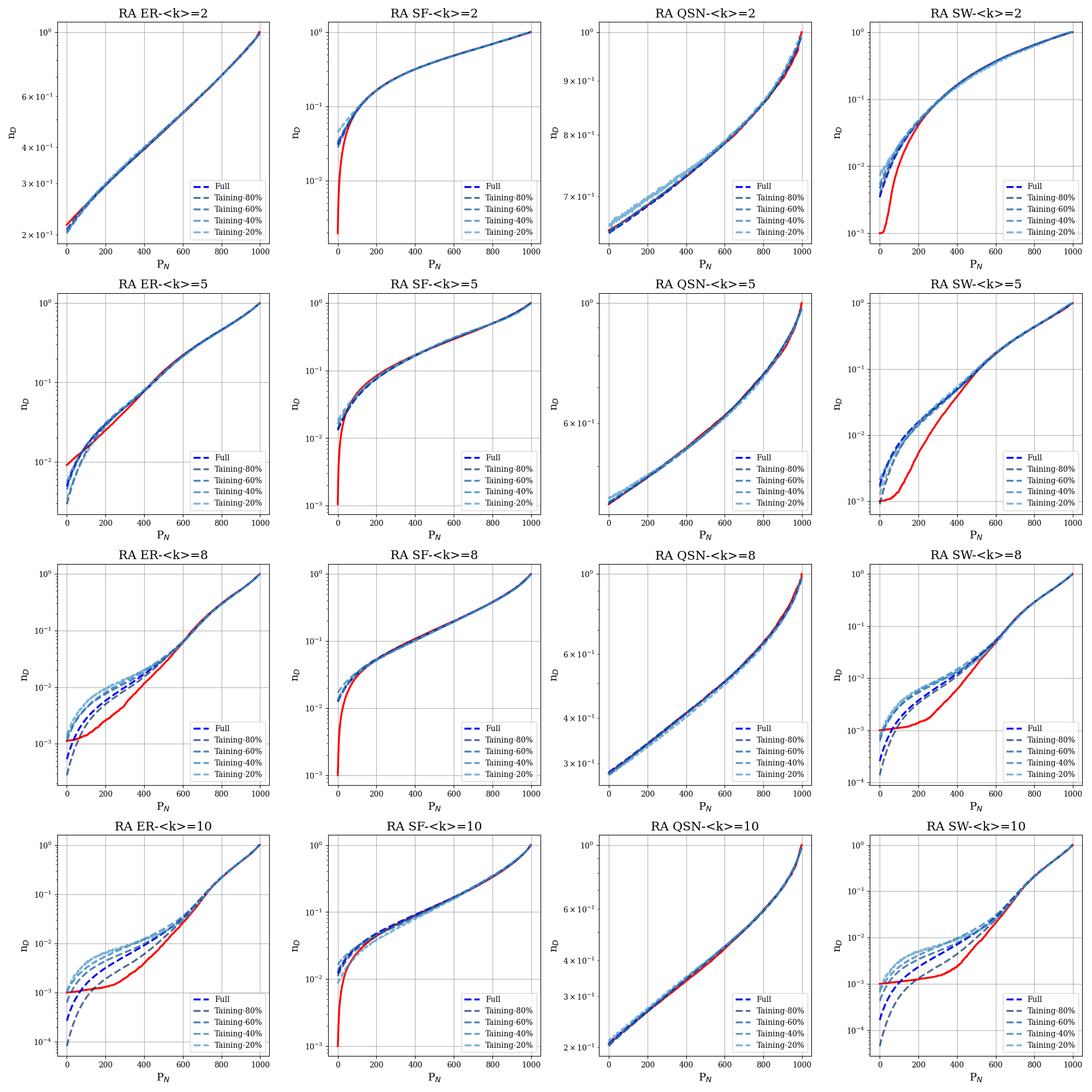}
  \caption{NCR-HoK's controlled robustness curve prediction results for various networks under 20\%, 40\%, 60\%, and 80\% reduction of the training dataset, respectively.}
  \label{fig7}
\end{figure*}
\subsection{Robustness Learning for Real-World Networks}
In addition to evaluating the performance of NCR-HoK on synthetic datasets, we also selected some real-world datasets from Network Repository\footnote{\url{http://networkrepository.com/}} to assess the effectiveness and reliability of our model. The chosen datasets cover areas such as biological proteins, internet data, and oil industrial data, with detailed information presented in table \ref{tab:table 4}. Since the node count in these datasets slightly exceeds 1000 nodes, following the method described in \cite{28}, we randomly select and remove some nodes until the desired size was reached. For each network, we resize randomly 10 times and then averaged the prediction results and errors. The predicted results can be seen in table \ref{tab:table 5} and Fig.\ref{fig6}.

\begin{figure*}[!t]
  \centering
  \includegraphics[width=5.5in]{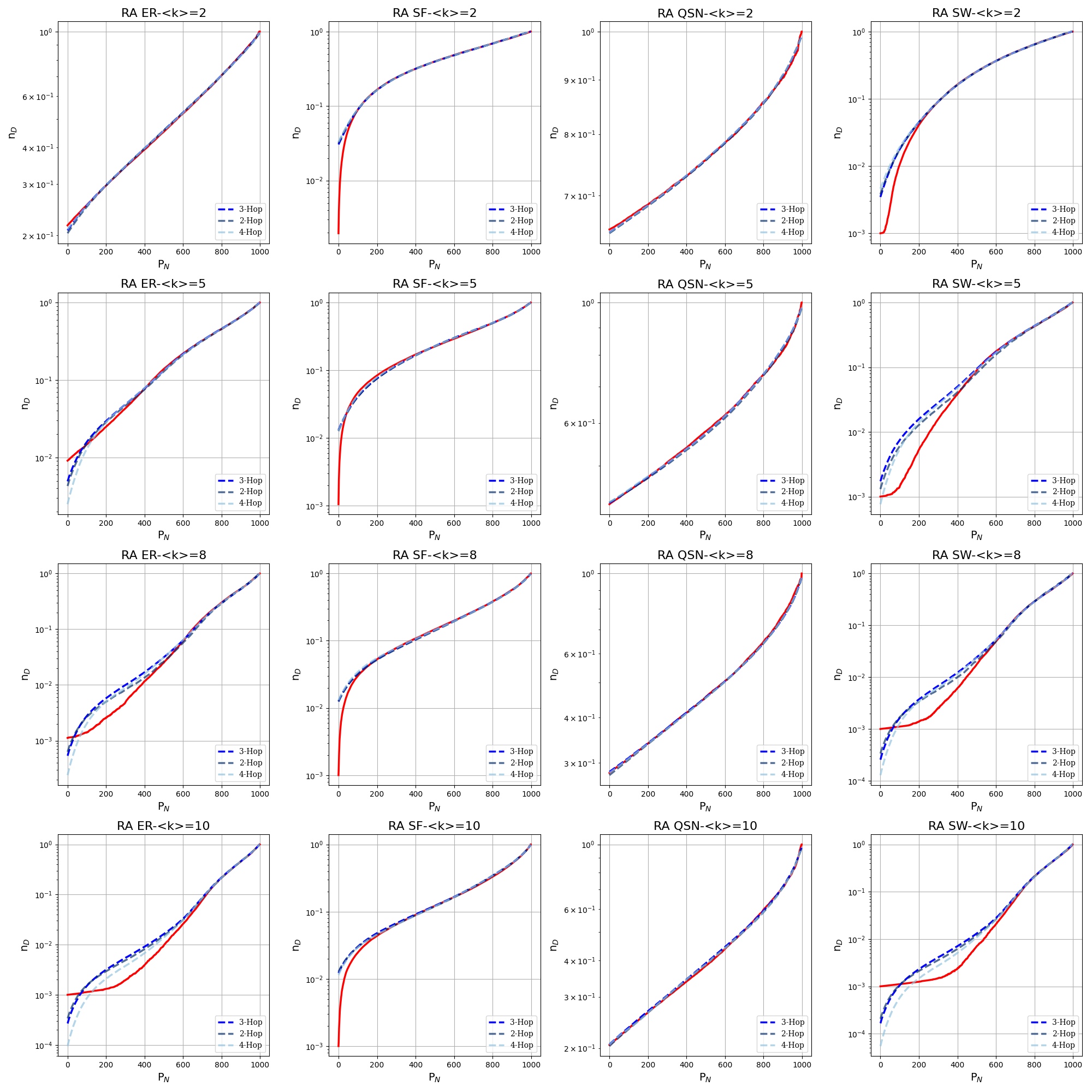}
  \caption{The prediction results of NCR-HoK's controllable robustness curves for various network topologies for K-Hop's K-values of 2, 3, and 4 cases, respectively.}
  \label{fig8}
\end{figure*}

\begin{figure*}[!t]
  \centering
  \includegraphics[width=5.5in]{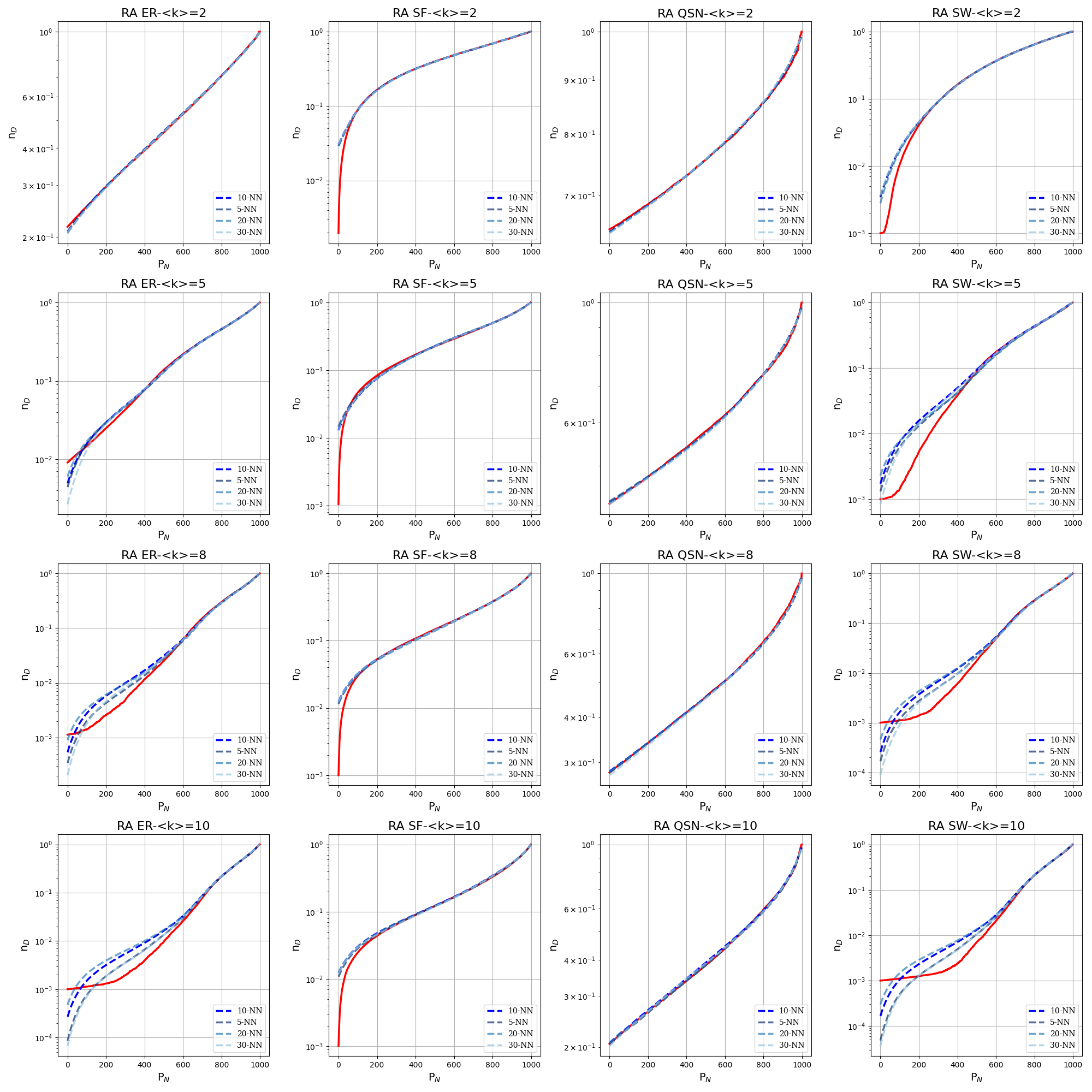}
  \caption{The prediction results of NCR-HoK's controllable robustness curves for various network topologies for K-NN's K-values of 5, 10, 20 and 30 cases, respectively.}
  \label{fig9}
\end{figure*}

{
\setlength{\tabcolsep}{3pt}
\begin{table}[H]
\centering
\caption{Results of ablation experiments with $BC_{GAT}$ module, \textit{w.o. BC} indicates no $BC_{GAT}$ module, \textit{w.o. S-HGNN} indicates only using a single HGNN module,  \textit{w.o. Dual HGNN} indicates no Dual HGNN module, \textit{Raw} indicates original setting.}
\label{tab:table abu}
\begin{tabular}{|ccc|cccc|}
\hline
\multicolumn{3}{|c|}{}    & \textbf{\textless{}k\textgreater{}=2}  & \textbf{\textless{}k\textgreater{}=5}  & \textbf{\textless{}k\textgreater{}=8}  & \textbf{\textless{}k\textgreater{}=10}    \\ \hline

\multicolumn{1}{|c|}{}   & \multicolumn{1}{l|}{}    & $\overline{er}$    & 0.037   & 0.027  & 0.017    & 0.017    \\
\multicolumn{1}{|c|}{}   & \multicolumn{1}{c|}{\multirow{-2}{*}{\textbf{\textit{w.o. BC}}}}  & \cellcolor[HTML]{D9D9D9}$\overline{\sigma}$ & \cellcolor[HTML]{D9D9D9}0.020  & \cellcolor[HTML]{D9D9D9}0.014         & \cellcolor[HTML]{D9D9D9}0.012       & \cellcolor[HTML]{D9D9D9}0.011               \\ \cline{2-2}

\multicolumn{1}{|c|}{}   & \multicolumn{1}{l|}{}    & $\overline{er}$    & 0.214   & 0.098  & 0.061    & 0.047    \\
\multicolumn{1}{|c|}{}   & \multicolumn{1}{c|}{\multirow{-2}{*}{\textbf{\textit{w.o. S-HGNN}}}}  & \cellcolor[HTML]{D9D9D9}$\overline{\sigma}$ & \cellcolor[HTML]{D9D9D9}0.018  & \cellcolor[HTML]{D9D9D9}0.018         & \cellcolor[HTML]{D9D9D9}0.012       & \cellcolor[HTML]{D9D9D9}0.011               \\ \cline{2-2}

\multicolumn{1}{|c|}{}   & \multicolumn{1}{l|}{}    & $\overline{er}$    & 0.340   & 0.105  & 0.056    & 0.044    \\
\multicolumn{1}{|c|}{}   & \multicolumn{1}{c|}{\multirow{-2}{*}{\textbf{\textit{w.o. Dual HGNN}}}}  & \cellcolor[HTML]{D9D9D9}$\overline{\sigma}$ & \cellcolor[HTML]{D9D9D9}0.036  & \cellcolor[HTML]{D9D9D9}0.021         & \cellcolor[HTML]{D9D9D9}0.012       & \cellcolor[HTML]{D9D9D9}0.010               \\ \cline{2-2}

\multicolumn{1}{|c|}{}                               & \multicolumn{1}{l|}{}                                & $\overline{er}$                         & \textbf{0.016}                         & \textbf{0.015}                         & \textbf{0.012}                         & \textbf{0.010}                                               \\
\multicolumn{1}{|c|}{\multirow{-8}{*}{\textbf{ER}}}  & \multicolumn{1}{c|}{\multirow{-2}{*}{\textbf{\textit{Raw}}}}  & \cellcolor[HTML]{D9D9D9}$\overline{\sigma}$ & \cellcolor[HTML]{D9D9D9}\textbf{0.012} & \cellcolor[HTML]{D9D9D9}\textbf{0.011} & \cellcolor[HTML]{D9D9D9}\textbf{0.009} & \cellcolor[HTML]{D9D9D9}\textbf{0.008} \\ \hline

\multicolumn{1}{|l|}{}                               & \multicolumn{1}{l|}{}                                & $\overline{er}$                         & 0.041                                  & 0.021                                  & 0.030                         & 0.030                                                    \\
\multicolumn{1}{|l|}{}                               & \multicolumn{1}{c|}{\multirow{-2}{*}{\textbf{\textit{w.o. BC}}}}  & \cellcolor[HTML]{D9D9D9}$\overline{\sigma}$ & \cellcolor[HTML]{D9D9D9}0.020         & \cellcolor[HTML]{D9D9D9}0.017          & \cellcolor[HTML]{D9D9D9}0.020         & \cellcolor[HTML]{D9D9D9}0.021                    \\ \cline{2-2}

\multicolumn{1}{|c|}{}   & \multicolumn{1}{l|}{}    & $\overline{er}$    & 0.359   & 0.287  & 0.225    & 0.200    \\
\multicolumn{1}{|c|}{}   & \multicolumn{1}{c|}{\multirow{-2}{*}{\textbf{\textit{w.o. S-HGNN}}}}  & \cellcolor[HTML]{D9D9D9}$\overline{\sigma}$ & \cellcolor[HTML]{D9D9D9}0.020  & \cellcolor[HTML]{D9D9D9}0.024        & \cellcolor[HTML]{D9D9D9}0.024       & \cellcolor[HTML]{D9D9D9}0.027               \\ \cline{2-2}

\multicolumn{1}{|c|}{}   & \multicolumn{1}{l|}{}    & $\overline{er}$    & 0.449   & 0.407  & 0.258    & 0.220    \\
\multicolumn{1}{|c|}{}   & \multicolumn{1}{c|}{\multirow{-2}{*}{\textbf{\textit{w.o. Dual HGNN}}}}  & \cellcolor[HTML]{D9D9D9}$\overline{\sigma}$ & \cellcolor[HTML]{D9D9D9}0.020  & \cellcolor[HTML]{D9D9D9}0.035         & \cellcolor[HTML]{D9D9D9}0.035       & \cellcolor[HTML]{D9D9D9}0.033               \\ \cline{2-2}

\multicolumn{1}{|l|}{}                               & \multicolumn{1}{l|}{}                                & $\overline{er}$                         & \textbf{0.016}                         & \textbf{0.019}                         & \textbf{0.021}                         & \textbf{0.021}                                            \\

\multicolumn{1}{|c|}{\multirow{-8}{*}{\textbf{SF}}}  & \multicolumn{1}{c|}{\multirow{-2}{*}{\textbf{\textit{o. BC}}}}  & \cellcolor[HTML]{D9D9D9}$\overline{\sigma}$ & \cellcolor[HTML]{D9D9D9}\textbf{0.012} & \cellcolor[HTML]{D9D9D9}\textbf{0.015} & \cellcolor[HTML]{D9D9D9}\textbf{0.015} & \cellcolor[HTML]{D9D9D9}\textbf{0.016}  \\ \hline

\multicolumn{1}{|l|}{}                               & \multicolumn{1}{l|}{}                                & $\overline{er}$                         & 0.049                        & 0.017                                  & 0.021                        & 0.029                                                      \\
\multicolumn{1}{|l|}{}                               & \multicolumn{1}{c|}{\multirow{-2}{*}{\textbf{\textit{w.o. BC}}}}  & \cellcolor[HTML]{D9D9D9}$\overline{\sigma}$ & \cellcolor[HTML]{D9D9D9}0.019          & \cellcolor[HTML]{D9D9D9}0.012          & \cellcolor[HTML]{D9D9D9}0.011          & \cellcolor[HTML]{D9D9D9}0.014                  \\ \cline{2-2}

\multicolumn{1}{|c|}{}   & \multicolumn{1}{l|}{}    & $\overline{er}$    & 0.185   & 0.125  & 0.100   & 0.090    \\
\multicolumn{1}{|c|}{}   & \multicolumn{1}{c|}{\multirow{-2}{*}{\textbf{\textit{w.o. S-HGNN}}}}  & \cellcolor[HTML]{D9D9D9}$\overline{\sigma}$ & \cellcolor[HTML]{D9D9D9}0.019  & \cellcolor[HTML]{D9D9D9}0.019         & \cellcolor[HTML]{D9D9D9}0.015       & \cellcolor[HTML]{D9D9D9}0.018               \\ \cline{2-2}

\multicolumn{1}{|c|}{}   & \multicolumn{1}{l|}{}    & $\overline{er}$    & 0.202   & 0.145  & 0.017    & 0.017    \\
\multicolumn{1}{|c|}{}   & \multicolumn{1}{c|}{\multirow{-2}{*}{\textbf{\textit{w.o. Dual HGNN}}}}  & \cellcolor[HTML]{D9D9D9}$\overline{\sigma}$ & \cellcolor[HTML]{D9D9D9}0.022  & \cellcolor[HTML]{D9D9D9}0.020         & \cellcolor[HTML]{D9D9D9}0.016       & \cellcolor[HTML]{D9D9D9}0.016               \\ \cline{2-2}

\multicolumn{1}{|l|}{}                               & \multicolumn{1}{l|}{}                                & $\overline{er}$                         & \textbf{0.015}                         & \textbf{0.015}                                  & \textbf{0.013}                                  & \textbf{0.012}                                         \\

\multicolumn{1}{|c|}{\multirow{-8}{*}{\textbf{QSN}}} & \multicolumn{1}{c|}{\multirow{-2}{*}{\textbf{\textit{o. BC}}}}  & \cellcolor[HTML]{D9D9D9}$\overline{\sigma}$ & \cellcolor[HTML]{D9D9D9}\textbf{0.011} & \cellcolor[HTML]{D9D9D9}\textbf{0.011} & \cellcolor[HTML]{D9D9D9}\textbf{0.009} & \cellcolor[HTML]{D9D9D9}\textbf{0.008}  \\ \hline
\multicolumn{1}{|l|}{}                               & \multicolumn{1}{l|}{}                                & $\overline{er}$                         & 0.013                                  & 0.030                         & 0.016                                  & 0.016                                                                 \\
\multicolumn{1}{|l|}{}                               & \multicolumn{1}{c|}{\multirow{-2}{*}{\textbf{\textit{w.o. BC}}}}  & \cellcolor[HTML]{D9D9D9}$\overline{\sigma}$ & \cellcolor[HTML]{D9D9D9}0.010          & \cellcolor[HTML]{D9D9D9}0.014         & \cellcolor[HTML]{D9D9D9}0.012         & \cellcolor[HTML]{D9D9D9}0.010                \\ \cline{2-2}

\multicolumn{1}{|c|}{}   & \multicolumn{1}{l|}{}    & $\overline{er}$    & 0.139   & 0.082  & 0.053    & 0.042    \\
\multicolumn{1}{|c|}{}   & \multicolumn{1}{c|}{\multirow{-2}{*}{\textbf{\textit{w.o. S-HGNN}}}}  & \cellcolor[HTML]{D9D9D9}$\overline{\sigma}$ & \cellcolor[HTML]{D9D9D9}0.013  & \cellcolor[HTML]{D9D9D9}0.013         & \cellcolor[HTML]{D9D9D9}0.012       & \cellcolor[HTML]{D9D9D9}0.010               \\ \cline{2-2}

\multicolumn{1}{|c|}{}   & \multicolumn{1}{l|}{}    & $\overline{er}$    & 0.127   & 0.090  & 0.017    & 0.017    \\
\multicolumn{1}{|c|}{}   & \multicolumn{1}{c|}{\multirow{-2}{*}{\textbf{\textit{w.o. Dual HGNN}}}}  & \cellcolor[HTML]{D9D9D9}$\overline{\sigma}$ & \cellcolor[HTML]{D9D9D9}0.013  & \cellcolor[HTML]{D9D9D9}0.016         & \cellcolor[HTML]{D9D9D9}0.011       & \cellcolor[HTML]{D9D9D9}0.010              \\ \cline{2-2}

\multicolumn{1}{|l|}{}                               & \multicolumn{1}{l|}{}                                & $\overline{er}$                         & \textbf{0.012}                         & \textbf{0.014}                         & \textbf{0.012}                         & \textbf{0.010}                              \\

\multicolumn{1}{|c|}{\multirow{-8}{*}{\textbf{SW}}}  & \multicolumn{1}{c|}{\multirow{-2}{*}{\textbf{\textit{o. BC}}}}  & \cellcolor[HTML]{D9D9D9}$\overline{\sigma}$ & \cellcolor[HTML]{D9D9D9}\textbf{0.009} & \cellcolor[HTML]{D9D9D9}\textbf{0.009} & \cellcolor[HTML]{D9D9D9}\textbf{0.008} & \cellcolor[HTML]{D9D9D9}\textbf{0.007} \\ \hline
\end{tabular}
\end{table}
}

From table \ref{tab:table 5} and Fig.\ref{fig6}, it can be seen that NCR-HoK is better than other baselines, both in terms of accuracy and stability of prediction. Especially, on DDG, DEL and ORS datasets, NCR-HoK is able to predict the trend of the network's controllable robustness in the presence of random attacks very accurately. This suggests that NCR-HoK is somewhat applicable to the analysis of controllable robustness for real-life networks. For real-world networks including DW5, DW7 and LSH, we think their unique structural properties might not be fully captured by our current hypergraph construction methods and this is an area that needs to be improved in the future for our model.

\subsection{Ablation Studies}
To investigate the effectiveness of each component in our NCR-HoK model, we conduct a series of experiments. We evaluate the performance when removing the betweenness centrality feature (\textit{w.o. BC}), removing both hypergraph attention channels (\textit{w.o. Dual HGNN}), and using only a single hypergraph attention channel (\textit{w.o. S-HGNN}). The performance of these variants of our method is compared against the full model (\textit{Raw}). As shown in the table \ref{tab:table abu}:

\subsubsection{Effectiveness of the $BC_{GAT}$ module} Across all datasets and thresholds, the models with the $BC_{GAT}$ module (\textit{o. BC}) consistently outperform the ones without it (\textit{w.o. BC}), as indicated by the lower values of $\overline{er}$. This trend highlights the significant contribution of the $BC_{GAT}$ component to capturing the critical structural features of the network that influence controllable robustness. Notably, for real-world networks like QSN and SW, the improvement is substantial, especially under higher degree thresholds, demonstrating the model's scalability and robustness.

\subsubsection{Contribution of the Dual HGNN Architecture} The experiments also underscore the importance of the hypergraph-based learning channels. When the entire Dual HGNN component is removed ($w.o. Dual HGNN$), leaving only the GAT module, the model's performance significantly deteriorates, demonstrating the necessity of hypergraph high-order information. Similarly, the $w.o. S-HGNN$ experiment validates the importance of having two independent hypergraph channels by using only a single HGNN module. This variant exhibits a significant performance drop, highlighting the advantages of the dual-channel architecture.

\subsection{Parameter Analysis}
To assess the impact of data volume on model performance, as well as the contribution and effectiveness of key model components, we conduct parameter analysis experiments on the data samples used for model training and the k-values in hypergraph generation methods such as K-Hop and K-NN.

\subsubsection{Training Split}
According to the experimental setup in Section V. A, for each network topology with different average degrees $\textless{}k\textgreater{}$, the original training sample consisted of 800 samples. In order to explore the impact of the amount of training data on model performance, the size of the training data is reduce to 20\%, 40\%, 60\%, and 80\% respectively, while the size of the test data samples remained unchanged. Some experimental results can be seen from \textbf{Supplementary Materials VII. D} due to page limitation.

Fig.\ref{fig7} shows that for both types of network structures, SF and QSN, NCR-HoK is able to predict the controllable robustness curves of the network structures well for dataset sizes ranging from 20\% to 100\%. For the QSN network, when the dataset size is reduced to 20\%, the prediction results of NCR-HoK (shown by the lightest blue line) show fluctuations, but the prediction curves are still generally in line with the trend of the controllable robustness curves. As for the ER and SW network structures, it can be seen that NCR-HoK is able to fit the network's controllable robustness curve well when the average network out-degree is low, but as the average network out-degree increases, the pre-prediction effect of NCR-HoK on the network's controllable robustness curve gradually deteriorates. The best prediction is achieved with 80\% training data volume while changing the size of the training dataset. And as can be seen from the results in table \ref{tab:table S1}, the average error and standard deviation of NCR-HoR's predictions of controllable robustness profiles for a variety of network topologies remain at a low level for different training data sizes.

\subsubsection{The k-value of the K-Hop and K-NN}
In this section, in order to explore the impact of hypergraph structures constructed by K-Hop, K-NN methods on complex networks and embedding spaces on the performance of NCR-HoR, we analyze the ablation experiments by varying the K-value of the K-Hop and K-NN methods. 

Similarly, following the experimental setup in Section V. A, we generate initial hypergraphs using the 2-Hop and 4-Hop. In the embedding space, we generate hypergraphs using the 5-NN, 20-NN, and 30-NN.

From Figs. \ref{fig8} and \ref{fig9}, we can observe that NCR-HoK consistently gives better predictions of the controllable robustness profiles of various networks under different K-value settings of the K-Hop and K-NN methods. When a larger value of K is chosen, such as 4-Hop or 30-NN, the predicted controllable robustness curves are closer to the true curves and more accurate. This may be due to the fact that the choice of K-value is equivalent to the feeling field of constructing the hypergraph, but when the K-value is larger, the constructed hypergraph covers more comprehensive high-order knowledge, so the prediction performance of the model will be better. The table of mean error and standard deviation results is detailed in the \textbf{Supplementary Materials Tables S3 and S4}.

\begin{figure}[h!]
  \centering
  \includegraphics[width=3.5in]{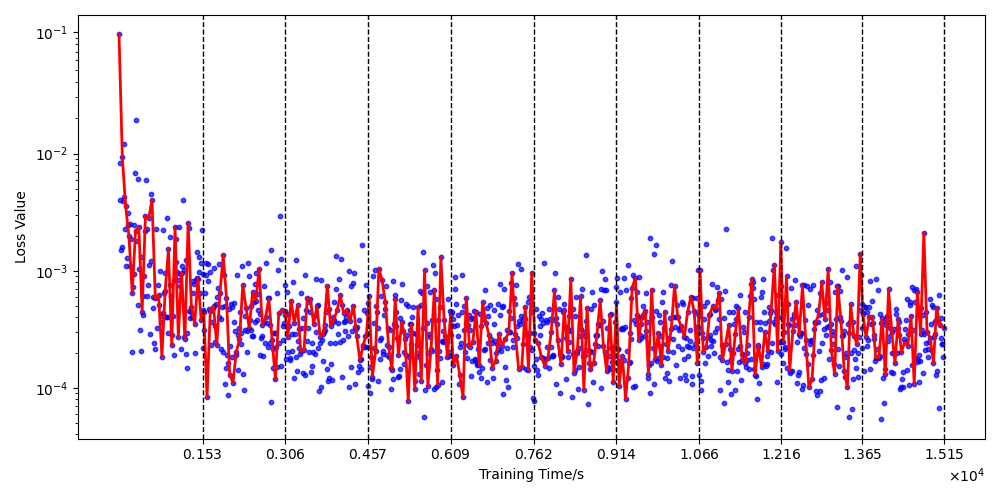}
  \caption{The training process of NCR-HoK for 3200 networks with 1000 nodes.}
  \label{fig10}
\end{figure}

\begin{table}[t]
\centering
\caption{The average time taken to process each graph for RAS, PCR\cite{27}, iPCR\cite{28}, CRL-SGNN\cite{CRL-SGNN} and NCR-HoK.}
\label{tab:table 6}
\begin{tabular}{cc}
\hline
\textbf{Controllability Robustness} & \textbf{Unit: Second} \\ \hline
RAS                                   & 23.581                \\
PCR                                   & 0.306                 \\
iPCR                                  & 0.325                 \\
CRL-SGNN                                  & 0.029                \\
NCR-HoK                                   & 0.085                 \\ \hline
\end{tabular}
\end{table}

\subsection{Running Time Comparison}
In addition to accuracy, another evaluation metric widely used to test the performance of network robustness learning algorithms is running time consumption. In this subsection, the time required to train the NCR-HoK model, the decreasing trend of training loss, and the time required to process a network with a node size of 1000 are discussed. All experiments, including baseline models, are conducted on the same computational platform to ensure fair and consistent comparisons. Baseline models are implemented using the default settings described in their original papers.

According to the experimental setup described in Section V. A, the average processing time per graph during model training is 0.118 seconds. In comparison, the PCR and iPCR models require 2.667 seconds and 5.898 seconds per graph, respectively, making NCR-HoK more efficient and simpler to train. As shown in Figure \ref{fig10}, training NCR-HoK for one epoch takes only 1530 seconds, and the loss decreases rapidly. 

Table \ref{tab:table 6} summarizes the average runtime across multiple independent tests using random attack simulations, PCR, iPCR, CRL-SGNN, and NCR-HoK on ER, SF, QSN, and SW networks, each with 1000 nodes and average degrees of $\textless{}k\textgreater{} = 2, 5, 8, 10$. The results in the table indicate the average processing time per graph. It can be observed that, on graphs with 1000 nodes, compared to NCR-HoK (which is based on a spatial GNN framework), CRL-SGNN has a training and inference time of approximately 0.029 seconds per graph, which gives it a clear advantage due to its spectral GNN foundation. However, NCR-HoK still demonstrates significant advantages in terms of efficiency over the baseline methods including PCR and iPCR as well as traditional attack simulation approaches.

\section{Conclusion and Feature Work}

This paper introduces NCR-HoK, a graph attention network that exploits high-order knowledge for robustness learning and controllability-robustness curve prediction. As the first systematic study of high-order hypergraph structures on controllable robustness, NCR-HoK integrates a node-feature encoder, hypergraphs built on high-order relations, and a dual-hypergraph attention module. The model simultaneously captures 1) explicit graph topology, 2) high-order neighborhood connections, and 3) latent structures in the embedding space—without expensive attack simulations. Experiments on synthetic and real-world networks demonstrate superior performance with low computational cost. Future work will extend controllability-robustness learning to broader classes of complex networks. However, to date, NCR-HoK cannot make predictions for time-varying, dynamic complex networks. This remains one of the existing research challenges. Therefore, future work will focus on developing controllability-based robustness learning and prediction methods for dynamic networks as well as more complex types of network such as attributed networks and heterogeneous networks.

\bibliographystyle{IEEEtran}
\bibliography{sample-base}

\newpage
\clearpage

\begin{strip}
\begin{center}
    {\LARGE \textbf{Supplemental Materials for “High-order Knowledge Based Network Controllability Robustness Prediction: A Hypergraph Neural Network Approach”}}\\[1em]
    {\large Shibing Mo, Jiarui Zhang, Jiayu Xie, Xiangyi Teng, and Jing Liu}\\[1em]
\end{center}
\end{strip}

\section{Supplemental Materials}

\setcounter{table}{0}   
\setcounter{figure}{0}
\renewcommand{\thetable}{S\arabic{table}}
\renewcommand{\thefigure}{S\arabic{figure}}

\newcounter{Sfigure}
\setcounter{Sfigure}{1}
\renewcommand{\thefigure}{S\arabic{Sfigure}}

\subsection{Related Work}

\subsubsection{The application of hypergraph neural networks}

Hypergraph neural networks — a term used in the literature with several related meanings (e.g., supernetworks that embed many candidate architectures, graph-of-graphs or hypergraph representations where each node is itself a graph, and extensions of hypergraph/hypergraph laplacians) — have already been applied across a remarkably broad set of domains and tasks. In neural architecture search, the differentiable search paradigm treats the search space as a single over-arching hypergraph (or supernetwork) that contains all candidate cells/paths and is optimized end-to-end, enabling efficient architecture optimization for image classification and related tasks \cite{maile2021darts}. In problems that compare or index whole graphs, researchers build a hypergraph of graphs (each node represents a data graph and edges encode pairwise similarities) so that graph-level classification, retrieval, or similarity search can be cast as node classification/lookup on the hypergraph; this idea underpins recent graph-graph similarity networks and neural hypergraph containment/search systems designed for large graph databases and efficient hypergraph queries \cite{yue2022graph, wang2023neural, wang2023neural}. In neuroscience and medical imaging, hypergraph and graph-super-resolution methods use graph neural models to upsample or refine low-resolution brain connectivity maps into higher-resolution connectomes, improving downstream analyses that require fine-grained regional networks \cite{isallari2021brain}. Engineering applications have also adopted hypergraph-style architectures: spatial–temporal hypergraph feature extractors and multi-scale hypergraph convolutional modules have been proposed for rotating machinery fault diagnosis and structural health monitoring, where mapping multi-sensor signals into graphs and then into hypergraph representations yields robust features for anomaly detection \cite{+5, lu2022feature}. Additionally, the interactive concept of supergraph neural networks can be applied to integrate multi-modal and multi-feature scenarios in single-cell multi-omics \cite{zhou2024schiclassifier, yang2024sccross}. In image processing, graph-based and hypergraph-inspired GNNs support multi-image and cross-scale super-resolution / restoration tasks by modeling nonlocal self-similarities across patches as graph nodes and using graph message passing to aggregate complementary information from multiple observations \cite{zhou2020cross}. Recommender systems and social recommendation have benefited from hypergraph and hypergraph constructions (often implemented via HGNN \cite{+5} variants) that explicitly model multi-party interactions, social groups, and multi-view modalities to better capture group influence and high-order affinities; empirical studies report improved recommendation and robustness when hypergraph/hypergraph attention mechanisms are used \cite{xia2021retracted, yu2024meta}. From a systems and scalability perspective, the hypergraph abstraction is useful for distributed or federated GNN training: constructing a hypergraph to summarize or index many local graphs (or to cluster nodes/partitions) can reduce communication and speed up training on massive graph collections \cite{zhu2025simplifying}. Beyond these concrete applications, surveys and reviews of hypergraph and hypergraph methodologies highlight how hypergraph laplacians, hypergraph generalizations, and hypergraph kernels provide principled theoretical foundations that have catalyzed diverse applications in multimodal fusion, biological network analysis, document and text classification, and knowledge-graph style queries; these reviews also point to ongoing research on efficiency, attention-based weighting of hyperedges, and multiset / AllSet \cite{chien2021allset} style propagation rules that extend the expressive power of conventional GNNs \cite{zhi2024review, bai2025hyperbolic}. Finally, recent work on neural similarity search, graph coding for efficient hypergraph search, and specialized index structures shows that hypergraph neural methods are not only beneficial for predictive modeling but also for practical graph-database operations such as containment search and large-scale retrieval, making them attractive for industrial graph-mining, cheminformatics, and program analysis pipelines\cite{wang2023neural, imai2020efficient, chang2025neural}.

\subsection{Formalizing High-order Knowledge}

This subsection gives a concise formal definition of high-order knowledge, states two propositions that make precise why hypergraph-derived features are strictly richer than 1-hop (pairwise) features for many controllability/robustness targets, and explains how these results connect to the Dual HGNN formulas \ref{eq:eq11}-\ref{eq:eq16}.

\subsubsection{Notation}

Let $G = (V, E)$ be a simple graph with $|V| = n$ nodes and adjacency matrix $A \in \{0, 1\}^{n \times m}$. For an integer $K \ge 1$ denote the set of walk matrices $\{A, A^2, \dots, A^K\}$. For a hypergraph constructed on $V$ with $m$ hyperedges, let $H \in \{0, 1\}^{n \times m}$ be its node-hyperedge incidence matrix and let $W \in \mathbb{R}^{m \times m}$ be a diagonal matrix of hyperedge weights. Define the node-projection matrix $B := HWH^T \in \mathbb{R}^{n \times n}$. When needed, we write entrywise relations as $B_{ij}$. We denote by $\mathcal{A}_K$ the algebra generated by the matrices $\{I, A, A^2, \dots, A^K\}$ under linear combinations and (optional) elementwise boolean thresholding.

\subsubsection{High-order knowledge Definition}

A feature object (matrix/tensor/collection) $\mathcal{H}$ computed from $G$ is said to encode \textbf{high-order knowledge of order $K$} if $\mathcal{H}$ depends nontrivially on walk information of length up to $K$; equivalently, 
\begin{itemize}
    \item $\mathcal{H}$ lies in the algebra generated by $\{A, A^2, \dots, A^K\}$ but not in the subalgebra generated by $\{I, A\}$ alone, or 
    \item $\mathcal{H}$ is produced by a hypergraph incidence $H$ whose construction uses $K$-ball / $K$-step relations (e.g., K-Hop hyperedges) or a kernel/diffusion $\Psi(A)$ incorporating powers up to degree $K$ (e.g., embeddings derived from spectral/propagation maps).
\end{itemize}
Intuitively, high-order refers to information that cannot be recovered from only immediate-neighbor (1-hop) relations.

\subsubsection{Propositions}

\begin{figure}[!ht]
\centering     
\subfigure[]{\includegraphics[width=60mm]{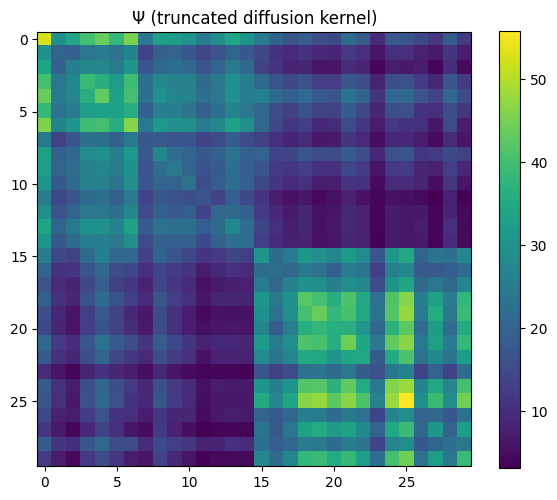}}
\subfigure[]{\includegraphics[width=60mm]{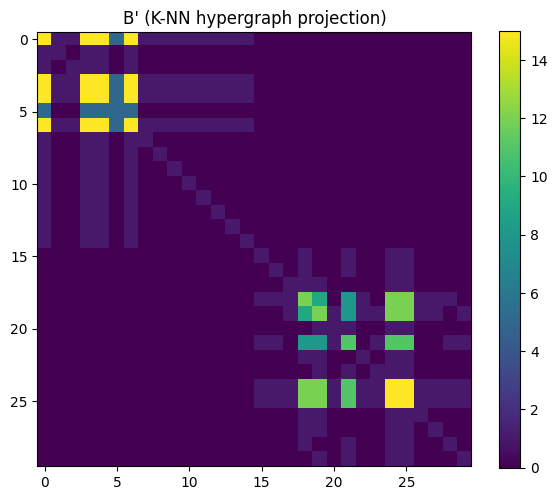}}
\caption{Visualization results of the corresponding matrix to compare large entries of $\psi$ and $B'$ (with top pairs overlapping).}
\label{fig11}
\end{figure}

We state two propositions that formalize the intuition that K-Hop and K-NN hypergraphs (as used in this work) encode strictly more information than 1-hop adjacency for many targets of interest.

i) \textit{Proposition 1 (K-Hop hypergraph sufficiency)}. Let $HG_{K\text{-Hop}}$ be the hypergraph constructed by taking, for each node $v \in V$, the K-ball $\mathcal{N}_K(v) := \{u : \exists k \le K, (A^k)_{uv} > 0\}$ as a hyperedge (or by using a collection of such balls), and let $H$ be its incidence matrix. Then the node-projection matrix $B = HWH^T$ is a function of the family $\{A, A^2, \dots, A^K\}$. Consequently, under (Assum1) there exists a map $\mathcal{F}$ (possibly nonlinear) such that $R(G) = g(\sum_{k=1}^K \alpha_k A^k) = g(\mathcal{F}(B))$. Hence any predictor that has access to $B$ (or learnable transforms of $B$) can in principle recover the dependence of $R(G')$ expressed in (Assum1).

\textit{Proof (sketch)}. For each center $v$ define the indicator vector $\mathbf{1}_{\mathcal{N}_K(v)} \in \{0, 1\}^n$ where the $i$-th entry equals 1 iff there exists a walk of length $\le K$ from $v$ to $i$; equivalently this indicator depends only on the booleanized sum $\mathbf{1}(\sum_{k=1}^K A^k > 0)$. If the hyperedge set contains the K-balls (one per center) and weights in $W$ are chosen deterministically (e.g., all ones), then $B = \sum_{v \in V} \mathbf{1}_{\mathcal{N}_K(v)}\mathbf{1}_{\mathcal{N}_K(v)}^T$, which shows that each entry $B_{ij}$ is a count of how many K-balls contain both $i$ and $j$. Therefore every entry of $B$ is a function of the booleanized matrices $\{(\mathbf{1}(A^k > 0))\}_{k \le K}$ (and hence of $\{A^k\}_{k \le K}$ up to thresholding). As the right-hand side in (Assum1) depends only on $\{A^k\}_{k \le K}$, there exists a (possibly many-to-one) map $\mathcal{F}$ from $B$ to $\sum_{k \le K} \alpha_k A^k$ (or to sufficient statistics for $g$). Thus the stated equality follows.

\textit{Remark}. Exact algebraic invertibility (i.e., uniquely recovering $A^k$ from $B$) is not required: it suffices that the information in $B$ contains the sufficient statistics used by $g$. In practice, the Dual HGNN learns $\mathcal{F}$ or an approximation thereof.

ii) \textit{Proposition 2 (K-NN hypergraph and diffusion kernels)}. Let $\bar{\Phi} : V \to \mathbb{R}^d$ be node embeddings such that the pairwise similarity matrix $S$ with entries $S_{ij} = \langle \bar{\Phi}(i), \bar{\Phi}(j) \rangle$ approximates a diffusion kernel of the adjacency, $S \approx \Psi(A) := \sum_{k=0}^{\infty} \beta_k A^k$, with the truncation $\sum_{k=0}^K \beta_k A^k$ dominating contributions up to order $K$. Constructing a K-NN hypergraph by linking each node to its $K_{NN}$ nearest neighbors in embedding space yields a projection matrix $B'$ whose large entries correspond to large entries of $\Psi(A)$. If $R(G)$ depends on structural equivalence or diffusion-based proximity (i.e. on $\Psi(A)$), then $B'$ is informative for predicting $R(G)$ while 1-hop adjacency alone is not.

\textit{Proof (sketch)}. 

Under the approximation $S \approx \Psi(A)$ assume the approximation error is small in operator norm on the subspace of interest. Thresholding or taking the K-NN of $S$ selects pairs with the largest values of $\Psi(A)_{ij}$, which reflect sums of weighted walk counts between nodes up to order $K$. The resulting incidence matrix $H'$ and projection $B' = H'W'H'^T$ therefore encode (thresholded) diffusion information and role-similarity that go beyond immediate adjacency. If $R(G)$ depends on such diffusion or role-based features, a model using $B'$ can access them while a model restricted to functions of $A$ with degree $\le 1$ cannot, concluding the claim.

Besides, we provide some numerical analysis as follows:

Fig. \ref{fig11} (a) $\psi$ represents the node embedding $\psi$ (spectral embedding) generated by extracting the eigenvector of the diffusion kernel with a truncation order of 40. The Fig. \ref{fig11} (b) visualizes the projection matrix $B'$ computed after constructing a hypergraph using K-NN.

The heatmap of $\psi$ displays larger diffusion affinity entries within two communities (two blocks of the matrix). These high values result from the accumulation of paths spanning multiple steps. The heatmap of $B'$ reveals the “co-occurrence” structure preserved after K-NN hypergraph projection: most node pairs frequently co-occurring within the same hyperedge manifest as large values in $B'$. The top-40 overlap rate ($\approx45\%$) indicates that under this setting, the K-NN hypergraph indeed captures a portion of $\psi$'s major entries (particularly node pairs that are significantly close in the embedding space).

\begin{table*}[!t]
\centering
\caption{$BC_{GAT}$ Model Network Structure}
\begin{tabular}{|c|c|c|c|c|}
\hline
\textbf{Layer Name} & \textbf{Layer Type} & \textbf{Input Channels} & \textbf{Output Channels} & \textbf{Number of Parameters} \\
\hline
gat1 & GATConv (4 heads) & 1 & 64 & $1 \times 64 \times 4 + 2 \times 64 \times 4 + 64 \times 4 = \textbf{1024}$ \\
\hline
gat2 & GATConv (1 head) & $64 \times 4 = 256$ & 64 & $256 \times 64 + 2 \times 64 + 64 = \textbf{16576}$ \\
\hline
fc & Fully Connected & 64 & 1 & $64 \times 1 + 1 = \textbf{65}$ \\
\hline
\end{tabular}
\label{tab:table S0}
\end{table*}

\subsubsection{Connection to Dual HGNN}

The Dual HGNN processes hypergraph structure via two alternating steps: node-to-hyperedge aggregation and hyperedge-to-node aggregation (formulas \ref{eq:eq11}-\ref{eq:eq16}). These steps are learnable functions of the incidence matrix $H$ and node/hyperedge features. Concretely, the message passing performed by the Dual HGNN can be written schematically as
\[
X_E \leftarrow f_E(H^T, X), \quad X \leftarrow f_V(H, X_E),
\]
where $X$ and $X_E$ are node and hyperedge feature matrices and each $f$ includes attention weights that are learned (see formulas \ref{eq:eq11}-\ref{eq:eq16}). Because the node-projection matrix $B=HWH^T$ is a sufficient statistic for the high-order relations described above (Propositions 1--2), the Dual HGNN has the representational capacity to learn maps of the form $X_{\text{out}} = \text{NN}(\mathcal{T}(B,X)) \approx \text{NN}(\mathcal{T}'(\{A^k\}_{k \le K}, X))$, for suitable neural transforms $\mathcal{T}, \mathcal{T}'$ and a readout NN. In other words, by attending to hyperedges and aggregating across them the Dual HGNN implements learnable functions of high-order statistics (walk counts, diffusion affinities, and role-similarity) that are relevant to controllability/robustness targets and which are generally inaccessible to models restricted to 1-hop adjacency information.

In summary, under mild and natural assumptions on the dependence of the target on multi-step connectivity, K-Hop and K-NN hypergraphs provide a formal instantiation of high-order knowledge; the Dual HGNN is a principled mechanism to transform these high-order statistics into predictive node embeddings and graph-level summaries.

\subsection{The Overview of $BC_{GAT}$ Model Network Structure}

Table \ref{tab:table S0} shows the network architecture of the $BC_{GAT}$ model.

\subsection{The Experimental Result of Training Split in Section V. D}

The specific experimental results are shown in Table S2.

\begin{table}[h]
\centering
\caption{The mean error $\overline{er}$ and mean standard deviation $\overline{\sigma}$ of the controlled robustness curves of NCR-HoK's prediction for different network topologies are reduced by 20\%, 40\%, 60\%, and 80\% for the training dataset, respectively.}
\label{tab:table S1}
\scalebox{0.9}{
\begin{tabular}{|ccc|cccc|}
\hline
\multicolumn{3}{|c|}{\textbf{RA Attack}}                                                                                                     & \textbf{\textless{}k\textgreater{}=2} & \textbf{\textless{}k\textgreater{}=5} & \multicolumn{1}{c}{\textbf{\textless{}k\textgreater{}=8}} & \textbf{\textless{}k\textgreater{}=10} \\ \hline
\multicolumn{1}{|c|}{}                                     & \multicolumn{1}{c|}{}                               & $\overline{er}$                         & 0.019                                 & 0.015                                 & 0.016                                                      & 0.012                                  \\
\multicolumn{1}{|c|}{}                                     & \multicolumn{1}{c|}{\multirow{-2}{*}{\textbf{ER}}}  & \cellcolor[HTML]{DBDBDB}$\overline{\sigma}$  & \cellcolor[HTML]{DBDBDB}0.014         & \cellcolor[HTML]{DBDBDB}0.011         & \cellcolor[HTML]{DBDBDB}0.010                              & \cellcolor[HTML]{DBDBDB}0.008          \\ \cline{2-2}
\multicolumn{1}{|c|}{}                                     & \multicolumn{1}{c|}{}                               & $\overline{er}$                         & 0.020                                 & 0.022                                 & 0.023                                                      & 0.023                                  \\
\multicolumn{1}{|c|}{}                                     & \multicolumn{1}{c|}{\multirow{-2}{*}{\textbf{SF}}}  & \cellcolor[HTML]{DBDBDB}$\overline{\sigma}$ & \cellcolor[HTML]{DBDBDB}0.016         & \cellcolor[HTML]{DBDBDB}0.016         & \cellcolor[HTML]{DBDBDB}0.017                              & \cellcolor[HTML]{DBDBDB}0.017          \\ \cline{2-2}
\multicolumn{1}{|c|}{}                                     & \multicolumn{1}{c|}{}                               & $\overline{er}$                         & 0.016                                 & 0.018                                 & 0.016                                                      & 0.016                                  \\
\multicolumn{1}{|c|}{}                                     & \multicolumn{1}{c|}{\multirow{-2}{*}{\textbf{QSN}}} & \cellcolor[HTML]{DBDBDB}$\overline{\sigma}$ & \cellcolor[HTML]{DBDBDB}0.012         & \cellcolor[HTML]{DBDBDB}0.012         & \cellcolor[HTML]{DBDBDB}0.011                              & \cellcolor[HTML]{DBDBDB}0.011          \\ \cline{2-2}
\multicolumn{1}{|c|}{}                                     & \multicolumn{1}{c|}{}                               & $\overline{er}$                         & 0.016                                 & 0.015                                 & 0.014                                                      & 0.011                                  \\
\multicolumn{1}{|c|}{\multirow{-8}{*}{\textbf{20\%}}}      & \multicolumn{1}{c|}{\multirow{-2}{*}{\textbf{SW}}}  & \cellcolor[HTML]{DBDBDB}$\overline{\sigma}$ & \cellcolor[HTML]{DBDBDB}0.010         & \cellcolor[HTML]{DBDBDB}0.011         & \cellcolor[HTML]{DBDBDB}0.009                              & \cellcolor[HTML]{DBDBDB}0.008          \\ \hline
\multicolumn{1}{|c|}{}                                     & \multicolumn{1}{c|}{}                               & $\overline{er}$                         & 0.017                                 & 0.015                                 & 0.014                                                      & 0.012                                  \\
\multicolumn{1}{|c|}{}                                     & \multicolumn{1}{c|}{\multirow{-2}{*}{\textbf{ER}}}  & \cellcolor[HTML]{DBDBDB}$\overline{\sigma}$ & \cellcolor[HTML]{DBDBDB}0.013         & \cellcolor[HTML]{DBDBDB}0.011         & \cellcolor[HTML]{DBDBDB}0.009                              & \cellcolor[HTML]{DBDBDB}0.008          \\ \cline{2-2}
\multicolumn{1}{|c|}{}                                     & \multicolumn{1}{c|}{}                               & $\overline{er}$                         & 0.016                                 & 0.019                                 & 0.020                                                      & 0.022                                  \\
\multicolumn{1}{|c|}{}                                     & \multicolumn{1}{c|}{\multirow{-2}{*}{\textbf{SF}}}  & \cellcolor[HTML]{DBDBDB}$\overline{\sigma}$ & \cellcolor[HTML]{DBDBDB}0.012         & \cellcolor[HTML]{DBDBDB}0.015         & \cellcolor[HTML]{DBDBDB}0.015                              & \cellcolor[HTML]{DBDBDB}0.016          \\ \cline{2-2}
\multicolumn{1}{|c|}{}                                     & \multicolumn{1}{c|}{}                               & $\overline{er}$                         & 0.015                                 & 0.016                                 & 0.014                                                      & 0.014                                  \\
\multicolumn{1}{|c|}{}                                     & \multicolumn{1}{c|}{\multirow{-2}{*}{\textbf{QSN}}} & \cellcolor[HTML]{DBDBDB}$\overline{\sigma}$ & \cellcolor[HTML]{DBDBDB}0.011         & \cellcolor[HTML]{DBDBDB}0.012         & \cellcolor[HTML]{DBDBDB}0.010                              & \cellcolor[HTML]{DBDBDB}0.010          \\ \cline{2-2}
\multicolumn{1}{|c|}{}                                     & \multicolumn{1}{c|}{}                               & $\overline{er}$                         & 0.013                                 & 0.014                                 & 0.013                                                      & 0.011                                  \\
\multicolumn{1}{|c|}{\multirow{-8}{*}{\textbf{40\%}}}      & \multicolumn{1}{c|}{\multirow{-2}{*}{\textbf{SW}}}  & \cellcolor[HTML]{DBDBDB}$\overline{\sigma}$ & \cellcolor[HTML]{DBDBDB}0.009         & \cellcolor[HTML]{DBDBDB}0.010         & \cellcolor[HTML]{DBDBDB}0.008                              & \cellcolor[HTML]{DBDBDB}0.008          \\ \hline
\multicolumn{1}{|c|}{}                                     & \multicolumn{1}{c|}{}                               & $\overline{er}$                         & 0.016                                 & 0.015                                 & 0.014                                                      & 0.014                                  \\
\multicolumn{1}{|c|}{}                                     & \multicolumn{1}{c|}{\multirow{-2}{*}{\textbf{ER}}}  & \cellcolor[HTML]{DBDBDB}$\overline{\sigma}$ & \cellcolor[HTML]{DBDBDB}0.012         & \cellcolor[HTML]{DBDBDB}0.011         & \cellcolor[HTML]{DBDBDB}0.009                              & \cellcolor[HTML]{DBDBDB}0.008          \\ \cline{2-2}
\multicolumn{1}{|c|}{}                                     & \multicolumn{1}{c|}{}                               & $\overline{er}$                         & 0.016                                 & 0.019                                 & 0.020                                                      & 0.021                                  \\
\multicolumn{1}{|c|}{}                                     & \multicolumn{1}{c|}{\multirow{-2}{*}{\textbf{SF}}}  & \cellcolor[HTML]{DBDBDB}$\overline{\sigma}$ & \cellcolor[HTML]{DBDBDB}0.012         & \cellcolor[HTML]{DBDBDB}0.014         & \cellcolor[HTML]{DBDBDB}0.015                              & \cellcolor[HTML]{DBDBDB}0.015          \\ \cline{2-2}
\multicolumn{1}{|c|}{}                                     & \multicolumn{1}{c|}{}                               & $\overline{er}$                         & 0.015                                 & 0.015                                 & 0.013                                                      & 0.013                                  \\
\multicolumn{1}{|c|}{}                                     & \multicolumn{1}{c|}{\multirow{-2}{*}{\textbf{QSN}}} & \cellcolor[HTML]{DBDBDB}$\overline{\sigma}$ & \cellcolor[HTML]{DBDBDB}0.011         & \cellcolor[HTML]{DBDBDB}0.011         & \cellcolor[HTML]{DBDBDB}0.010                              & \cellcolor[HTML]{DBDBDB}0.009          \\ \cline{2-2}
\multicolumn{1}{|c|}{}                                     & \multicolumn{1}{c|}{}                               & $\overline{er}$                         & 0.013                                 & 0.015                                 & 0.012                                                      & 0.011                                  \\
\multicolumn{1}{|c|}{\multirow{-8}{*}{\textbf{60\%}}}      & \multicolumn{1}{c|}{\multirow{-2}{*}{\textbf{SW}}}  & \cellcolor[HTML]{DBDBDB}$\overline{\sigma}$ & \cellcolor[HTML]{DBDBDB}0.009         & \cellcolor[HTML]{DBDBDB}0.010         & \cellcolor[HTML]{DBDBDB}0.008                              & \cellcolor[HTML]{DBDBDB}0.007          \\ \hline
\multicolumn{1}{|c|}{}                                     & \multicolumn{1}{c|}{}                               & $\overline{er}$                         & 0.016                                 & 0.014                                 & 0.013                                                      & 0.011                                  \\
\multicolumn{1}{|c|}{}                                     & \multicolumn{1}{c|}{\multirow{-2}{*}{\textbf{ER}}}  & \cellcolor[HTML]{DBDBDB}$\overline{\sigma}$ & \cellcolor[HTML]{DBDBDB}0.012         & \cellcolor[HTML]{DBDBDB}0.011         & \cellcolor[HTML]{DBDBDB}0.009                              & \cellcolor[HTML]{DBDBDB}0.008          \\ \cline{2-2}
\multicolumn{1}{|c|}{}                                     & \multicolumn{1}{c|}{}                               & $\overline{er}$                         & 0.016                                 & 0.019                                 & 0.021                                                      & 0.021                                  \\
\multicolumn{1}{|c|}{}                                     & \multicolumn{1}{c|}{\multirow{-2}{*}{\textbf{SF}}}  & \cellcolor[HTML]{DBDBDB}$\overline{\sigma}$ & \cellcolor[HTML]{DBDBDB}0.012         & \cellcolor[HTML]{DBDBDB}0.015         & \cellcolor[HTML]{DBDBDB}0.015                              & \cellcolor[HTML]{DBDBDB}0.015          \\ \cline{2-2}
\multicolumn{1}{|c|}{}                                     & \multicolumn{1}{c|}{}                               & $\overline{er}$                         & 0.015                                 & 0.015                                 & 0.013                                                      & 0.013                                  \\
\multicolumn{1}{|c|}{}                                     & \multicolumn{1}{c|}{\multirow{-2}{*}{\textbf{QSN}}} & \cellcolor[HTML]{DBDBDB}$\overline{\sigma}$ & \cellcolor[HTML]{DBDBDB}0.011         & \cellcolor[HTML]{DBDBDB}0.011         & \cellcolor[HTML]{DBDBDB}0.010                              & \cellcolor[HTML]{DBDBDB}0.009          \\ \cline{2-2}
\multicolumn{1}{|c|}{}                                     & \multicolumn{1}{c|}{}                               & $\overline{er}$                         & 0.012                                 & 0.014                                 & 0.012                                                      & 0.010                                  \\
\multicolumn{1}{|c|}{\multirow{-8}{*}{\textbf{80\%}}}      & \multicolumn{1}{c|}{\multirow{-2}{*}{\textbf{SW}}}  & \cellcolor[HTML]{DBDBDB}$\overline{\sigma}$ & \cellcolor[HTML]{DBDBDB}0.009         & \cellcolor[HTML]{DBDBDB}0.009         & \cellcolor[HTML]{DBDBDB}0.008                              & \cellcolor[HTML]{DBDBDB}0.007          \\ \hline
\multicolumn{1}{|c|}{}                                     & \multicolumn{1}{c|}{}                               & $\overline{er}$                         & 0.016                                 & 0.015                                 & 0.012                                                      & 0.010                                  \\
\multicolumn{1}{|c|}{}                                     & \multicolumn{1}{c|}{\multirow{-2}{*}{\textbf{ER}}}  & \cellcolor[HTML]{DBDBDB}$\overline{\sigma}$ & \cellcolor[HTML]{DBDBDB}0.012         & \cellcolor[HTML]{DBDBDB}0.011         & \cellcolor[HTML]{DBDBDB}0.009                              & \cellcolor[HTML]{DBDBDB}0.008          \\ \cline{2-2}
\multicolumn{1}{|c|}{}                                     & \multicolumn{1}{c|}{}                               & $\overline{er}$                         & 0.016                                 & 0.019                                 & 0.021                                                      & 0.021                                  \\
\multicolumn{1}{|c|}{}                                     & \multicolumn{1}{c|}{\multirow{-2}{*}{\textbf{SF}}}  & \cellcolor[HTML]{DBDBDB}$\overline{\sigma}$ & \cellcolor[HTML]{DBDBDB}0.012         & \cellcolor[HTML]{DBDBDB}0.015         & \cellcolor[HTML]{DBDBDB}0.015                              & \cellcolor[HTML]{DBDBDB}0.016          \\ \cline{2-2}
\multicolumn{1}{|c|}{}                                     & \multicolumn{1}{c|}{}                               & $\overline{er}$                         & 0.015                                 & 0.015                                 & 0.013                                                      & 0.012                                  \\
\multicolumn{1}{|c|}{}                                     & \multicolumn{1}{c|}{\multirow{-2}{*}{\textbf{QSN}}} & \cellcolor[HTML]{DBDBDB}$\overline{\sigma}$ & \cellcolor[HTML]{DBDBDB}0.011         & \cellcolor[HTML]{DBDBDB}0.011         & \cellcolor[HTML]{DBDBDB}0.009                              & \cellcolor[HTML]{DBDBDB}0.008          \\ \cline{2-2}
\multicolumn{1}{|c|}{}                                     & \multicolumn{1}{c|}{}                               & $\overline{er}$                         & 0.012                                 & 0.014                                 & 0.012                                                      & 0.010                                  \\
\multicolumn{1}{|c|}{\multirow{-8}{*}{\textbf{Full(Raw)}}} & \multicolumn{1}{c|}{\multirow{-2}{*}{\textbf{SW}}}  & \cellcolor[HTML]{DBDBDB}$\overline{\sigma}$ & \cellcolor[HTML]{DBDBDB}0.009         & \cellcolor[HTML]{DBDBDB}0.009         & \cellcolor[HTML]{DBDBDB}0.008                              & \cellcolor[HTML]{DBDBDB}0.007          \\ \hline
\end{tabular}}
\end{table}

\subsection{The Experimental Results of the Choice of K-value} \label{sec: Kvalue}

\subsubsection{$K$-Hop Experiments on the different choice of $K$}

Based on the data in table \ref{tab:table S2}, the choice of $K$=3 (3-Hop) appears to be a practical and well-balanced decision, representing an excellent trade-off between model performance and computational cost. As shown in the table results:
\begin{itemize}
    \item Marginal Performance Gains with Higher $K$: When comparing the results of 3-Hop to 4-Hop, the performance improvements in terms of mean error and standard deviation are minimal to non-existent across all network types. For example, in SW networks with $<k>=10$, both 3-Hop and 4-Hop achieve the same mean error of 0.010 and standard deviation of 0.007. Since a larger K-value increases the computational complexity of constructing the hypergraph, choosing $K$=4 offers no significant accuracy benefit over $K$=3.
    \item Consistent Improvement Over K=2: While 3-Hop doesn't show a large advantage over 4-Hop, it does offer slight but consistent improvements over 2-Hop in several cases. For instance, in QSN networks with $<k>=10$, the mean error drops from 0.013 at 2-Hop to 0.012 at 3-Hop. This suggests that moving from $K$=2 to $K$=3 captures more valuable high-order information.
    \item Identical Performance on SF Networks: For SF networks, the performance is identical across all three $K$-values. In this scenario, selecting the lowest computationally expensive option that still performs well on other network types (like $K$=3) is the most logical choice.
\end{itemize}

In summary, $K$=3 was likely chosen because it hits a sweet spot. It captures more complex network structures than $K$=2, leading to slightly better performance, while avoiding the increased computational cost of $K$=4, which provides no meaningful improvement in accuracy.

\begin{table}[!h]
\centering
\caption{The mean error er and mean standard deviation a of the controlled robustness curves of NCR-HoK's prediction for different network topologies for K-Hop’s K-values of 2, 3, and 4 cases, respectively.}
\label{tab:table S2}
\scalebox{1.0}{
\begin{tabular}{|ccc|cccc|}
\hline
\multicolumn{3}{|c|}{\textbf{RA Attack}}                                                                                                                                                                                     & \multicolumn{1}{c}{\textbf{\textless{}k\textgreater{}=2}} & \multicolumn{1}{c}{\textbf{\textless{}k\textgreater{}=5}} & \multicolumn{1}{c}{\textbf{\textless{}k\textgreater{}=8}} & \multicolumn{1}{c|}{\textbf{\textless{}k\textgreater{}=10}} \\ \hline
\multicolumn{1}{|l|}{}                                                                                       & \multicolumn{1}{l|}{}                               & $\overline{er}$                                                       & 0.016                                                     & 0.014                                                     & 0.012                                                      & 0.011                                                       \\
\multicolumn{1}{|l|}{}                                                                                       & \multicolumn{1}{c|}{\multirow{-2}{*}{\textbf{ER}}}  & \cellcolor[HTML]{DBDBDB}$\overline{\sigma}$                               & \cellcolor[HTML]{DBDBDB}0.013                             & \cellcolor[HTML]{DBDBDB}0.011                             & \cellcolor[HTML]{DBDBDB}0.010                              & \cellcolor[HTML]{DBDBDB}0.008                               \\ \cline{2-2}
\multicolumn{1}{|l|}{}                                                                                       & \multicolumn{1}{l|}{}                               & $\overline{er}$                                                       & 0.016                                                     & 0.019                                                     & 0.021                                                      & 0.021                                                       \\
\multicolumn{1}{|l|}{}                                                                                       & \multicolumn{1}{c|}{\multirow{-2}{*}{\textbf{SF}}}  & \cellcolor[HTML]{DBDBDB}$\overline{\sigma}$                               & \cellcolor[HTML]{DBDBDB}0.013                             & \cellcolor[HTML]{DBDBDB}0.015                             & \cellcolor[HTML]{DBDBDB}0.016                              & \cellcolor[HTML]{DBDBDB}0.016                               \\ \cline{2-2}
\multicolumn{1}{|l|}{}                                                                                       & \multicolumn{1}{l|}{}                               & $\overline{er}$                                                       & 0.015                                                     & 0.015                                                     & 0.013                                                      & 0.013                                                       \\
\multicolumn{1}{|l|}{}                                                                                       & \multicolumn{1}{c|}{\multirow{-2}{*}{\textbf{QSN}}} & \cellcolor[HTML]{DBDBDB}$\overline{\sigma}$                               & \cellcolor[HTML]{DBDBDB}0.011                             & \cellcolor[HTML]{DBDBDB}0.011                             & \cellcolor[HTML]{DBDBDB}0.010                              & \cellcolor[HTML]{DBDBDB}0.009                               \\ \cline{2-2}
\multicolumn{1}{|l|}{}                                                                                       & \multicolumn{1}{l|}{}                               & $\overline{er}$                                                       & 0.012                                                     & 0.015                                                     & 0.012                                                      & 0.010                                                       \\
\multicolumn{1}{|l|}{\multirow{-8}{*}{\textbf{2-Hop}}}                                                       & \multicolumn{1}{c|}{\multirow{-2}{*}{\textbf{SW}}}  & \cellcolor[HTML]{DBDBDB}$\overline{\sigma}$                               & \cellcolor[HTML]{DBDBDB}0.009                             & \cellcolor[HTML]{DBDBDB}0.010                             & \cellcolor[HTML]{DBDBDB}0.009                              & \cellcolor[HTML]{DBDBDB}0.008                               \\ \hline
\multicolumn{1}{|l|}{}                                                                                       & \multicolumn{1}{l|}{}                               & $\overline{er}$                                                       & 0.016                                                     & 0.015                                                     & 0.012                                                      & 0.010                                                       \\
\multicolumn{1}{|l|}{}                                                                                       & \multicolumn{1}{c|}{\multirow{-2}{*}{\textbf{ER}}}  & \cellcolor[HTML]{DBDBDB}$\overline{\sigma}$                               & \cellcolor[HTML]{DBDBDB}0.012                             & \cellcolor[HTML]{DBDBDB}0.011                             & \cellcolor[HTML]{DBDBDB}0.009                              & \cellcolor[HTML]{DBDBDB}0.008                               \\ \cline{2-2}
\multicolumn{1}{|l|}{}                                                                                       & \multicolumn{1}{l|}{}                               & $\overline{er}$                                                       & 0.016                                                     & 0.019                                                     & 0.021                                                      & 0.021                                                       \\
\multicolumn{1}{|l|}{}                                                                                       & \multicolumn{1}{c|}{\multirow{-2}{*}{\textbf{SF}}}  & \cellcolor[HTML]{DBDBDB}$\overline{\sigma}$                               & \cellcolor[HTML]{DBDBDB}0.012                             & \cellcolor[HTML]{DBDBDB}0.015                             & \cellcolor[HTML]{DBDBDB}0.015                              & \cellcolor[HTML]{DBDBDB}0.016                               \\ \cline{2-2}
\multicolumn{1}{|l|}{}                                                                                       & \multicolumn{1}{l|}{}                               & $\overline{er}$                                                       & 0.015                                                     & 0.015                                                     & 0.013                                                      & 0.012                                                       \\
\multicolumn{1}{|l|}{}                                                                                       & \multicolumn{1}{c|}{\multirow{-2}{*}{\textbf{QSN}}} & \cellcolor[HTML]{DBDBDB}$\overline{\sigma}$                               & \cellcolor[HTML]{DBDBDB}0.011                             & \cellcolor[HTML]{DBDBDB}0.011                             & \cellcolor[HTML]{DBDBDB}0.009                              & \cellcolor[HTML]{DBDBDB}0.008                               \\ \cline{2-2}
\multicolumn{1}{|l|}{}                                                                                       & \multicolumn{1}{l|}{}                               & $\overline{er}$                                                       & 0.012                                                     & 0.014                                                     & 0.012                                                      & 0.010                                                       \\
\multicolumn{1}{|l|}{\multirow{-8}{*}{\textbf{\begin{tabular}[c]{@{}l@{}}3-Hop\\    \\ (Raw)\end{tabular}}}} & \multicolumn{1}{l|}{\multirow{-2}{*}{\textbf{SW}}}  & \cellcolor[HTML]{DBDBDB}$\overline{\sigma}$                               & \cellcolor[HTML]{DBDBDB}0.009                             & \cellcolor[HTML]{DBDBDB}0.009                             & \cellcolor[HTML]{DBDBDB}0.008                              & \cellcolor[HTML]{DBDBDB}0.007                               \\ \hline
\multicolumn{1}{|l|}{}                                                                                       & \multicolumn{1}{l|}{}                               & $\overline{er}$                                                       & 0.016                                                     & 0.015                                                     & 0.012                                                      & 0.011                                                       \\
\multicolumn{1}{|l|}{}                                                                                       & \multicolumn{1}{c|}{\multirow{-2}{*}{\textbf{ER}}}  & \multicolumn{1}{l|}{\cellcolor[HTML]{DBDBDB}$\overline{\sigma}$} & \cellcolor[HTML]{DBDBDB}0.012                             & \cellcolor[HTML]{DBDBDB}0.011                             & \cellcolor[HTML]{DBDBDB}0.009                              & \cellcolor[HTML]{DBDBDB}0.008                               \\ \cline{2-2}
\multicolumn{1}{|l|}{}                                                                                       & \multicolumn{1}{l|}{}                               & $\overline{er}$                                                       & 0.016                                                     & 0.019                                                     & 0.021                                                      & 0.021                                                       \\
\multicolumn{1}{|l|}{}                                                                                       & \multicolumn{1}{c|}{\multirow{-2}{*}{\textbf{SF}}}  & \multicolumn{1}{l|}{\cellcolor[HTML]{DBDBDB}$\overline{\sigma}$} & \cellcolor[HTML]{DBDBDB}0.012                             & \cellcolor[HTML]{DBDBDB}0.015                             & \cellcolor[HTML]{DBDBDB}0.016                              & \cellcolor[HTML]{DBDBDB}0.016                               \\ \cline{2-2}
\multicolumn{1}{|l|}{}                                                                                       & \multicolumn{1}{l|}{}                               & $\overline{er}$                                                       & 0.015                                                     & 0.014                                                     & 0.013                                                      & 0.013                                                       \\
\multicolumn{1}{|l|}{}                                                                                       & \multicolumn{1}{c|}{\multirow{-2}{*}{\textbf{QSN}}} & \multicolumn{1}{l|}{\cellcolor[HTML]{DBDBDB}$\overline{\sigma}$} & \cellcolor[HTML]{DBDBDB}0.011                             & \cellcolor[HTML]{DBDBDB}0.010                             & \cellcolor[HTML]{DBDBDB}0.009                              & \cellcolor[HTML]{DBDBDB}0.009                               \\ \cline{2-2}
\multicolumn{1}{|l|}{}                                                                                       & \multicolumn{1}{l|}{}                               & $\overline{er}$                                                       & 0.013                                                     & 0.014                                                     & 0.012                                                      & 0.010                                                       \\
\multicolumn{1}{|l|}{\multirow{-8}{*}{\textbf{4-Hop}}}                                                       & \multicolumn{1}{c|}{\multirow{-2}{*}{\textbf{SW}}}  & \cellcolor[HTML]{DBDBDB}$\overline{\sigma}$                               & \cellcolor[HTML]{DBDBDB}0.009                             & \cellcolor[HTML]{DBDBDB}0.009                             & \cellcolor[HTML]{DBDBDB}0.008                              & \cellcolor[HTML]{DBDBDB}0.007                               \\ \hline
\end{tabular}}
\end{table}

\subsubsection{$K$-NN Experiments on the different choice of $K$}

Based on the data in table \ref{tab:table S3}, the choice of $K$=10 (10-NN) is a reasonable and justified default because the model's performance shows almost no sensitivity to the K-value in the K-NN hypergraph construction. As shown in the table results:
\begin{itemize}
    \item Consistent Performance Across All K-Values: When comparing the results for 5-NN, 10-NN, 20-NN, and 30-NN, the mean error and standard deviation are remarkably stable across all network types and average degrees. For example: 
    \begin{itemize}
        \item For SF networks with an average degree of $<k>=8$, the mean error is consistently 0.021 and the standard deviation is 0.015, regardless of whether $K$ is 5, 10, 20, or 30.
        \item For SW networks with $<k>=10$, both the mean error (0.010) and standard deviation (0.007) are identical across all four K-NN settings.
    \end{itemize}
    \item No Advantage when Increasing $K$: Unlike the $K$-Hop analysis where a larger $K$ sometimes offered marginal benefits, here there is no discernible improvement from increasing $K$ from 10 to 20 or 30. Since a larger $K$-value increases the density of the hypergraph and thus the computational cost, choosing a higher value offers no practical benefit.
\end{itemize}

In summary, because the model's prediction accuracy is insensitive to this specific hyperparameter, choosing $K$=10 serves as a sensible middle-ground. It's large enough to capture local neighborhood structures in the embedding space but not so large as to add unnecessary computational complexity for no performance gain.

\begin{table}[!h]
\centering
\caption{The mean error er and mean standard deviation a of the controlled robustness curves of NCR-HoK's prediction for different network topologies for K-NN’s K-values of 5, 10, 20 and 30 cases, respectively.}
\label{tab:table S3}
\scalebox{1.0}{
\begin{tabular}{|ccc|cccc|}
\hline
\multicolumn{3}{|c|}{\textbf{RA Attack}}                                                                                                                                                                                     & \multicolumn{1}{c}{\textbf{\textless{}k\textgreater{}=2}} & \multicolumn{1}{c}{\textbf{\textless{}k\textgreater{}=5}} & \multicolumn{1}{c}{\textbf{\textless{}k\textgreater{}=8}} & \multicolumn{1}{c|}{\textbf{\textless{}k\textgreater{}=10}} \\ \hline
\multicolumn{1}{|l|}{}                                                                                       & \multicolumn{1}{l|}{}                               & $\overline{er}$                                                       & 0.016                                                     & 0.014                                                     & 0.012                                                     & 0.010                                 \\
\multicolumn{1}{|l|}{}                                                                                       & \multicolumn{1}{c|}{\multirow{-2}{*}{\textbf{ER}}}  & \cellcolor[HTML]{DBDBDB}$\overline{\sigma}$                               & \cellcolor[HTML]{DBDBDB}0.012                             & \cellcolor[HTML]{DBDBDB}0.011                             & \cellcolor[HTML]{DBDBDB}0.009                             & \cellcolor[HTML]{DBDBDB}0.008         \\ \cline{2-2}
\multicolumn{1}{|l|}{}                                                                                       & \multicolumn{1}{l|}{}                               & $\overline{er}$                                                       & 0.016                                                     & 0.019                                                     & 0.021                                                     & 0.021                                 \\
\multicolumn{1}{|l|}{}                                                                                       & \multicolumn{1}{c|}{\multirow{-2}{*}{\textbf{SF}}}  & \cellcolor[HTML]{DBDBDB}$\overline{\sigma}$                               & \cellcolor[HTML]{DBDBDB}0.012                             & \cellcolor[HTML]{DBDBDB}0.015                             & \cellcolor[HTML]{DBDBDB}0.016                             & \cellcolor[HTML]{DBDBDB}0.015          \\ \cline{2-2}
\multicolumn{1}{|l|}{}                                                                                       & \multicolumn{1}{l|}{}                               & $\overline{er}$                                                      & 0.015                                                     & 0.014                                                     & 0.013                                                     & 0.012                                 \\
\multicolumn{1}{|l|}{}                                                                                       & \multicolumn{1}{c|}{\multirow{-2}{*}{\textbf{QSN}}} & \cellcolor[HTML]{DBDBDB}$\overline{\sigma}$                               & \cellcolor[HTML]{DBDBDB}0.011                             & \cellcolor[HTML]{DBDBDB}0.011                             & \cellcolor[HTML]{DBDBDB}0.010                             & \cellcolor[HTML]{DBDBDB}0.009         \\ \cline{2-2}
\multicolumn{1}{|l|}{}                                                                                       & \multicolumn{1}{l|}{}                               & $\overline{er}$                                                       & 0.012                                                     & 0.014                                                     & 0.011                                                     & 0.010                                  \\
\multicolumn{1}{|l|}{\multirow{-8}{*}{\textbf{5-NN}}}                                                       & \multicolumn{1}{c|}{\multirow{-2}{*}{\textbf{SW}}}  & \cellcolor[HTML]{DBDBDB}$\overline{\sigma}$                               & \cellcolor[HTML]{DBDBDB}0.010                             & \cellcolor[HTML]{DBDBDB}0.010                             & \cellcolor[HTML]{DBDBDB}0.009                             & \cellcolor[HTML]{DBDBDB}0.007          \\ \hline
\multicolumn{1}{|l|}{}                                                                                       & \multicolumn{1}{l|}{}                               & $\overline{er}$                                                       & 0.016                                                     & 0.015                                                     & 0.012                                                     & 0.010                                  \\
\multicolumn{1}{|l|}{}                                                                                       & \multicolumn{1}{c|}{\multirow{-2}{*}{\textbf{ER}}}  & \cellcolor[HTML]{DBDBDB}$\overline{\sigma}$                               & \cellcolor[HTML]{DBDBDB}0.012                             & \cellcolor[HTML]{DBDBDB}0.011                             & \cellcolor[HTML]{DBDBDB}0.009                             & \cellcolor[HTML]{DBDBDB}0.008         \\ \cline{2-2}
\multicolumn{1}{|l|}{}                                                                                       & \multicolumn{1}{l|}{}                               & $\overline{er}$                                                       & 0.016                                                     & 0.019                                                     & 0.021                                                     & 0.021                                 \\
\multicolumn{1}{|l|}{}                                                                                       & \multicolumn{1}{c|}{\multirow{-2}{*}{\textbf{SF}}}  & \cellcolor[HTML]{DBDBDB}$\overline{\sigma}$                               & \cellcolor[HTML]{DBDBDB}0.012                             & \cellcolor[HTML]{DBDBDB}0.015                             & \cellcolor[HTML]{DBDBDB}0.015                             & \cellcolor[HTML]{DBDBDB}0.016          \\ \cline{2-2}
\multicolumn{1}{|l|}{}                                                                                       & \multicolumn{1}{l|}{}                               & $\overline{er}$                                                       & 0.015                                                     & 0.015                                                     & 0.013                                                     & 0.012                                \\
\multicolumn{1}{|l|}{}                                                                                       & \multicolumn{1}{c|}{\multirow{-2}{*}{\textbf{QSN}}} & \cellcolor[HTML]{DBDBDB}$\overline{\sigma}$                               & \cellcolor[HTML]{DBDBDB}0.011                             & \cellcolor[HTML]{DBDBDB}0.011                             & \cellcolor[HTML]{DBDBDB}0.009                             & \cellcolor[HTML]{DBDBDB}0.008          \\ \cline{2-2}
\multicolumn{1}{|l|}{}                                                                                       & \multicolumn{1}{l|}{}                               & $\overline{er}$                                                       & 0.012                                                     & 0.014                                                     & 0.012                                                     & 0.010                                \\
\multicolumn{1}{|l|}{\multirow{-8}{*}{\textbf{\begin{tabular}[c]{@{}l@{}}10-NN\\    \\ (Raw)\end{tabular}}}} & \multicolumn{1}{c|}{\multirow{-2}{*}{\textbf{SW}}}  & \cellcolor[HTML]{DBDBDB}$\overline{\sigma}$                               & \cellcolor[HTML]{DBDBDB}0.009                             & \cellcolor[HTML]{DBDBDB}0.009                             & \cellcolor[HTML]{DBDBDB}0.008                             & \cellcolor[HTML]{DBDBDB}0.007          \\ \hline
\multicolumn{1}{|l|}{}                                                                                       & \multicolumn{1}{l|}{}                               & $\overline{er}$                                                       & 0.016                                                     & 0.014                                                     & 0.013                                                     & 0.012                                  \\
\multicolumn{1}{|l|}{}                                                                                       & \multicolumn{1}{c|}{\multirow{-2}{*}{\textbf{ER}}}  & \multicolumn{1}{l|}{\cellcolor[HTML]{DBDBDB}$\overline{\sigma}$} & \cellcolor[HTML]{DBDBDB}0.012                             & \cellcolor[HTML]{DBDBDB}0.011                             & \cellcolor[HTML]{DBDBDB}0.010                             & \cellcolor[HTML]{DBDBDB}0.008          \\ \cline{2-2}
\multicolumn{1}{|l|}{}                                                                                       & \multicolumn{1}{l|}{}                               & $\overline{er}$                                                       & 0.016                                                     & 0.019                                                     & 0.021                                                     & 0.021                                \\
\multicolumn{1}{|l|}{}                                                                                       & \multicolumn{1}{c|}{\multirow{-2}{*}{\textbf{SF}}}  & \multicolumn{1}{l|}{\cellcolor[HTML]{DBDBDB}$\overline{\sigma}$} & \cellcolor[HTML]{DBDBDB}0.013                             & \cellcolor[HTML]{DBDBDB}0.015                             & \cellcolor[HTML]{DBDBDB}0.016                             & \cellcolor[HTML]{DBDBDB}0.016          \\ \cline{2-2}
\multicolumn{1}{|l|}{}                                                                                       & \multicolumn{1}{l|}{}                               & $\overline{er}$                                                       & 0.015                                                     & 0.015                                                     & 0.013                                                     & 0.013                                 \\
\multicolumn{1}{|l|}{}                                                                                       & \multicolumn{1}{c|}{\multirow{-2}{*}{\textbf{QSN}}} & \multicolumn{1}{l|}{\cellcolor[HTML]{DBDBDB}$\overline{\sigma}$} & \cellcolor[HTML]{DBDBDB}0.011                             & \cellcolor[HTML]{DBDBDB}0.011                             & \cellcolor[HTML]{DBDBDB}0.010                             & \cellcolor[HTML]{DBDBDB}0.009         \\ \cline{2-2}
\multicolumn{1}{|l|}{}      & \multicolumn{1}{l|}{}          & $\overline{er}$               & 0.012             & 0.015              & 0.012             & 0.011      \\
\multicolumn{1}{|l|}{\multirow{-8}{*}{\textbf{20-NN}}}                                                       & \multicolumn{1}{c|}{\multirow{-2}{*}{\textbf{SW}}}  & \cellcolor[HTML]{DBDBDB}$\overline{\sigma}$                               & \cellcolor[HTML]{DBDBDB}0.009                             & \cellcolor[HTML]{DBDBDB}0.010                             & \cellcolor[HTML]{DBDBDB}0.008                             & \cellcolor[HTML]{DBDBDB}0.008          \\ \hline
\multicolumn{1}{|l|}{}                                                                                       & \multicolumn{1}{l|}{}                               & \multicolumn{1}{l|}{er}                                 & 0.016                                                     & 0.015                                                     & 0.012                                                     & 0.010                               \\
\multicolumn{1}{|l|}{}                                                                                       & \multicolumn{1}{c|}{\multirow{-2}{*}{\textbf{ER}}}  & \multicolumn{1}{l|}{\cellcolor[HTML]{DBDBDB}$\overline{\sigma}$} & \cellcolor[HTML]{DBDBDB}0.012                             & \cellcolor[HTML]{DBDBDB}0.011                             & \cellcolor[HTML]{DBDBDB}0.009                             & \cellcolor[HTML]{DBDBDB}0.008         \\ \cline{2-2}
\multicolumn{1}{|l|}{}                                                                                       & \multicolumn{1}{l|}{}                               & $\overline{er}$                                                       & 0.016                                                     & 0.019                                                     & 0.021                                                     & 0.021                                  \\
\multicolumn{1}{|l|}{}                                                                                       & \multicolumn{1}{c|}{\multirow{-2}{*}{\textbf{SF}}}  & \multicolumn{1}{l|}{\cellcolor[HTML]{DBDBDB}$\overline{\sigma}$} & \cellcolor[HTML]{DBDBDB}0.013                             & \cellcolor[HTML]{DBDBDB}0.015                             & \cellcolor[HTML]{DBDBDB}0.016                             & \cellcolor[HTML]{DBDBDB}0.016         \\ \cline{2-2}
\multicolumn{1}{|l|}{}                                                                                       & \multicolumn{1}{l|}{}                               & $\overline{er}$                                                       & 0.015                                                     & 0.014                                                     & 0.013                                                     & 0.012                                  \\
\multicolumn{1}{|l|}{}                                                                                       & \multicolumn{1}{c|}{\multirow{-2}{*}{\textbf{QSN}}} & \multicolumn{1}{l|}{\cellcolor[HTML]{DBDBDB}$\overline{\sigma}$} & \cellcolor[HTML]{DBDBDB}0.011                             & \cellcolor[HTML]{DBDBDB}0.011                             & \cellcolor[HTML]{DBDBDB}0.010                             & \cellcolor[HTML]{DBDBDB}0.009         \\ \cline{2-2}
\multicolumn{1}{|l|}{}                                                                                       & \multicolumn{1}{l|}{}                               & $\overline{er}$                                                       & 0.013                                                     & 0.014                                                     & 0.012                                                     & 0.010                                \\
\multicolumn{1}{|l|}{\multirow{-8}{*}{\textbf{30-NN}}}                                                       & \multicolumn{1}{c|}{\multirow{-2}{*}{\textbf{SW}}}  & \cellcolor[HTML]{DBDBDB}$\overline{\sigma}$                               & \cellcolor[HTML]{DBDBDB}0.009                             & \cellcolor[HTML]{DBDBDB}0.010                           & \cellcolor[HTML]{DBDBDB}0.009                             & \cellcolor[HTML]{DBDBDB}0.007  \\ \hline
\end{tabular}}
\end{table}

\subsection{Analysis on More Different Attack Scenarios} \label{sec: TBA TDA}

We supplemented our analysis with Target Degree Based Attacks (TDA)  and Target Betweenness Based Attacks (TBA) results. The experimental results, presented in tables \ref{tab:table td} and \ref{tab:table tb}, demonstrate that NCR-HoK maintains a significant advantage under both attack scenarios, further highlighting the robustness of our approach.

\subsubsection{Performance under TDA} As shown in table \ref{tab:table td}, our model achieves the top rank (\#1) in overall average performance for every network type (ER, SF, QSN, and SW) under Target Degree Based Attacks. For instance, on SF networks, NCR-HoK's average error is 0.010, which is substantially lower than the next best model, CRL-SGNN (0.035). Furthermore, our model consistently records the lowest average standard deviation, indicating that its predictions are not only accurate but also highly stable.

\subsubsection{Performance under TBA} The results under Target Betweenness Based Attacks, detailed in table \ref{tab:table tb}, reinforce this conclusion. Our model again secures the top rank (\#1) for overall average error on every network, where it still achieves a competitive first-place rank (\#1).

{
\setlength{\tabcolsep}{3pt}
\begin{table}[!h]
\centering
\caption{The average errors $\overline{er}$ and average standard deviations $\overline{\sigma}$ of the controllable robustness curve predicting for various types of networks under the TDA condition are compared among the NCR-HoK, PCR\cite{27}, iPCR\cite{28} models and CRL-SGNN\cite{CRL-SGNN} , along with a comprehensive ranking.}
\label{tab:table td}
\scalebox{0.9}{
\begin{tabular}{|ccc|cccc|c|}
\hline
\multicolumn{3}{|c|}{}    & \textbf{\textless{}k\textgreater{}=2}  & \textbf{\textless{}k\textgreater{}=5}  & \textbf{\textless{}k\textgreater{}=8}  & \textbf{\textless{}k\textgreater{}=10} & \textbf{Average}    \\ \hline
\multicolumn{1}{|c|}{}   & \multicolumn{1}{l|}{}    & $\overline{er}$    & 0.039    & 0.040  & 0.018  & 0.013    & 0.028(\#3)    \\
\multicolumn{1}{|c|}{}   & \multicolumn{1}{c|}{\multirow{-2}{*}{\textbf{PCR}}}  & \cellcolor[HTML]{D9D9D9}$\overline{\sigma}$ & \cellcolor[HTML]{D9D9D9}0.017   & \cellcolor[HTML]{D9D9D9}0.016          & \cellcolor[HTML]{D9D9D9}0.010          & \cellcolor[HTML]{D9D9D9}0.008          & \cellcolor[HTML]{D9D9D9}0.013(\#2)          \\ \cline{2-2}

\multicolumn{1}{|c|}{}                               & \multicolumn{1}{l|}{}                                & $\overline{er}$                         &    0.053                               &  0.004                                 & 0.008                                  &  0.006                                 & 0.018(\#2)                                  \\
\multicolumn{1}{|c|}{}                               & \multicolumn{1}{c|}{\multirow{-2}{*}{\textbf{iPCR}}} & \cellcolor[HTML]{D9D9D9}$\overline{\sigma}$ & \cellcolor[HTML]{D9D9D9}0.070          & \cellcolor[HTML]{D9D9D9}0.004          & \cellcolor[HTML]{D9D9D9}0.007          & \cellcolor[HTML]{D9D9D9}0.006          & \cellcolor[HTML]{D9D9D9}0.022(\#3)          \\ \cline{2-2}

\multicolumn{1}{|c|}{}                               & \multicolumn{1}{l|}{}                                & $\overline{er}$                         &   0.013                                &   0.070                                 & 0.019                                   & 0.049                                & 0.038(\#4)                                  \\
\multicolumn{1}{|c|}{}                               & \multicolumn{1}{c|}{\multirow{-2}{*}{\textbf{CRL-SGNN}}} & \cellcolor[HTML]{D9D9D9}$\overline{\sigma}$ & \cellcolor[HTML]{D9D9D9}0.014          & \cellcolor[HTML]{D9D9D9}0.047          & \cellcolor[HTML]{D9D9D9}0.022          & \cellcolor[HTML]{D9D9D9}0.051         & \cellcolor[HTML]{D9D9D9}0.034(\#4)          \\ \cline{2-2}

\multicolumn{1}{|c|}{}                               & \multicolumn{1}{l|}{}                                & $\overline{er}$                         & 0.011                        &  0.020                        &  0.012             &  0.011                 & \textbf{0.014(\#1)}                        \\
\multicolumn{1}{|c|}{\multirow{-8}{*}{\textbf{ER}}}  & \multicolumn{1}{c|}{\multirow{-2}{*}{\textbf{Our}}}  & \cellcolor[HTML]{D9D9D9}$\overline{\sigma}$ & \cellcolor[HTML]{D9D9D9}0.007 & \cellcolor[HTML]{D9D9D9}0.011 & \cellcolor[HTML]{D9D9D9}0.007 & \cellcolor[HTML]{D9D9D9}0.007 & \cellcolor[HTML]{D9D9D9}\textbf{0.008(\#1)} \\ \hline

\multicolumn{1}{|l|}{}                               & \multicolumn{1}{l|}{}                                & $\overline{er}$                         &   0.073                               &  0.061                               &  0.047                       &    0.042                              &  0.056(\#4)                                \\
\multicolumn{1}{|l|}{}                               & \multicolumn{1}{c|}{\multirow{-2}{*}{\textbf{PCR}}}  & \cellcolor[HTML]{D9D9D9}$\overline{\sigma}$ & \cellcolor[HTML]{D9D9D9}0.004        & \cellcolor[HTML]{D9D9D9}0.008      & \cellcolor[HTML]{D9D9D9}0.009         & \cellcolor[HTML]{D9D9D9}0.010        & \cellcolor[HTML]{D9D9D9}0.008(\#2)          \\ \cline{2-2}
\multicolumn{1}{|l|}{}                               & \multicolumn{1}{l|}{}                                & $\overline{er}$                         & 0.010          &  0.182               &  0.012               & 0.008                    & 0.053(\#3)                                  \\
\multicolumn{1}{|l|}{}                               & \multicolumn{1}{c|}{\multirow{-2}{*}{\textbf{iPCR}}} & \cellcolor[HTML]{D9D9D9}$\overline{\sigma}$ & \cellcolor[HTML]{D9D9D9}0.009          & \cellcolor[HTML]{D9D9D9}0.198        & \cellcolor[HTML]{D9D9D9}0.014          & \cellcolor[HTML]{D9D9D9}0.010        & \cellcolor[HTML]{D9D9D9}0.058(\#4)          \\ \cline{2-2}
\multicolumn{1}{|l|}{}                               & \multicolumn{1}{l|}{}                                & $\overline{er}$                         & 0.026                   &  0.033               & 0.025               &  0.057                     & 0.035(\#2)                         \\

\multicolumn{1}{|l|}{}                               & \multicolumn{1}{c|}{\multirow{-2}{*}{\textbf{CRL-SGNN}}} & \cellcolor[HTML]{D9D9D9}$\overline{\sigma}$ & \cellcolor[HTML]{D9D9D9}0.032         & \cellcolor[HTML]{D9D9D9}0.049        & \cellcolor[HTML]{D9D9D9}0.035         & \cellcolor[HTML]{D9D9D9}0.082       & \cellcolor[HTML]{D9D9D9}0.050(\#2)          \\ \cline{2-2}
\multicolumn{1}{|l|}{}                               & \multicolumn{1}{l|}{}                                & $\overline{er}$                         &  0.005                       & 0.010                    & 0.011                       & 0.013                     & \textbf{0.010(\#1)}                         \\

\multicolumn{1}{|c|}{\multirow{-8}{*}{\textbf{SF}}}  & \multicolumn{1}{c|}{\multirow{-2}{*}{\textbf{Our}}}  & \cellcolor[HTML]{D9D9D9}$\overline{\sigma}$ & \cellcolor[HTML]{D9D9D9}0.003    & \cellcolor[HTML]{D9D9D9}0.005     & \cellcolor[HTML]{D9D9D9}0.006     & \cellcolor[HTML]{D9D9D9}0.007     & \cellcolor[HTML]{D9D9D9}\textbf{0.005(\#1)} \\ \hline
\multicolumn{1}{|l|}{}                               & \multicolumn{1}{l|}{}                                & $\overline{er}$                         &   0.017                      &   0.021                                &   0.017                      & 0.017                         & 0.018(\#2)                                  \\
\multicolumn{1}{|l|}{}                               & \multicolumn{1}{c|}{\multirow{-2}{*}{\textbf{PCR}}}  & \cellcolor[HTML]{D9D9D9}$\overline{\sigma}$ & \cellcolor[HTML]{D9D9D9}0.007          & \cellcolor[HTML]{D9D9D9}0.011         & \cellcolor[HTML]{D9D9D9}0.010       & \cellcolor[HTML]{D9D9D9}0.010         & \cellcolor[HTML]{D9D9D9}0.010(\#2)          \\ \cline{2-2}
\multicolumn{1}{|l|}{}                               & \multicolumn{1}{l|}{}                                & $\overline{er}$                         & 0.009                       &   0.114                       &  0.011                               & 0.008                                & 0.036(\#3)                       \\
\multicolumn{1}{|l|}{}                               & \multicolumn{1}{c|}{\multirow{-2}{*}{\textbf{iPCR}}} & \cellcolor[HTML]{D9D9D9}$\overline{\sigma}$ & \cellcolor[HTML]{D9D9D9}0.011        & \cellcolor[HTML]{D9D9D9}0.129         & \cellcolor[HTML]{D9D9D9}0.013          & \cellcolor[HTML]{D9D9D9}0.011       & \cellcolor[HTML]{D9D9D9}0.041(\#4)          \\ \cline{2-2}
\multicolumn{1}{|l|}{}                               & \multicolumn{1}{l|}{}                                & $\overline{er}$                         & 0.074                        &  0.059         &    0.016          &   0.031                             & 0.045(\#4)                                  \\

\multicolumn{1}{|l|}{}                               & \multicolumn{1}{c|}{\multirow{-2}{*}{\textbf{CRL-SGNN}}} & \cellcolor[HTML]{D9D9D9}$\overline{\sigma}$ & \cellcolor[HTML]{D9D9D9}0.026         & \cellcolor[HTML]{D9D9D9}0.073        & \cellcolor[HTML]{D9D9D9}0.020        & \cellcolor[HTML]{D9D9D9}0.037         & \cellcolor[HTML]{D9D9D9}0.039(\#3)          \\ \cline{2-2}

\multicolumn{1}{|l|}{}                               & \multicolumn{1}{l|}{}                                & $\overline{er}$                         & 0.010                       &  0.018                              & 0.014                                &  0.014                                & \textbf{0.014(\#1)}                                  \\

\multicolumn{1}{|c|}{\multirow{-8}{*}{\textbf{QSN}}} & \multicolumn{1}{c|}{\multirow{-2}{*}{\textbf{Our}}}  & \cellcolor[HTML]{D9D9D9}$\overline{\sigma}$ & \cellcolor[HTML]{D9D9D9}0.006     & \cellcolor[HTML]{D9D9D9}0.012     & \cellcolor[HTML]{D9D9D9}0.010    & \cellcolor[HTML]{D9D9D9}0.009    & \cellcolor[HTML]{D9D9D9}\textbf{0.009(\#1)}  \\ \hline
\multicolumn{1}{|l|}{}                               & \multicolumn{1}{l|}{}                                & $\overline{er}$                         &   0.010                              &   0.011                     &   0.012                          &  0.011                               & 0.011(\#2)                                  \\
\multicolumn{1}{|l|}{}                               & \multicolumn{1}{c|}{\multirow{-2}{*}{\textbf{PCR}}}  & \cellcolor[HTML]{D9D9D9}$\overline{\sigma}$ & \cellcolor[HTML]{D9D9D9}0.005         & \cellcolor[HTML]{D9D9D9}0.007         & \cellcolor[HTML]{D9D9D9}0.008          & \cellcolor[HTML]{D9D9D9}0.007          & \cellcolor[HTML]{D9D9D9}0.007(\#2)          \\ \cline{2-2}
\multicolumn{1}{|l|}{}                               & \multicolumn{1}{l|}{}                                & $\overline{er}$                         &  0.009                                &  0.079                       &  0.011                             & 0.008                                &0.027(\#3)                                  \\
\multicolumn{1}{|l|}{}                               & \multicolumn{1}{c|}{\multirow{-2}{*}{\textbf{iPCR}}} & \cellcolor[HTML]{D9D9D9}$\overline{\sigma}$ & \cellcolor[HTML]{D9D9D9}0.013          & \cellcolor[HTML]{D9D9D9}0.093         & \cellcolor[HTML]{D9D9D9}0.014        & \cellcolor[HTML]{D9D9D9}0.012         & \cellcolor[HTML]{D9D9D9}0.033(\#3)          \\ \cline{2-2}
\multicolumn{1}{|l|}{}                               & \multicolumn{1}{l|}{}                                & $\overline{er}$                         & 0.023                   &  0.047               &  0.033                      & 0.029                    & 0.033(\#4)                        \\

\multicolumn{1}{|l|}{}                               & \multicolumn{1}{c|}{\multirow{-2}{*}{\textbf{CRL-SGNN}}} & \cellcolor[HTML]{D9D9D9}$\overline{\sigma}$ & \cellcolor[HTML]{D9D9D9}0.031           & \cellcolor[HTML]{D9D9D9}0.054         & \cellcolor[HTML]{D9D9D9}0.039     & \cellcolor[HTML]{D9D9D9}0.035        & \cellcolor[HTML]{D9D9D9}0.039(\#4)          \\ \cline{2-2}
\multicolumn{1}{|l|}{}                               & \multicolumn{1}{l|}{}                                & $\overline{er}$                         &  0.006                      & 0.012                     &  0.011                        &  0.010                        & \textbf{0.010(\#1)}                        \\

\multicolumn{1}{|c|}{\multirow{-8}{*}{\textbf{SW}}}  & \multicolumn{1}{c|}{\multirow{-2}{*}{\textbf{Our}}}  & \cellcolor[HTML]{D9D9D9}$\overline{\sigma}$ & \cellcolor[HTML]{D9D9D9}0.004     & \cellcolor[HTML]{D9D9D9}0.007     & \cellcolor[HTML]{D9D9D9}0.007     & \cellcolor[HTML]{D9D9D9}0.007    & \cellcolor[HTML]{D9D9D9}\textbf{0.006(\#1)} \\ \hline
\end{tabular}}
\end{table}}

{
\setlength{\tabcolsep}{3pt}
\begin{table}[t]
\centering
\caption{The average errors $\overline{er}$ and average standard deviations $\overline{\sigma}$ of the controllable robustness curve predicting for various types of networks under the TBA condition are compared among the NCR-HoK, PCR\cite{27}, iPCR\cite{28} models and CRL-SGNN\cite{CRL-SGNN} , along with a comprehensive ranking.}
\label{tab:table tb}
\scalebox{0.9}{
\begin{tabular}{|ccc|cccc|c|}
\hline
\multicolumn{3}{|c|}{}    & \textbf{\textless{}k\textgreater{}=2}  & \textbf{\textless{}k\textgreater{}=5}  & \textbf{\textless{}k\textgreater{}=8}  & \textbf{\textless{}k\textgreater{}=10} & \textbf{Average}    \\ \hline
\multicolumn{1}{|c|}{}   & \multicolumn{1}{l|}{}    & $\overline{er}$    & 0.047    & 0.033  & 0.018    & 0.016    & 0.029(\#3)    \\
\multicolumn{1}{|c|}{}   & \multicolumn{1}{c|}{\multirow{-2}{*}{\textbf{PCR}}}  & \cellcolor[HTML]{D9D9D9}$\overline{\sigma}$ & \cellcolor[HTML]{D9D9D9}0.017   & \cellcolor[HTML]{D9D9D9}0.015          & \cellcolor[HTML]{D9D9D9}0.012          & \cellcolor[HTML]{D9D9D9}0.010          & \cellcolor[HTML]{D9D9D9}0.014(\#2)          \\ \cline{2-2}

\multicolumn{1}{|c|}{}                               & \multicolumn{1}{l|}{}                                & $\overline{er}$                         &   0.080                                &   0.013                                 & 0.011                                  &  0.013                                 & 0.029(\#3)                                  \\
\multicolumn{1}{|c|}{}                               & \multicolumn{1}{c|}{\multirow{-2}{*}{\textbf{iPCR}}} & \cellcolor[HTML]{D9D9D9}$\overline{\sigma}$ & \cellcolor[HTML]{D9D9D9}0.086          & \cellcolor[HTML]{D9D9D9}0.009          & \cellcolor[HTML]{D9D9D9}0.011          & \cellcolor[HTML]{D9D9D9}0.013          & \cellcolor[HTML]{D9D9D9}0.030(\#3)          \\ \cline{2-2}

\multicolumn{1}{|c|}{}                               & \multicolumn{1}{l|}{}                                & $\overline{er}$                         &   0.014                                & 0.041                                  & 0.015                                  & 0.038                                 & 0.027(\#2)                                  \\
\multicolumn{1}{|c|}{}                               & \multicolumn{1}{c|}{\multirow{-2}{*}{\textbf{CRL-SGNN}}} & \cellcolor[HTML]{D9D9D9}$\overline{\sigma}$ & \cellcolor[HTML]{D9D9D9}0.015          & \cellcolor[HTML]{D9D9D9}0.052          & \cellcolor[HTML]{D9D9D9}0.014          & \cellcolor[HTML]{D9D9D9}0.038         & \cellcolor[HTML]{D9D9D9}0.030(\#3)          \\ \cline{2-2}

\multicolumn{1}{|c|}{}                               & \multicolumn{1}{l|}{}                                & $\overline{er}$                         &  0.017                       &  0.014                        &  0.016             &  0.012                 & \textbf{0.015(\#1)}                         \\
\multicolumn{1}{|c|}{\multirow{-8}{*}{\textbf{ER}}}  & \multicolumn{1}{c|}{\multirow{-2}{*}{\textbf{Our}}}  & \cellcolor[HTML]{D9D9D9}$\overline{\sigma}$ & \cellcolor[HTML]{D9D9D9}0.012 & \cellcolor[HTML]{D9D9D9}0.010 & \cellcolor[HTML]{D9D9D9}0.010 & \cellcolor[HTML]{D9D9D9}0.009 & \cellcolor[HTML]{D9D9D9}\textbf{0.010(\#1)} \\ \hline

\multicolumn{1}{|l|}{}                               & \multicolumn{1}{l|}{}                                & $\overline{er}$                         &    0.110                              & 0.069                                & 0.050                        &    0.044                              &   0.068(\#4)                               \\
\multicolumn{1}{|l|}{}                               & \multicolumn{1}{c|}{\multirow{-2}{*}{\textbf{PCR}}}  & \cellcolor[HTML]{D9D9D9}$\overline{\sigma}$ & \cellcolor[HTML]{D9D9D9}0.014        & \cellcolor[HTML]{D9D9D9}0.017      & \cellcolor[HTML]{D9D9D9}0.015         & \cellcolor[HTML]{D9D9D9}0.015        & \cellcolor[HTML]{D9D9D9}0.015(\#2)          \\ \cline{2-2}
\multicolumn{1}{|l|}{}                               & \multicolumn{1}{l|}{}                                & $\overline{er}$                         & 0.014          & 0.180                & 0.013                & 0.013                    & 0.055(\#3)                                  \\
\multicolumn{1}{|l|}{}                               & \multicolumn{1}{c|}{\multirow{-2}{*}{\textbf{iPCR}}} & \cellcolor[HTML]{D9D9D9}$\overline{\sigma}$ & \cellcolor[HTML]{D9D9D9}0.012          & \cellcolor[HTML]{D9D9D9}0.204        & \cellcolor[HTML]{D9D9D9}0.012          & \cellcolor[HTML]{D9D9D9}0.011        & \cellcolor[HTML]{D9D9D9}0.060(\#4)          \\ \cline{2-2}
\multicolumn{1}{|l|}{}                               & \multicolumn{1}{l|}{}                                  & $\overline{er}$                         &  0.038                       &   0.027                  & 0.022                       & 0.040                     & 0.032(\#2)                          \\

\multicolumn{1}{|l|}{}                               & \multicolumn{1}{c|}{\multirow{-2}{*}{\textbf{CRL-SGNN}}} & \cellcolor[HTML]{D9D9D9}$\overline{\sigma}$ & \cellcolor[HTML]{D9D9D9}0.042    & \cellcolor[HTML]{D9D9D9}0.030     & \cellcolor[HTML]{D9D9D9}0.021     & \cellcolor[HTML]{D9D9D9}0.042     & \cellcolor[HTML]{D9D9D9}0.034(\#3)         \\ \cline{2-2}
\multicolumn{1}{|l|}{}                               & \multicolumn{1}{l|}{}                                & $\overline{er}$                         &      0.010              &  0.014               &  0.014              &    0.013                   & \textbf{0.013(\#1)}                         \\

\multicolumn{1}{|c|}{\multirow{-8}{*}{\textbf{SF}}}  & \multicolumn{1}{c|}{\multirow{-2}{*}{\textbf{Our}}}  & \cellcolor[HTML]{D9D9D9}$\overline{\sigma}$ & \cellcolor[HTML]{D9D9D9}0.008         & \cellcolor[HTML]{D9D9D9}0.010        & \cellcolor[HTML]{D9D9D9}0.010         & \cellcolor[HTML]{D9D9D9}0.010       & \cellcolor[HTML]{D9D9D9}\textbf{0.010(\#1)}   \\ \hline

\multicolumn{1}{|l|}{}                               & \multicolumn{1}{l|}{}                                & $\overline{er}$                         &    0.027                    &  0.023                                 &  0.021                       & 0.019                         & 0.023(\#2)                                  \\
\multicolumn{1}{|l|}{}                               & \multicolumn{1}{c|}{\multirow{-2}{*}{\textbf{PCR}}}  & \cellcolor[HTML]{D9D9D9}$\overline{\sigma}$ & \cellcolor[HTML]{D9D9D9}0.012          & \cellcolor[HTML]{D9D9D9}0.012         & \cellcolor[HTML]{D9D9D9}0.013       & \cellcolor[HTML]{D9D9D9}0.012         & \cellcolor[HTML]{D9D9D9}0.012(\#2)          \\ \cline{2-2}
\multicolumn{1}{|l|}{}                               & \multicolumn{1}{l|}{}                                & $\overline{er}$                         & 0.017                       &   0.127                        &   0.014                               & 0.014                                & 0.043(\#4)                       \\
\multicolumn{1}{|l|}{}                               & \multicolumn{1}{c|}{\multirow{-2}{*}{\textbf{iPCR}}} & \cellcolor[HTML]{D9D9D9}$\overline{\sigma}$ & \cellcolor[HTML]{D9D9D9}0.013        & \cellcolor[HTML]{D9D9D9}0.137         & \cellcolor[HTML]{D9D9D9}0.012          & \cellcolor[HTML]{D9D9D9}0.013       & \cellcolor[HTML]{D9D9D9}0.044(\#4)          \\ \cline{2-2}
\multicolumn{1}{|l|}{}                               & \multicolumn{1}{l|}{}                                & $\overline{er}$                         &  0.038                       &  0.043                                &  0.020                              &  0.036                              & 0.034(\#3)                                  \\

\multicolumn{1}{|l|}{}                               & \multicolumn{1}{c|}{\multirow{-2}{*}{\textbf{CRL-SGNN}}} & \cellcolor[HTML]{D9D9D9}$\overline{\sigma}$ & \cellcolor[HTML]{D9D9D9}0.041          & \cellcolor[HTML]{D9D9D9}0.051         & \cellcolor[HTML]{D9D9D9}0.025        & \cellcolor[HTML]{D9D9D9}0.042          & \cellcolor[HTML]{D9D9D9}0.040(\#3)          \\ \cline{2-2}
\multicolumn{1}{|l|}{}                               & \multicolumn{1}{l|}{}                                & $\overline{er}$                         &  0.014                      & 0.015                               &  0.014                               & 0.013                                 & \textbf{0.014(\#1)}                                  \\

\multicolumn{1}{|c|}{\multirow{-8}{*}{\textbf{QSN}}} & \multicolumn{1}{c|}{\multirow{-2}{*}{\textbf{Our}}}  & \cellcolor[HTML]{D9D9D9}$\overline{\sigma}$ & \cellcolor[HTML]{D9D9D9}0.008     & \cellcolor[HTML]{D9D9D9}0.009     & \cellcolor[HTML]{D9D9D9}0.009    & \cellcolor[HTML]{D9D9D9}0.009    & \cellcolor[HTML]{D9D9D9}\textbf{0.009(\#1)}  \\ \hline
\multicolumn{1}{|l|}{}                               & \multicolumn{1}{l|}{}                                & $\overline{er}$                         &   0.014                              &  0.021                      & 0.018                            & 0.016                                & 0.017(\#2)                                  \\
\multicolumn{1}{|l|}{}                               & \multicolumn{1}{c|}{\multirow{-2}{*}{\textbf{PCR}}}  & \cellcolor[HTML]{D9D9D9}$\overline{\sigma}$ & \cellcolor[HTML]{D9D9D9}0.006         & \cellcolor[HTML]{D9D9D9}0.013         & \cellcolor[HTML]{D9D9D9}0.012          & \cellcolor[HTML]{D9D9D9}0.010          & \cellcolor[HTML]{D9D9D9}\textbf{0.010(\#1)}          \\ \cline{2-2}
\multicolumn{1}{|l|}{}                               & \multicolumn{1}{l|}{}                                & $\overline{er}$                         & 0.012                                 &  0.096                       & 0.016                              & 0.013                                &0.034(\#4)                                  \\
\multicolumn{1}{|l|}{}                               & \multicolumn{1}{c|}{\multirow{-2}{*}{\textbf{iPCR}}} & \cellcolor[HTML]{D9D9D9}$\overline{\sigma}$ & \cellcolor[HTML]{D9D9D9}0.013          & \cellcolor[HTML]{D9D9D9}0.099         & \cellcolor[HTML]{D9D9D9}0.015        & \cellcolor[HTML]{D9D9D9}0.013         & \cellcolor[HTML]{D9D9D9}0.035(\#4)          \\ \cline{2-2}
\multicolumn{1}{|l|}{}                               & \multicolumn{1}{l|}{}                                & $\overline{er}$                         &  0.015                  &  0.037               &  0.034                      & 0.040                    & 0.032(\#3)                        \\

\multicolumn{1}{|l|}{}                               & \multicolumn{1}{c|}{\multirow{-2}{*}{\textbf{CRL-SGNN}}} & \cellcolor[HTML]{D9D9D9}$\overline{\sigma}$ & \cellcolor[HTML]{D9D9D9}0.015           & \cellcolor[HTML]{D9D9D9}0.042         & \cellcolor[HTML]{D9D9D9}0.037     & \cellcolor[HTML]{D9D9D9}0.047        & \cellcolor[HTML]{D9D9D9}0.033(\#3)          \\ \cline{2-2}
\multicolumn{1}{|l|}{}                               & \multicolumn{1}{l|}{}                                & $\overline{er}$                         &  0.009                      & 0.013                     &  0.014                        & 0.013                         & \textbf{0.012(\#1)}                        \\

\multicolumn{1}{|c|}{\multirow{-8}{*}{\textbf{SW}}}  & \multicolumn{1}{c|}{\multirow{-2}{*}{\textbf{Our}}}  & \cellcolor[HTML]{D9D9D9}$\overline{\sigma}$ & \cellcolor[HTML]{D9D9D9}0.004     & \cellcolor[HTML]{D9D9D9}0.013     & \cellcolor[HTML]{D9D9D9}0.014     & \cellcolor[HTML]{D9D9D9}0.013    & \cellcolor[HTML]{D9D9D9}0.011(\#2) \\ \hline
\end{tabular}}
\end{table}}

\end{document}